%% file: 00_aaai_23.tex
\documentclass[letterpaper]{article} 
\usepackage{aaai23}  
\usepackage{times}  
\usepackage{helvet}  
\usepackage{courier}  
\usepackage[hyphens]{url}  
\usepackage{graphicx} 
\def\UrlFont{\rm}  
\usepackage{natbib}  
\usepackage{caption} 
\usepackage{booktabs}       
\usepackage{amsfonts}       
\usepackage{nicefrac}       
\usepackage{microtype}      
\usepackage{xcolor}         
\usepackage{soul}
\usepackage[ruled,linesnumbered]{algorithm2e}
\usepackage{amsmath}
\usepackage{multirow}
\usepackage[flushleft]{threeparttable}
\usepackage{epsfig}  
\usepackage{subcaption}
\usepackage{paralist}
\usepackage{enumitem}
\usepackage{amssymb}
\usepackage{listings}
\usepackage{wrapfig,lipsum,booktabs}
\usepackage{newfloat}
\usepackage{float}
\usepackage{listings}
\DeclareCaptionStyle{ruled}{labelfont=normalfont,labelsep=colon,strut=off} 
\floatstyle{ruled}
\newfloat{listing}{tb}{lst}{}
\floatname{listing}{Listing}

\DeclareMathOperator*{\argmax}{arg\,max}
\DeclareMathOperator*{\argmin}{arg\,min}
\usepackage{mathtools}
\DeclarePairedDelimiter\abs{\lvert}{\rvert}%
\DeclarePairedDelimiter\norm{\lVert}{\rVert}%
\makeatletter
\let\oldabs\abs
\def\abs{\@ifstar{\oldabs}{\oldabs*}}
\let\oldnorm\norm
\def\norm{\@ifstar{\oldnorm}{\oldnorm*}}
\makeatother

\newcommand{\best}[1]{\color{red}{\textbf{#1}}}
\newcommand{\secondbest}[1]{\color{blue}{#1}}

\newcommand{\beginsupplement}{%
        \setcounter{table}{0}
        \renewcommand{\thetable}{\thesection\arabic{table}}%
        \setcounter{figure}{0}
        \renewcommand{\thefigure}{\thesection\arabic{figure}}%
     }

\title{Zero-Cost Operation Scoring in Differentiable Architecture Search}
\author {
  Lichuan Xiang\textsuperscript{1$\dagger$},  {\L}ukasz Dudziak\textsuperscript{2$\dagger$}, Mohamed S. Abdelfattah\textsuperscript{3$\dagger$} \\
  \textbf{Thomas Chau\textsuperscript{2}, Nicholas D. Lane\textsuperscript{2,4}, Hongkai Wen\textsuperscript{1,2}
  } \\
}
\affiliations {
  \textsuperscript{1} University of Warwick, UK \hspace*{10pt} 
  \textsuperscript{2} Samsung AI Center Cambridge, UK\\
  \textsuperscript{3} Cornell University, USA \hspace*{10pt} 
  \textsuperscript{4} University of Cambridge, UK \\
  \footnotesize{$^{\dagger}$\textit{Indicates equal contributions}} \\
  \texttt{\{l.xiang.2, hongkai.wen\}@warwick.ac.uk} \\
  \texttt{\{l.dudziak, thomas.chau, nic.lane\}@samsung.com} \\
  \texttt{mohamed@cornell.edu} \\
}

\usepackage{bibentry}

\begin{document}

\maketitle
\input{01-abstract}

\input{10-intro}

\input{20-related}

\input{30-scoring}

\input{40-zcpt}

\input{50-results}

\input{60-conclusion}
\bibliography{main}

\newpage
\appendix
\onecolumn
\section{Appendix}
\input{90_appendix}


\end{document}

%% file: 01-abstract.tex
\begin{abstract}

We formalize and analyze a fundamental component of differentiable neural architecture search (NAS): local ``operation scoring'' at each operation choice.
We view existing operation scoring functions as inexact proxies for accuracy, and we find that they perform poorly when analyzed empirically on NAS benchmarks.
From this perspective, we introduce a novel \textit{perturbation-based zero-cost operation scoring} (Zero-Cost-PT) approach, which utilizes zero-cost proxies that were recently studied in multi-trial NAS but degrade significantly on larger search spaces, typical for differentiable NAS. 
We conduct a thorough empirical evaluation on a number of NAS benchmarks and large search spaces, from NAS-Bench-201, NAS-Bench-1Shot1, NAS-Bench-Macro, to DARTS-like and MobileNet-like spaces, showing significant improvements in both search time and accuracy. 
On the ImageNet classification task on the DARTS search space, our approach improved accuracy compared to the best current training-free methods (TE-NAS) while being over $10\times$ faster (total searching time 25 minutes on a single GPU), and observed significantly better transferability on architectures searched on the CIFAR-10 dataset with an accuracy increase of 1.8 pp. 
Our code is available at: \url{https://github.com/zerocostptnas/zerocost_operation_score}.

\end{abstract}

%% file: 10-intro.tex

\section{Introduction}\label{sec:intro}



One of the biggest problems in neural architecture search (NAS) is the computational cost -- even training a single deep network can require enormous computational resources, and many NAS methods need to train tens, if not hundreds, of networks in order to converge to a good architecture~\citep{agingevo,nao2018,brpnas}. 
A related problem concerns search space size---a larger NAS search space would typically contain better architectures but requires a longer searching time~\citep{agingevo}.
Differentiable architecture search (DARTS) was first proposed to tackle those challenges, showcasing promising results when searching for a network in a set of over $10^{18}$ possible variations~\citep{darts}.
Unfortunately, DARTS has significant robustness issues, as demonstrated through many recent works~\citep{arber2020understanding,shu2020understanding,yu2020evaluation}. 
It also requires a careful selection of hyperparameters, making it somewhat difficult to adapt to a new task. 
Recently, \cite{darts-pt} showed that operation selection in DARTS based on the magnitude of architectural parameters ($\alpha$) is fundamentally wrong and will always simply select skip connections over other more meaningful operations. 
They proposed an alternative operation selection method based on \textit{perturbation}, where the importance of an operation is determined by the decrease of the supernet's validation accuracy when it is removed. 
Then the most important operations are selected by exhaustively comparing them with other alternatives on each single edge of the supernet until the final architecture is found.

In a parallel line of work that aims to speed up NAS, \textit{proxies} are often used instead of training accuracy to quickly obtain an indication of performance without expensive full training for each searched model.
Conventional proxies typically consist of a reduced form of training with fewer epochs, less data or a smaller DNN architecture~\citep{econas}.
Most recently, \textit{zero-cost proxies}, which are extreme types of NAS proxies that do not require any training, have gained interest and are shown to empirically outperform conventional training-based proxies and deliver outstanding results on common NAS benchmarks~\citep{zero-cost,nwot}.
However, their efficient usage on a large search space, typical for differentiable NAS, has been shown to be more challenging and thus remains an open problem~\citep{nwot}.

The objective of our paper is to shed some light onto the implicit proxies that are used for operation scoring in differentiable NAS, and to discover new proxies in this setting that have the potential of improving both search speed and quality.
We decompose differentiable NAS into its two constituent parts: (1) supernet training and (2) operation scoring.
We focus on the second component and formalize the concept of ``operation scoring'' that happens during local operation selection at each edge in a supernet.
Through this lens, we are able to empirically compare the efficacy of existing differentiable NAS operation scoring functions.
We find that existing methods act as a proxy for accuracy and perform poorly on NAS benchmarks. Consequently, we propose new operation scoring functions based on zero-cost proxies that outperform existing methods on both search speed and accuracy.
Our main contributions are:
\begin{itemize}[leftmargin=8mm, topsep=-2pt]
    \setlength{\itemsep}{0pt}
    \setlength{\parskip}{0pt}
    \setlength{\parsep}{0pt}
    \item Formalize \textit{operation scoring} in differentiable NAS and perform a first-of-its-kind analysis of the implicit proxies that are present in existing methods.
    \item Propose, evaluate and compare \textit{perturbation-based zero-cost operation scoring} (Zero-Cost-PT) for differentiable NAS building upon recent work on training-free NAS proxies.
    \item Perform a thorough empirical evaluation of Zero-Cost-PT in multiple search spaces and datasets, including DARTS, DARTS subspaces S1-S4, MobileNet-like space, and 3 popular NAS benchmarks: NAS-Bench-201, NAS-Bench-1shot1 and NAS-Bench-Macro.
\end{itemize}

%% file: 20-related.tex
\section{Related work}\label{sec:related}

\textbf{Classic NAS and Proxies.}
Zoph \& Lee were among the first to propose an automated method to search neural network architectures, using a reinforcement learning agent to maximize rewards coming from training different models~\citep{nas}. 
Since then, a number of alternative approaches have been proposed in order to reduce the significant cost introduced by training each proposed model. 
In general, reduced training can be found in many NAS works~\citep{enas,econas}, and different proxies have been proposed, e.g. searching for a model on a smaller dataset and then transferring the architecture to the larger target dataset~\citep{agingevo,nasbenchasr}, or incorporating a predictor into the search process~\citep{npenas2020,brpnas,weak_predictor2021,neural_predictor2019}.


\noindent\textbf{Zero-cost Proxies.}
Recently, zero-cost proxies~\citep{nwot,zero-cost} for NAS emerged from pruning-at-initialisation literature~\citep{synflow,grasp,snip,blockswap}.
Such proxies can be formulated as architecture scoring functions $S(A)$ that evaluate the ``saliency'' of a given architecture $A$ in achieving accuracy at initialization without the expensive training process. 
In this paper, we adopt the recently proposed zero-cost proxies~\citep{zero-cost,nwot}, namely \texttt{grad\_norm}, \texttt{snip}, \texttt{grasp}, \texttt{synflow}, \texttt{fisher} and \texttt{nwot}.
Those metrics either aggregate the saliency of model parameters to compute the score of an architecture~\citep{zero-cost}, or use the overlapping of activations between different samples within a minibatch of data as a performance indicator~\citep{nwot}.
In a similar vein, \cite{chen2020tenas} proposed the use of training-free scoring for operations based on the neural tangent kernel~\citep{ntk} and number of linear regions; operations with the lowest score are \textit{pruned} from the supernet iteratively until a subnetwork is found.

\noindent\textbf{Differentiable NAS and Operation Perturbation.}
Liu et al. first proposed to search for a neural network's architecture by parameterizing it with continuous values (called architectural parameters $\alpha$) in a differentiable way. 
Their method constructs a \textit{supernet}, i.e., a superposition of all networks in the search space, and optimizes the architectural parameters ($\alpha$) together with supernet weights ($w$). 
The final architecture is extracted from the supernet by preserving operations with the largest $\alpha$. Despite the significant reduction in searching time, the stability and generalizability of DARTS have been challenged, e.g., it may produce trivial models dominated by skip connections~\citep{arber2020understanding}. 
SDARTS~\citep{chen2020stabilizing} proposed to overcome such issues by smoothing the loss landscape, while SGAS~\citep{li2020sgas} considered a greedy algorithm to select and prune operations sequentially. 
The recent DARTS-PT~\citep{darts-pt} proposed a perturbation-based operation selection strategy, showing promising results on DARTS space. In DARTS-PT operations are no longer selected by optimizing architectural parameters ($\alpha$), but via a scoring function evaluating the impact on a supernet's validation accuracy when they are removed. 

%% file: 30-scoring.tex
\section{Rethinking Operation Scoring }
\label{sec:scoring}

In the context of differentiable NAS, a supernet would contain multiple candidate operations on each edge as shown in Figure~\ref{discretize-perturb}.
Operation scoring functions assign a score to rank operations and select the best one.
In this section, we empirically quantify the effectiveness of existing operation scoring methods in differentiable NAS, with a specific focus on DARTS~\citep{darts} and the recently-proposed DARTS-PT~\citep{darts-pt}.
Concretely, we view these scoring functions as proxies for final subnetwork accuracies and we evaluate them on that basis to quantify how well these functions perform.
We challenge many  assumptions made in previous work and show that we can outperform existing methods with lightweight alternatives.

\subsection{Operation Scoring Preliminaries}
\label{sub:op-eval}

\begin{figure*}[t]
\centering
\caption{Visualization of perturbation and discretization of an edge in a supernet. \textbf{Middle}: a supernet is composed of three edges $\{e^{(i)}\}_{i=1,2,3}$, each consisting of three possible operations $\{o^{(i)}\}_{i=1,2,3}$ which are applied in parallel to the same input. \textbf{Left}: edge $e^{(2)}$ is perturbed by removing $o^{(1)}$ from the set of candidate operations assigned to this edge. \textbf{Right}: the same edge $e^{(2)}$ is discretized with operation $o^{(1)}$ by removing all other candidate operations leaving $o^{(1)}$ as the only choice left.}
 \includegraphics[width=0.75\textwidth, trim=0 0 0 0cm]{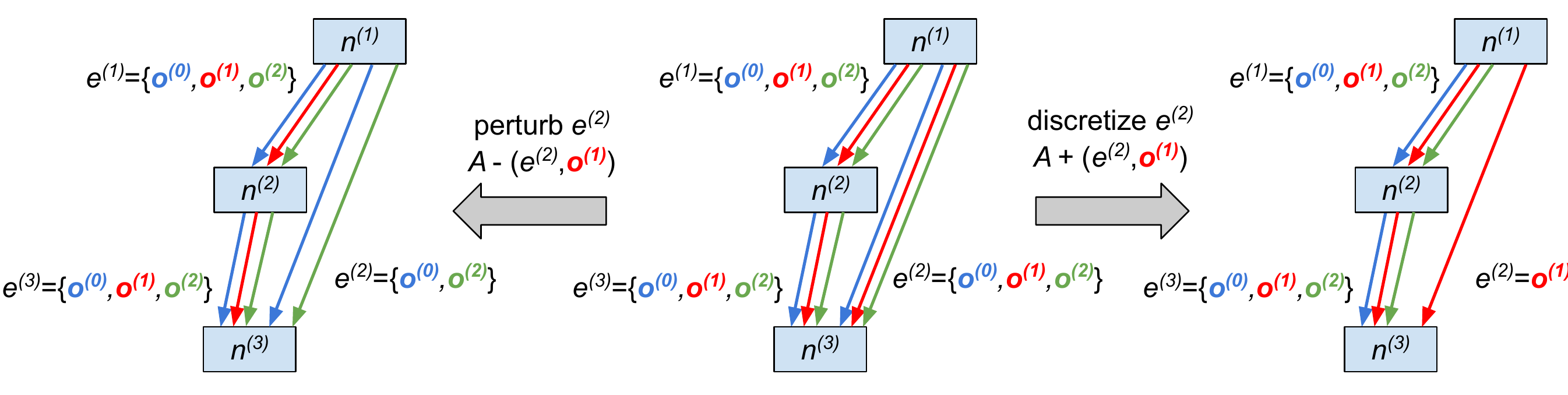}
  \label{discretize-perturb}

\end{figure*}

For a supernet $A$ we want to be able to start \textit{discretizing} edges in order to derive a subnetwork.
When discretizing we replace an edge composed of multiple candidate operations and their respective (optional) architectural parameters $\alpha$ with only one operation selected from the candidates. 
We will denote the process of discretization of an edge $e$ with operation $o$, given a model $A$, as: $A+(e,o)$.
Analogously, the \textit{perturbation} of a supernet $A$ by removing an operation $o$ from an edge $e$ will be denoted as $A - (e,o)$.
Figure~\ref{discretize-perturb} illustrates discretization and perturbation.
Furthermore, we will use $\mathcal{A}$, $\mathcal{E}$ and $\mathcal{O}$ to refer to the set of all possible network architectures, edges in the supernet and candidate operations, respectively. More details about notation can be found in Appendix~\ref{sec:app:notation}.

NAS can then be performed by iterative discretization of edges in the supernet, yielding in the process a sequence of partially discretized architectures: $A_0,A_1,...,A_{|\mathcal{E}|}$, where $A_0$ is the original supernet, $A_{|\mathcal{E}|}$ is the final fully-discretized subnetwork (result of NAS), and $A_t$ is $A_{t-1}$ after discretizing a next edge, i.e., $A_t = A_{t-1}+(e_t,o_t)$ where $t$ is an iteration counter.
The problem of finding the sequence of $(e_t, o_t)$ that maximizes the performance of the resulting network $A_{|\mathcal{E}|}$ has an optimal substructure and can be reduced to the problem of finding the optimal policy $\pi: \mathcal{A} \times \mathcal{E} \rightarrow \mathcal{O}$ that is used to decide on an operation to assign to an edge at each iteration, given current model (state).
This policy function is defined by means of an analogous scoring function $f: \mathcal{A} \times \mathcal{E} \times \mathcal{O} \rightarrow \mathbb{R}$, that assigns scores to the possible values of the policy function, and then taking $\argmax$ or $\argmin$ over $f$, depending on the type of scores produced by $f$.
\footnote{Since a scoring function clearly defines a relevant policy function, we will sometimes talk about a scoring function even though the context might be directly related to a policy function -- in those cases, it should be understood as the policy function that follows from the relevant scoring function (and vice versa).}

We begin by defining the optimal scoring function that we will later use to assess the quality of different empirical approaches.
For a given partially-discretized model $A_t$, let us denote the set of all possible fully-discretized networks that can be obtained from $A_t$ after a next edge $e$ is discretized with an operation $o$ as $\mathcal{A}_{t,e,o}$.
Our optimal scoring function can then be defined as:
\begin{equation}\label{eq:opsel:best}
    \pi_{\text{best-acc}}(A_t,e) = \argmax_{o \in \mathcal{O}_e} \max_{A_{|\mathcal{E}|} \in \mathcal{A}_{t,e,o}} V^{*}(A_{|\mathcal{E}|})
\end{equation}
where $V^{*}$ is the validation accuracy of a network after converged (we will use $V$ to denote validation accuracy without training), and $\mathcal{O}_e \subseteq \mathcal{O}$ is the subset of candidate operations that are considered for edge $e$.
It is easy to see that this policy meets Bellman's principle of optimality~\citep{Bellman:DynamicProgramming} -- the definition follows directly from it and therefore is the optimal solution to our problem.
However, it might be more practical to consider the expected achievable accuracy when an operation is selected, instead of the best.
Therefore we define the function $\pi_{\text{avg-acc}}$:
\begin{equation}\label{eq:opsel:avg}
    \pi_{\text{avg-acc}}(A_t,e) = \argmax_{o \in \mathcal{O}_e} \mathop{\mathbb{E}}_{A_{|\mathcal{E}|} \in \mathcal{A}_{t,e,o}} V^{*}(A_{|\mathcal{E}|})
\end{equation}
In practice, we are unable to use either $\pi_{\text{best-acc}}$ or $\pi_{\text{avg-acc}}$ since we would need to have the final validation accuracy $V^*$ of all the networks in the search space.
Here we consider the following practical alternatives from DARTS~\citep{darts} and the recent DARTS-PT~\citep{darts-pt}:
\begin{align}
    \pi_{\text{darts}}(A_t,e) & = \argmax_{o \in \mathcal{O}_e} \alpha_{e,o} \label{eq:opsel:darts} \\
     \pi_{\text{disc-acc}}(A_t,e) &= \argmax_{o \in \mathcal{O}_e} V^{*}(A_t+(e,o)) \label{eq:opsel:disc-acc} \\
      \pi_{\text{darts-pt}}(A_t,e) &= \argmin_{o \in \mathcal{O}_e} V(A_t-(e,o)) \label{eq:opsel:darts-pt}
\end{align}
%
where $\alpha_{e,o}$ is the architectural parameter assigned to operation $o$ on edge $e$ as presented in DARTS~\citep{darts}.
$\pi_{\text{disc-acc}}$ uses accuracy of a supernet after an operation $o$ is assigned to an edge $e$ -- this is referred to as ``discretization accuracy'' in DARTS-PT and is assumed to be a good operation scoring function~\citep{darts-pt}, it could approximate $f_{\text{avg-acc}}$.
$\pi_{\text{darts-pt}}$ is the perturbation-based approach used by DARTS-PT -- it is presented as a practical and lightweight alternative to $\pi_{\text{disc-acc}}$~\citep{darts-pt}.

\noindent\textbf{Zero-Cost Operation Scoring.} 
We argue that the scoring functions \ref{eq:opsel:darts}-\ref{eq:opsel:darts-pt} are merely proxies for the best achievable accuracy (Eq.~\ref{eq:opsel:best}).
As such, we see an opportunity to use a new class of training-free proxies that are very fast to compute and have been shown to work well within multi-trial NAS, albeit not in differentiable NAS, nor within large search spaces.
We present the following scoring functions that use a zero-cost proxy $S$ instead of validation accuracy when discretizing an edge or perturbing an operation.
Note that the supernet is randomly-initialized and untrained.
\begin{align}
    \pi_{\text{disc-zc}}(A_t,e) &= \argmax_{o \in \mathcal{O}_e} S(A_t+(e,o)) \label{eq:opsel:disc-zc} \\
    \pi_{\text{zc-pt}}(A_t,e) &= \argmin_{o \in \mathcal{O}_e} S(A_t-(e,o)) \label{eq:opsel:zc-pt}
\end{align}
%
In the rest of this paper, we consider the following proxies that have been proposed in recent zero-cost NAS literature: \texttt{grad\_norm}~{\citep{zero-cost}}, \texttt{snip}~{\citep{snip}}, \texttt{grasp}~{\citep{grasp}}, \texttt{synflow}~{\citep{synflow}}, \texttt{fisher}~{\citep{fisher-gaze}}, \texttt{zen\_score}~{\citep{ming_zennas_iccv2021}}, \texttt{tenas}~{\citep{chen2020tenas}} and \texttt{nwot}~{\citep{nwot}}. Detailed metrics descriptions are included in Appendix~{\ref{sec:app:zc-proxies}.
Note that in most existing work~\mbox{\citep{zero-cost,nwot}}, zero-cost metrics are not used to score operations but to select architectures based on their end-to-end scores, their effectiveness on operation selection remains to discover, and we are going to show that building operation scoring function and algorithm upon on them is trivial, while TE-NAS~{\citep{chen2020tenas}} also uses them to score operations. However, as opposed to \textit{selecting} the optimal operations (via either \textit{discretization} or \textit{perturbation}), the \texttt{tenas} metric is used to iteratively \textit{prune} the weakest operations from a supernet~{\citep{chen2020tenas}}.} 

\subsection{Empirical Analysis on Operation Scoring}
\label{sub:op-eval-corr}

In this subsection, we investigate the performance of different operation scoring methods.
Because we want to compare with the optimal best-acc and avg-acc, 
we conduct experiments on two popular NAS benchmarks: NAS-Bench-201~{\citep{nasbench2}} and NAS-Bench-1Shot1~{\citep{nasbench-1shot1}}.
In the following, we discuss our findings using NAS-Bench-201, while results on NAS-Bench-1Shot1 can be found in Appendix~{\ref{sec:app:nb1s1}}.
We conduct our investigation in two settings, \textit{initial} and \textit{progressive}. 
The first setting compares operation scoring functions while making their first decision (iteration 0) during NAS.
The second, \textit{progressive}, setting takes into account retraining of a partially discretized supernet $A_t$ and a subsequent rescoring of operations, that might occur between iterations of different algorithms that we consider like darts-pt~\citep{darts-pt}.



\begin{figure*}[t]
 \centering
   \caption{\textbf{Left}: Rank correlation coefficient between different operation scoring metrics at the first iteration of NAS. \textbf{Right}: Rank correlation coefficient of different operation scoring functions vs. best-acc when invoked iteratively for each edge. In iteration $i$, only edge $i$ is discretized then all scores for all operations on the remaining edges is computed and correlated against best-acc.}
  \includegraphics[width=0.75\textwidth, trim=0 0 0 0cm]{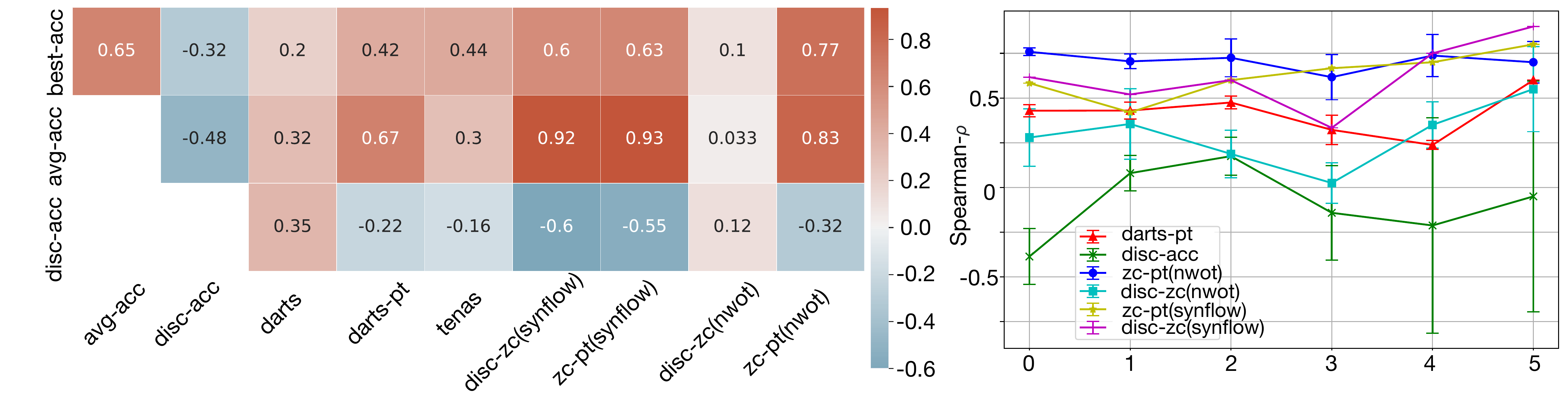}
  \label{op}

\end{figure*}

\noindent\textbf{Initial Operation Scoring.}
For the supernet $A_0$ we compute the operation scores for all operations on all edges, at the first iteration (iteration 0) of NAS, that is, $f(A_0,e,o)\,\, \forall\,\, e \in \mathcal{E},\,\, o \in \mathcal{O}_e$.
In our first experiment, we collect the scores produced by different scoring methods, per operation, per edge, then compute the Spearman rank correlation for operations on each edge, and finally average the rank correlation coefficient over all edges (details of our experiments and illustrative examples are provided in Appendix~\ref{sec:app:scores}). 
The resulting averaged rank correlation is indicative of how well an operation scoring method would do when making the first discretization decision, relative to a perfect ``oracle'' search. 
We plot the rank correlation coefficients in Figure~\ref{op}, showing many surprising findings. 
First, the original darts $\alpha$ score is weakly and inversely correlated with the oracle scores, further supporting arguments in prior work that this is not an effective operation scoring method. 
Second, disc-acc is inversely correlated to best-acc. 
This refutes the claim in the DARTS-PT paper that disc-acc is a reasonable operation score~\citep{darts-pt} -- these findings are aligned with prior work that has already shown that the supernet accuracy is unrelated to the final subnetwork accuracy~\citep{li2020sgas}.
Third, the darts-pt score does not track disc-acc, in fact, it is inversely-correlated to it as well, meaning that the darts-pt score is not a good approximation of disc-acc. 
However, darts-pt is weakly correlated to the ``oracle'' best-acc and avg-acc scores which supports (empirically) why it works well.
Fourth, tenas~\citep{chen2020tenas}, which also utilizes training-free operation scoring, performs fairly well, with Spearman-$\rho=$0.44, but still falls short of the performance of the two zc-pt variants ($\rho=$0.77 and 0.63).
Finally, our zc-pt, when using either \texttt{synflow} or \texttt{nwot} metric, is strongly correlated with both the best-acc and avg-acc metrics, indicating that there could be huge promise when using this scoring function within NAS.
Note that disc-zc, in particular when using \texttt{nwot} metric, is only weakly correlated with the oracle scores, suggesting that \textit{perturbation} is a more robust scoring paradigm than \textit{discretization}.
We provide more analysis on disc-zc \textit{vs.} later, and compare NAS results when using either scoring method in Appendix~{\ref{sec:app:pt_vs_disc}}.


In Table~\ref{tbl:op-strength}, we show the discovered NAS-Bench-201 architecture when applying the seven scoring functions (Eq.~\eqref{eq:opsel:best} -- \eqref{eq:opsel:zc-pt}) for operation selection on all edges.
As expected, best-acc chooses the best subnetwork, while avg-acc selects a very good model but not the best one, likely due to the large variance of accuracies in NAS-Bench-201.
zc-pt(\texttt{nwot}) selected one of the top models in NAS-Bench-201 as expected from the strong correlation with the oracle best-acc function; while tenas selected a good model, in the top 15\% of the NAS-Bench-201 dataset, commensurate with the average correlation shown in Figure~\ref{op}.
The remaining operation scoring functions failed to produce a good model in this experiment, suggesting that these metrics do not make a good initial choice of operations at iteration 0 of differentiable NAS. A similar analysis of NAS-Bench-1shot1 search space can be found in Appendix~{\ref{sec:app:nb1s1}}.
We provide a more detailed look into the failures of certain scoring methods below.


\begin{table}[t]
\setlength\tabcolsep{2pt}
\centering
\caption {Model selected based on maximizing each operation strength independently.}
\footnotesize
\resizebox{\columnwidth}{!}{
    \begin{threeparttable}
    \begin{tabular}{@{}l@{}cccccc@{}}
    \toprule
                              & \textbf{best-acc} & \textbf{avg-acc } & \textbf{disc-acc} & \textbf{darts-pt}\\ \toprule
     Avg. Error\tnote{1} [\%] & 5.63              & 6.24              & 13.55             & 19.43  \\ 
     Rank in NAS-Bench-201    & 1                 & 166               & 12,744            & 13,770       \\\toprule
                           &\textbf{zc-pt(\texttt{nwot})}    & \textbf{disc-zc(\texttt{nwot})} & \textbf{darts}  & \textbf{tenas}\\ \toprule
    Avg. Error\tnote{1} [\%]  & 5.81             & 22.96            & 45.7            & 7.19  \\ 
     Rank in NAS-Bench-201     & 14               & 14,274           & 15,231          & 1,817       \\\toprule
    \end{tabular}
    \begin{tablenotes}
    \scriptsize
    \item[1] Computed as the average of all available seeds for the selected model in NAS-Bench-201 CIFAR-10 dataset.
    \end{tablenotes}
    \end{threeparttable}
}
\label{tbl:op-strength}
\end{table}


\noindent\textbf{Analysis of the darts-pt and disc-acc scoring.}
As mentioned before, our zc-pt operation scoring function outperforms both darts-pt and disc-acc, despite the latter methods relying directly on accuracy.
This may sound counter-intuitive, but it becomes clearer if we note that the accuracy used by these methods (supernet accuracy) is not directly relevant to the NAS objective (subnet accuracy).
Regarding darts-pt, we argue that the unrolled estimation performed by a supernet, as described by \cite{darts-pt}, might lead to the observed preference towards selecting skip connections (see Table~\ref{tbl:raw-op-scores}).
This is because under this hypothesis, convolutional operations on an edge perform only refinement of the input, while most of the information is carried directly from the input through a skip connection.
Therefore, it can be expected that removal of the skip connection should have much severe effects on the supernet's performance.
More information can be found in Appendix~\ref{sec:app:additional-disc-acc}.


Regarding disc-acc, we perform an additional case study of how the supernet's accuracy changes after discretizing an edge with different operations, as a function of training epochs.
The experiment details are in Appendix~\ref{sec:app:additional-disc-acc}.
From the results we can see that as retraining progresses, lighter operations (\texttt{none} in particular) converge much faster than heavy, but potentially more meaningful, operations like \texttt{conv\_1x1}. 
What is more, even after sufficiently long training, all choices converge to roughly the same point.
We argue that this is caused by the fact that in the early NAS iterations the supernet is heavily overparamtrized, which hinders our ability to faithfully measure each operation's contribution based simply on the accuracy of the supernet.


\noindent\textbf{Analysis of the zc-pt and disc-zc scoring.}
We further perform detailed experiments on comparing both \textit{pertubation} and \textit{discretization} policies, especially when using \texttt{nwot} zero-cost metric.
Specifically, we create a toy model that allows us to observe how \texttt{disc-zc(nwot)} and \texttt{zc-pt(nwot)} behave in a simplified setting, as we vary the model's depth.
We observe that while both approaches behave similarly when a network is shallow, as we increase its depth \texttt{disc-zc} quickly degrades and becomes biased towards selecting skip connection, which is not the case for \texttt{zc-pt}.
This suggests that \texttt{nwot} might in fact prefer shallower networks -- however, unlike discretization, perturbation paradigm does not introduce the ability to reduce the supernet's depth (as long as each edge includes at least two meaningful operations among their candidates), which seems to robustify \texttt{nwot} significantly.
More detailed can be found in Appendix~\ref{sec:app:additional-op-scoring}.

While the above signals some major weaknesses of different proxies used in differentiable NAS, when used to perform initial scoring of operations, it's worthwhile to further analyze them in the \textit{progressive} setting which would show what happens in later NAS iterations.

\noindent\textbf{Progressive Operation Scoring.}
Until now, we have only investigated the performance of operation scoring functions in the first iteration of NAS.
This approach is relevant for methods like DARTS, where operation scoring function $f$ does not depend on $A_t$ in any way (only $A_0$), but is not truly representative of other methods that work iteratively.
Because of that, we extend our analysis to investigate what happens in later iterations of NAS.
To do that, we calculate the correlation of scoring functions in the \textit{progressive} setting by performing the following steps: (1) score operations on all undiscretized edges, (2) discretize edge $i$, (3) retrain for 5 epochs (darts-pt and disc-acc only), (4) increment $i$ and repeat from step 1 until all edges are discretized. At each iteration $i$, we calculate the scores for the operations on all remaining undiscretized edges and compute their Spearman-$\rho$ rank correlation coefficients with respect to best-acc.
This is plotted in Figure~\ref{op}, averaged over 4 seeds.

Our results confirm many of our \textit{initial} (iteration-0) analysis.
zc-pt(\texttt{nwot}) continues to be the best operation scoring function, and darts-pt is the second-best, improving in correlation from 0.4 to 0.6 between the first and last iterations, indeed showing that retraining and/or progressive discretization helps.
However, disc-acc continues to be unrepresentative of operation strength even when used in the iterative setting.
This is not what we expected, especially in the very last iteration when disc-acc is supposed to match a subnetwork exactly.
As Figure~\ref{op} shows, the variance in the last iteration is quite large -- we believe this happens because we do not train to convergence every time we discretize an edge, and instead we only train for 5 epochs.
Our progressive analysis provided further empirical evidence that supernet discretization accuracy should not be used as a proxy for subnetwork accuracy, contradicting \cite{darts-pt}.
However, we have confirmed that darts-pt does in fact improve when retraining is performed between NAS iterations, but could still be improved upon with zc-pt  -- it performed exceptionally well as a proxy for accuracy and has the potential to make differentiable NAS both much faster and of higher accuracy.

%% file: 40-zcpt.tex
\section{Zero-Cost-PT Neural Architecture Search}
\label{sec:zc-pt}
Based on our analysis of operation scoring, in this section, we propose a NAS algorithm called Zero-Cost-PT using zero-cost perturbation and perform ablation studies to find the best set of heuristics for our NAS methodology, including: edge discretization order, number of search and validation iterations, and the choice of the zero-cost metric.

\subsection{Architecture Search with Zero-cost Proxies}
\label{sub:arch-search}

\begin{figure*}[t]
 \centering
 \includegraphics[width=\textwidth, trim=0 0 0 0cm]{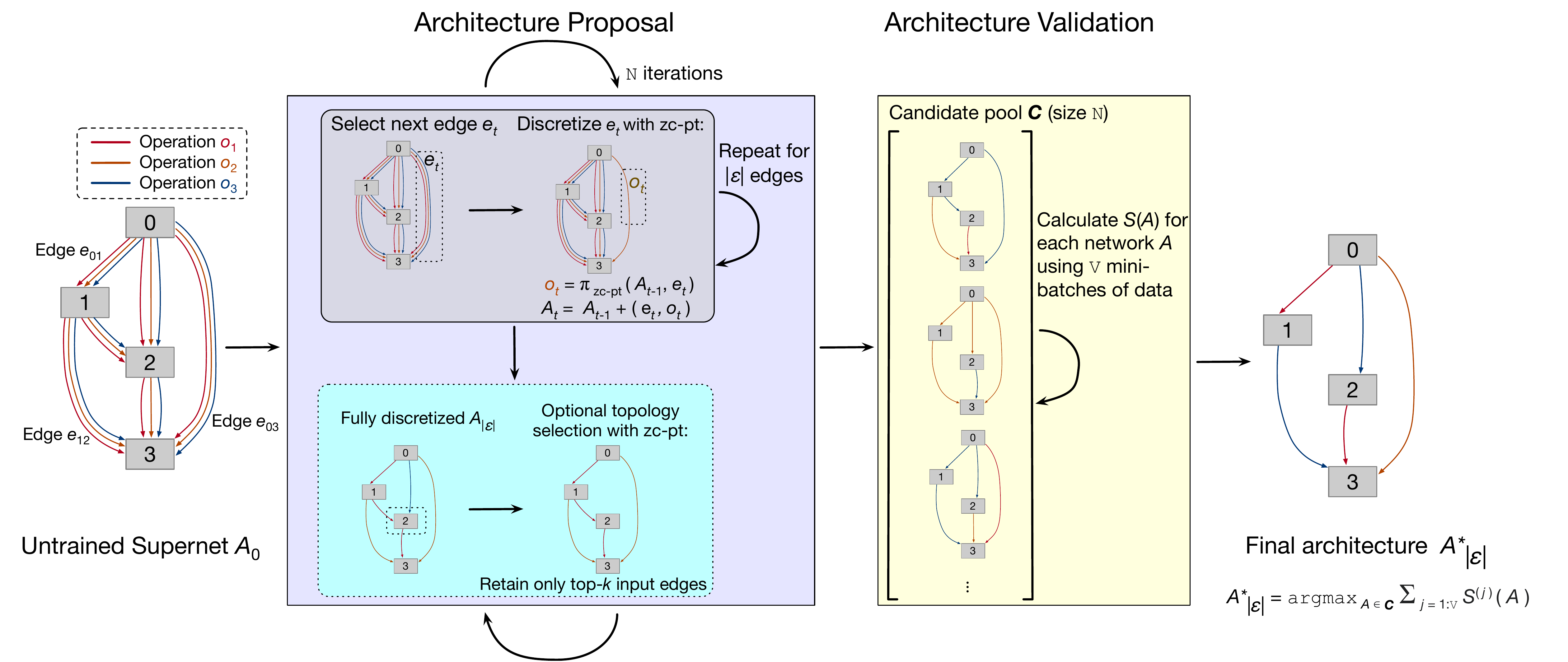}
\caption{Flowchart of the proposed Zero-Cost-PT algorithm.}
 \label{fig:flowchart}
\end{figure*}

Our algorithm contains two stages: \textit{architecture proposal} and \textit{validation}. It begins with an untrained supernet $A_0$ which contains a set of edges $\mathcal{E}$, the number of proposal iterations $\mathtt{N}$, and the number of validation iterations $\mathtt{V}$.
In each proposal iteration $i$, we discretize the supernet $A_0$ based on our proposed zero-cost-based perturbation function $f_{\text{zc-pt}}$ that achieved promising results in the previous section.
After all edges have been discretized, the final architecture is added to the set of candidates and we begin the process again for $i+1$ starting with the original $A_0$.
After $\mathtt{N}$ candidate architectures have been constructed, the validation stage begins.
We score the candidate architectures again using a selected zero-cost metric (the same which is used in $f_{\text{zc-pt}}$), but this time computing their end-to-end score rather than using the perturbation paradigm.
We calculate the zero-cost metric for each subnetwork using $\mathtt{V}$ different minibatches of data.
The final architecture is the one that achieves the best total score during the validation stage.
The full algorithm is outlined as Algorithm~\ref{algo:zero-cost-pt-complete} in Appendix~\ref{sec:app:algo} and the flowchart of our algorithms is in Figure~\ref{fig:flowchart} .
Our algorithm contains four main hyperparameters: $\mathtt{N}$, $\mathtt{V}$, ordering of edges to follow when discretizing, and the zero-cost metric to use ($S$).
In the following we present detailed ablations to decide on the best possible configuration of these.

\begin{table}[t]
\centering
\setlength\tabcolsep{4pt}

\caption {Comparison in test error (\%) with SOTA perturbation-based and zero-cost NAS on NAS-Bench-201 (Best in red, 2nd best in blue. Same for all following tables).}
\footnotesize
\resizebox{\columnwidth}{!}{
    \begin{threeparttable}
    \begin{tabular}{@{}lccc@{}}
    \toprule
    \textbf{Method}    & \textbf{CIFAR-10} &\textbf{CIFAR-100}    &\textbf{ImageNet-16}  \\ \toprule
    \multicolumn{4}{@{}l}{Zero-Cost-PT$_{\texttt{random}}$ with different proxies (Section~\ref{sub:ablation})} \\
    \texttt{tenas}~{\citep{chen2020tenas}}     &70.07$_{\pm 39.87}$	& 83.04$_{\pm 31.93}$	& 90.57$_{\pm 17.21}$\\
    \texttt{fisher}~{\citep{fisher-gaze}}      &10.64$_{\pm 1.27}$	&38.48$_{\pm 1.96}$	&82.85$_{\pm 12.63}$\\ 
    \texttt{grad\_norm}~{\citep{zero-cost}}     &10.55$_{\pm 1.11}$   &38.43$_{\pm 2.10}$	&80.71$_{\pm 12.10}$ \\ 
    \texttt{grasp}~{\citep{grasp}}     &9.81$_{\pm 3.42}$	&36.52$_{\pm 6.33}$	&64.27$_{\pm 8.82}$ \\
    \texttt{snip}~{\citep{snip}}       &8.32$_{\pm2.02}$	&34.00$_{\pm 4.03}$	&65.35$_{\pm 11.04}$\\
    \texttt{zen\_score}\tnote{1}~{\citep{ming_zennas_iccv2021}}     &6.24$_{\pm 0.00}$	&28.89$_{\pm 0.00}$	& 58.56$_{\pm 0.00}$\\
    \texttt{synflow}\tnote{1}~{\citep{synflow}}     &6.24$_{\pm 0.00}$	&{28.89$_{\pm 0.00}$}	&58.56$_{\pm 0.00}$\\

    \texttt{nwot}~{\citep{nwot}} &\best{5.97$_{\pm 0.17}$}	&\best{27.47$_{\pm 0.28}$}   &\best{53.82$_{\pm 0.77}$}	\\
    \bottomrule
    \multicolumn{4}{@{}l}{\textbf{Baselines and SOTA approaches} (Section~\ref{sub:results-201})} \\
    Random	&13.39$_{\pm 13.28}$	&39.17$_{\pm 12.58}$	&66.87$_{\pm 9.66}$  \\
    DARTS	&45.70$_{\pm 0.00}$	&84.39$_{\pm 0.00}$	&83.68$_{\pm 0.00}$ \\ 
    DARTS-PT \tnote{1}    &11.89$_{\pm 0.00}$  &45.72$_{\pm 6.26}$  &69.60$_{\pm 4.40}$ \\
    DARTS-PT (fix $\alpha$) \tnote{1, 2}    &{6.20$_{\pm 0.00}$}  &34.03$_{\pm 2.24}$	&61.36$_{\pm 1.91}$ \\ 
    NASWOT(\texttt{synflow}) \tnote{3}	&6.54$_{\pm 0.62}$	&29.53$_{\pm 2.13}$	&58.22$_{\pm 4.18}$ \\
    NASWOT(\texttt{nwot}) \tnote{3}	&7.04$_{\pm 0.80}$	&29.97$_{\pm 1.16}$	&55.57$_{\pm 2.07}$ \\ 
    TE-NAS    & \secondbest{6.10$_{\pm 0.47}$}  & 28.76$_{\pm 0.56}$  & 57.62$_{\pm 0.46}$    \\
    Zero-Cost-Disc(\texttt{nwot}~{\citep{nwot}}) &6.22$_{\pm 0.84}$	&\secondbest28.18$_{\pm 2.01}$   &\secondbest{55.14$_{\pm 1.77}$}	\\
    \bottomrule
    \end{tabular}
    \begin{tablenotes}
    \scriptsize
        \item[1] Only 1 model was selected across all 4 seeds in both cases.
        \item[2] Results on CIFAR-10 taken from~\citep{darts-pt}. Results on other datasets computed using official code in~\citep{darts-pt} across 4 seeds. 
        \item[3] Using N=1000 for both proxies and averaged over 500 runs as in~\citep{nwot}.
        
    \end{tablenotes}
    \end{threeparttable}
}
\label{tbl:201}
\end{table}

\subsection{Ablation Study}
\label{sub:ablation}

We conduct ablations of the proposed Zero-Cost-PT approach on NAS-Bench-201~\citep{nasbench2}. More results on additional NAS search spaces and benchmarks are reported later in Section~\ref{sec:results} and Appendix~\ref{sec:app:nv_darts}. NAS-Bench-201 constructed a unified cell-based search space, where each architecture has been trained on three different datasets, CIFAR-10, CIFAR-100 and ImageNet-16-120\footnote{\scriptsize{We use the three random seeds available in NAS-Bench-201: 777, 888, 999.}}.
In our experiments, we take a randomly initialised supernet for this search space and apply our Zero-Cost-PT algorithm to search for architectures without any training.
We search with four random seeds (0, 1, 2, 3) and report the average and standard deviation of test errors of the obtained architectures.
All searches are performed on CIFAR-10, and obtained architectures are then additionally evaluated on the other two datasets.



\noindent \textbf{Different Zero-cost Metrics.}
Since our focus is to understand how existing zero-cost metrics can be successfully applied to a large-space NAS, we begin our investigation by analysis how different metrics behave when used in the proposed combination with perturbation-based search. 
It is also worth noting that our formulation and analysis are general and can be extended to new zero-cost proxies that may emerge in the future.
For now we only consider \texttt{random} edge discretization order (Zero-Cost-PT$_{\texttt{random}}$), and more details on edge discretization order will be presented later.
Table~\ref{tbl:201} compares the average test errors of architectures selected by different proxies on NAS-Bench-201.
We see that \texttt{nwot}, \texttt{synflow} and \texttt{zen\_score} perform considerably better across the three datasets than the others, where \texttt{nwot} offers around 0.27\% improvement over \texttt{synflow}. 
On the other hand, we notice \texttt{tenas} fails in this case, as the metric was designed for \textit{pruning} operations rather than \textit{selecting} them (more details are discussed in Appendix~{\ref{sec:app:ablations}}). Other than that, even the naive \texttt{grad\_norm} outperforms the state-of-the-art DARTS-PT on this benchmark. This confirms it is the appropriate combination of zero-cost metrics and perturbation-based NAS paradigms as in Zero-Cost-PT that could become promising proxies to the actual trained accuracy. 
We also observed that the ranking of those metrics are quite stable on the three datasets (descending order in terms of error as in Table~\ref{tbl:201}), indicating that architectures discovered by our Zero-Cost-PT have good transferability.
In particular \texttt{nwot} consistently performs best, reducing test errors on all datasets by a considerable margin.

\begin{figure*}[t]
 \centering
   \caption{(left) Accuracy vs. score of architectures discovered on CIFAR-10 by Zero-Cost-PT with different $\mathtt{N}$. (right) Accuracy distribution of discovered architectures with different $\mathtt{N}$ and $\mathtt{V}$.}
  \includegraphics[width=0.75\textwidth, trim=0 0 0 0cm]{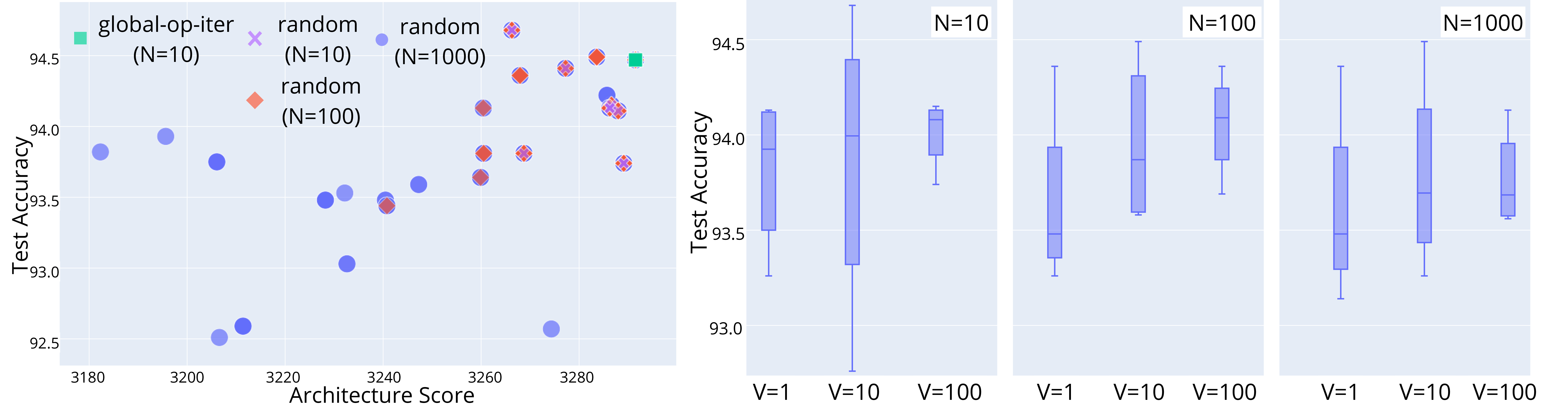}
 \label{fig:n-v-population}
\end{figure*}


\noindent \textbf{Architecture Proposal vs. Validation.}
We then study the impact of different architecture proposal iterations $\mathtt{N}$ and validation iterations $\mathtt{V}$ when Zero-Cost-PT uses \texttt{nwot} metric and \texttt{random} edge discretization order. 
Intuitively, larger $\mathtt{N}$ leads to more architecture candidates being found, while $\mathtt{V}$ indicates the amount of data used to rank the search candidates. 
As shown in Figure~\ref{fig:n-v-population}, we see larger $\mathtt{N}$ does lead to more architectures discovered, but not proportional to the value of $\mathtt{N}$ on NAS-Bench-201 space. 
For $\mathtt{N}$=100 we discover 27.8 distinct architectures on average, but when increased to $\mathtt{N}$=1000 the number only roughly doubles. 
We also see that even with $\mathtt{N}$=10, Zero-Cost-PT$_\texttt{random}$ can already discover top models in the space, demonstrating desirable balance between search quality and efficiency.
On the other hand, as shown in Figure~\ref{fig:n-v-population}, larger $\mathtt{V}$ tends to reduce the performance variance, especially for smaller $\mathtt{N}$. 
This is also expected as more validation iterations could stabilise the ranking of selected architecture candidates, helping Zero-Cost-PT to retain the most promising ones with a manageable overhead of $\mathtt{V}$ mini-batches. 

To further justify our finding on NAS-Bench-201, we performed similar ablations on DARTS-CNN space. We study the impact of different architecture proposal iterations \texttt{N} and validation iterations \texttt{V} when Zero-Cost-PT uses \texttt{random} as the search order and \texttt{nwot} metric. The further experiment details are in Appendix~\ref{sec:app:nv_darts}
 
We first consider an extreme case, setting architecture proposal iteration \texttt{N}=1, where Zero-Cost-PT only proposes one architecture candidate (with \texttt{random} edge discretization order), and with no validation stage performed. And then, in order to maximize the performance of our method, we balance exploration (higher \texttt{N} + random edge order) and exploitation (higher \texttt{V}) in the searching and validation phases respectively. 

Admittedly, the interplay between those two phases is crucial for our method. To further showcase how the validation phase complements the searching phase, we run additional ablations on the DARTS CNN space with \texttt{N}=10 and \texttt{V}=\{1,10,100\}, the results are shown in Table~\ref{tbl:zcpt-ab-overview}. The results are consistent with what is shown in the NAS-Bench-201: higher \texttt{V} produces better results on average but does not affect the best case that much (the best model is still upper-bounded by what was found with \texttt{N}=1).

\begin{table}[t]
\centering
\setlength\tabcolsep{20pt}
\caption {Ablation overview, the performance of Zero-Cost-PT$_{\texttt{random}}$ with \texttt{N}=\{1, 10\}, \texttt{V}=\{0, 1, 10, 100\}, \texttt{nwot} metric on DARTS CNN space.}
\footnotesize
\resizebox{\columnwidth}{!}{
\begin{threeparttable}[]
\begin{tabular}{@{}cccccc@{}}
\toprule
\multicolumn{2}{c}{\multirow{2}{*}{\textbf{N}}} & \multirow{2}{*}{\textbf{V}} & \multicolumn{2}{c}{\textbf{Test Error(\%)}} \\ \cmidrule(l){4-5} 
\multicolumn{2}{c}{}                    &     & Avg. & Best \\ \midrule
\multicolumn{2}{c}{1}                   & 0   & 2.81$_{\pm 0.29}$ & 2.43 \\ \midrule
\multicolumn{2}{c}{\multirow{3}{*}{10}} & 1   & 2.93$_{\pm0.14}$ & 2.65 \\
\multicolumn{2}{c}{}                    & 10  & 2.88$_{\pm 0.14}$ & 2.65 \\
\multicolumn{2}{c}{}                    & 100 & 2.64$_{\pm 0.16}$ & 2.43 \\ \bottomrule
\end{tabular}
\scriptsize
\end{threeparttable}
}
\label{tbl:zcpt-ab-overview}
\end{table}


\noindent \textbf{Edge Discretization Order.}
Finally, we investigate how different edge discretization orders may impact the performance of our Zero-Cost-PT approach, when the best performing \texttt{nwot} metric, $\mathtt{N}$=10 and $\mathtt{V}$=100 are used. We consider the following edge discretization orders:
\begin{itemize}[leftmargin=4mm, topsep=0pt]
    \itemsep0mm 
    \setlength{\itemsep}{0pt}
    \setlength{\parskip}{0pt}
    \setlength{\parsep}{0pt}
    \item \texttt{fixed}: discretizes the edges in a fixed order, where our experiments discretize from the input towards the output;
    \item \texttt{random}: discretizes the edges in a random order; 
    \item \texttt{global-op-iter}: iteratively evaluates $S(A - (e,o))$ for all operations on all edges in $\mathcal{E}$, selects the edge $e$ containing the operation $o^{\ast}$ with globally best score. Discretizes $e$ with $o^{\ast}$, then repeats to decide on the next edge (re-evaluating scores) until all edges have been discretized;
    
    \item \texttt{global-edge-iter}: similar to the above but iteratively selects edge $e$ from $\mathcal{E}$ based on the average score of all operations on each edge; 
    
    \item \texttt{global-op-once}: only evaluates $S(A - (e,o))$ for all operations once to obtain a ranking order of the operations and decide the edge order upfront based on it, then starts following the algorithm as usual, calculating scores of operations at each edge iteratively;
    
    \item \texttt{global-edge-once}: similar to the above but uses the average score of operations on edges to obtain the edge discretization order.
\end{itemize}

\begin{table}[t]
\centering
\setlength\tabcolsep{1pt}
\caption {Comparison with s SOTA differentiable NAS methods on the DARTS CNN search space (CIFAR-10).} 
\footnotesize
\resizebox{\columnwidth}{!}{
    \begin{threeparttable}
    \begin{tabular}{@{} l c c c c @{}}
    \toprule
    \multirow{2}{*}{\textbf{Method}} & \multicolumn{2}{c}{\textbf{Error [\%]}} & \textbf{Params} & \textbf{Cost}   \\
     & Avg. & Best & \textbf{[M]} & \textbf{[GPU-days]}   \\\toprule
    
    DARTS~\cite{darts}                       & 3.00$_{\pm 0.14}$ &-  & 3.3      & 0.4               \\
    SDARTS-RS~\cite{chen2020stabilizing}        & 2.67$_{\pm 0.03}$  &- & 3.4       & 0.4               \\
    SGAS~\cite{li2020sgas}                       & 2.66$_{\pm 0.24}$ &- & 3.7       & 0.25               \\\midrule
    DARTS-PT~\cite{darts-pt}    & \best{2.61$_{\pm 0.08}$}  & \secondbest{2.48} & 3.0       & 0.8              \\
    DARTS-PT$_{+\texttt{none}}$\tnote{1}    & 2.73$_{\pm 0.13}$  & 2.67 & 3.2       & 0.8              \\\midrule
    TE-NAS~\cite{chen2020tenas}    & {2.63$_{\pm 0.064}$}  & - & 3.8       & \secondbest{0.05}              \\\midrule
     max-param-random    & {2.94$_{\pm 0.098}$}  & 2.83 & 5.14       & -              \\
     NASWOT(\texttt{2500}) & {2.99$_{\pm 0.22}$} & 2.66
     & - & 0.018 \\
     NASWOT(\texttt{20000}) & {2.73$_{\pm 0.09}$} & 2.58
     & - & 0.083 \\
     NASWOT(\texttt{50000}) & {2.72$_{\pm 0.09}$} & 2.52
     & - & 0.208 \\
     Zero-Cost-EVO & {2.94$_{\pm 0.14}$} & 2.72
     & - & 0.018 \\
     \midrule
     Zero-Cost-PT$_\texttt{synflow}$                   & 3.88$_{\pm 0.56}$ &3.38 & 5.1      & -                 \\
    Zero-Cost-PT$_\texttt{zen\_score}$                   & 3.06$_{\pm 0.31}$ & 2.68 & 2.9      & -   \\
    Zero-Cost-PT$_\texttt{random}$                   & 2.64$_{\pm 0.16}$ & \best{2.43} & 4.7      & \best{0.018}                 \\
    Zero-Cost-PT$_\texttt{global-op-iter}$                   & \secondbest{2.62$_{\pm 0.09}$} & 2.49  & 4.6      & 0.17                \\\bottomrule
    \end{tabular}
    \begin{tablenotes}
    \scriptsize
        \item[1] 
        Results obtained by re-enabling \texttt{none} operation in DARTS-PT~\citep{darts-pt}.
    \end{tablenotes}
    \end{threeparttable}
}
\label{tbl:darts-c10}

\end{table}


Table.~\ref{tbl:201-search-order-complete} shows the performance of and \# of perturbations required by our Zero-Cost-PT approach when using different edge discretization order, under \texttt{nwot} metric, with $\mathtt{N}=10$ and $\mathtt{V}=100$.

We observe that \texttt{global-op-iter} consistently performs best across all three datasets since it iteratively explores the search space of remaining operations while greedily selecting the current best. 
On the other hand, we see that the performance of \texttt{global-op-once} is inferior since it determines the order of perturbation by assessing the importance of operations once and for all at the beginning, which may not be appropriate as discretization continues.
We observe similar behaviour in \texttt{global-edge-iter} and \texttt{global-edge-once}, both of which use the average importance of operations on edges to decide search order, leading to suboptimal performance. 
It is also worth pointing out that \texttt{fixed} performs relatively well comparing to the other variants, offering comparable performance with \texttt{random}. 
This shows that Zero-Cost-PT is generally robust to the edge discretization order.
In the following experiments, we use Zero-Cost-PT with \texttt{random} order with a moderate setting in architecture proposal iterations ($\mathtt{N}$=10) to balance exploration and exploitation during the search, while maintaining efficiency.

\begin{table}[t]
\centering
\caption {Test error (\%) of Zero-Cost-PT when using different search orders on NAS-Bench-201.}
\footnotesize
\resizebox{\columnwidth}{!}{
\begin{threeparttable}
\begin{tabular}{@{}lcccc@{}}
\toprule
\textbf{Search Order\tnote{1}}    & \textbf{\# of Perturbations\tnote{2}} & \textbf{C10} &\textbf{C100}    &\textbf{ImageNet-16}  \\ \toprule
\texttt{fixed} &\best{$|\mathcal{O}||\mathcal{E}|$}	    &5.98$_{\pm 0.50}$	&27.60$_{\pm 1.63}$   &54.23$_{\pm 0.93}$\\ 
\texttt{global-op-iter}	&$\frac{1}{2}|\mathcal{O}||\mathcal{E}|(|\mathcal{E}| + 1)$	&\best{5.69$_{\pm 0.19}$}	&\best{26.80$_{\pm 0.51}$}	&\best{53.64$_{\pm0.40}$}\\ 
\texttt{global-op-once}	&\secondbest{$2|\mathcal{O}||\mathcal{E}| - |\mathcal{O}|$}	&6.30$_{\pm 0.57}$	&28.96$_{\pm 1.66}$	&55.04$_{\pm 1.47}$\\ 
\texttt{global-edge-iter}	&$\frac{1}{2}|\mathcal{O}||\mathcal{E}|(|\mathcal{E}| + 1)$	&6.23$_{\pm 0.45}$	&28.42$_{\pm 0.59}$	&54.39$_{\pm 0.47}$\\ 
\texttt{global-edge-once}	&\secondbest{$2|\mathcal{O}||\mathcal{E}| - |\mathcal{O}|$}	&6.30$_{\pm 0.57}$	&28.96$_{\pm 1.66}$	&55.04$_{\pm 1.47}$\\ 
\texttt{random}	&\best{$|\mathcal{O}||\mathcal{E}|$}	&\secondbest{5.97$_{\pm 0.17}$}	&\secondbest{27.47$_{\pm 0.28}$}   &\secondbest{53.82$_{\pm 0.77}$}\\ \bottomrule 
\end{tabular}
\scriptsize
\begin{tablenotes}
 \item[1] All methods use \texttt{nwot} metric, $\mathtt{N}$=10 search iterations and $\mathtt{V}$=100 validation iteration.
 \item[2] Number of perturbations per search iteration.
\end{tablenotes}
\end{threeparttable}
}
\label{tbl:201-search-order-complete}
\end{table}

%% file: 50-results.tex
\section{Results}
\label{sec:results}

\begin{table}[t]
\centering
\setlength\tabcolsep{1pt}
\caption {Comparison with SOTA differentiable NAS methods on the DARTS CNN search space (ImageNet).}
\footnotesize
\resizebox{\columnwidth}{!}{
    \begin{threeparttable}
    \begin{tabular}{@{}lcccc@{}}
    \toprule
    \multirow{2}{*}{\textbf{Method}} & \multicolumn{2}{c}{\textbf{Error [\%]}} & \textbf{Params} & \textbf{Cost}   \\
     & Top-1 & Top-5 & \textbf{[M]} & \textbf{[GPU-days]}   \\\toprule
     DARTS~\cite{darts}       & 26.7        & 8.7         & 4.7 &  0.4  \\
     SDARTS-RS~\cite{chen2020stabilizing}  & 25.6 & 8.2 & - & 0.4  \\
     DARTS-PT~\cite{darts-pt} & 25.5        & 8.0         & 4.6 &  0.8  \\
     PC-DARTS~\cite{pcdarts}  & 25.1        & 7.8         & 5.3 &  0.1  \\
     SGAS~\cite{li2020sgas}  & \best{24.1} & 7.3 & 5.4 & 0.25  \\\midrule
     TE-NAS(C10)~\cite{chen2020tenas}      & 26.2        & 8.3          & 6.3 & \secondbest{0.05} \\
     TE-NAS~\cite{chen2020tenas}      & 24.5        & 7.5          & 5.4 & 0.17 \\\midrule
     Zero-Cost-PT\tnote{1} (best)      & \secondbest{24.4}        & 7.5           & 6.3 & \best{0.018} \\
     Zero-Cost-PT\tnote{1} (4 seeds)   & 24.6$_{\pm 0.13}$ & 7.6$_{\pm 0.09}$ & 6.3 & \best{0.018} \\
    \bottomrule
    \end{tabular}
    \begin{tablenotes}
    \scriptsize
    \item[1] We use the same training pipeline from DARTS~\citep{darts}.
    \end{tablenotes}
    \end{threeparttable}
}
\label{tbl:darts-imagenet}
\end{table}

In this section, we perform extensive empirical comparisons of Zero-Cost-PT with the state-of-the-art differentiable and zero-cost NAS algorithms on a number of search spaces. Due to space limit, in the following, we present results on NAS-Bench-201~\citep{nasbench2}, DARTS CNN space~\citep{darts} and the practical large search space MobileNet-like space. Results on NAS-Bench-1shot1~\citep{nasbench-1shot1}, NAS-Bench-Macro~\citep{nasbench-macro} and the four DARTS subspaces S1-S4~\citep{arber2020understanding}, together with detailed experimental settings and more baselines are in Appendix~\ref{sec:app:exp-details} \ref{sec:app:nb1s1}, \ref{sec:app:nbmacro}, \ref{sec:app:mobilenet}, .

\subsection{Tabular NAS Benchmarks}
\label{sub:results-201}

Table~\ref{tbl:201} shows the average test error (\%) of the competing approaches and our Zero-Cost-PT on the three datasets in NAS-Bench-201.
Here we include the naive random search and original DARTS as baselines, and compare our approach with the recent zero-cost NAS algorithm NASWOT~\citep{nwot}, TE-NAS~\citep{chen2020tenas}, as well as the perturbation-based NAS approaches DARTS-PT and DARTS-PT (fix $\alpha$)~\citep{darts-pt}.
As in all competing approaches, we perform a search on CIFAR-10 and evaluate the final model on all three datasets.
We see that on all datasets, our Zero-Cost-PT (with \texttt{nwot}) consistently offers superior performance, especially on CIFAR-100 and ImageNet-16.
On the other hand, the best existing perturbation-based algorithm, DARTS-PT (fix $\alpha$), fails on those two datasets, producing suboptimal results with small improvements compared to random search, suggesting that architectures discovered by DARTS-PT might not transfer well to other datasets.
TE-NAS is second best on CIFAR but as we show in the later section, performance deteriorates on larger datasets like ImageNet.

We compare the performance of Zero-Cost-DISC and our proposed Zero-Cost-PT on NAS-Bench-201~\citep{nasbench2}, as shown in Table~\ref{tbl:201}. We see that discretization (Zero-Cost-DISC) results in inferior performance compared to the proposed perturbation-based approach (Zero-Cost-PT) on all datasets, confirming our previous analysis on their correlations with the oracle metric.


\subsection{DARTS CNN Search Space}
\label{sub:results-darts}

We use the same settings as in DARTS-PT~\citep{darts-pt}, but instead of pre-training the supernet and fine-tuning it after each perturbation, we take an untrained supernet and directly perform our algorithm as in Section~\ref{sub:arch-search}.
Additional details,  baselines, ablations and discovered architectures can be found in Appendix~\ref{sec:app:nv_darts},  \ref{sec:app:exp-details}, and \ref{sec:app:models}.

\noindent\textbf{Results on CIFAR-10.} As shown in Table~\ref{tbl:darts-c10} the proposed Zero-Cost-PT approaches can achieve a much better average test error than the DARTS and are comparable to its newer variants SDARTS-RS~\citep{chen2020stabilizing} and SGAS~\citep{li2020sgas} at a much lower searching cost (especially when using \texttt{random} edge ordering).
There is a significant search cost reduction compared to DARTS-PT.
While DARTS-PT needs to perform retraining between iterations, Zero-Cost-PT only evaluates the score of the perturbed supernet with zero-cost proxies ($S_{\texttt{nwot}}$), requiring less than a minibatch of data.
Note that here the cost of Zero-Cost-PT reported in Table~\ref{tbl:darts-c10} is for $\mathtt{N}$=10 ( \texttt{random} edge discretization order), and thus a single proposal iteration only takes about a few minutes to run.
The $\texttt{global-op-iter}$ variant offers better performance with lower variance compared to $\texttt{random}$ but incurs slightly heavier computation.

\subsubsection{Different Zero-cost Searching Approaches.}
We additionally compare our Zero-Cost-PT to several alternative ways of performing zero-cost NAS, to further show its efficiency in utilising zero-cost proxies and assert its efficiency as a searching methodology. 

We start with the simplest baseline of maximizing number of parameters -- this is based on the observation that our method tends to select slightly larger models than some of the baselines in Table~\ref{tbl:darts-c10} and Table~\ref{tbl:darts-imagenet}.
We include details in the Appendix~\ref{sec:app:nv_darts} and summarize our finding here.
Overall, the test error (\%) of this baseline is 2.93$_{\pm0.23}$ (avg.) and 2.78 (min), vs. our  2.64$_{\pm0.16}$ (avg.) and 2.43 (min).
This confirms that simply selecting models with maximum FLOPs/Params is not an appropriate searching methodology in general, and that our methods performs more meanigngful architecture selection than simply maximising model size.


In Section~\ref{sub:results-201}, we compared our method to sampling-based zero-cost NAS in Table~\ref{tbl:201} (see NASWOT lines). Our results are empirically better on all three datasets. Additionally, our method computes the operation score per edge in a supernet, whereas the sampling-based approach computes the end-to-end network score. The relationship between the number of subnetworks and the number of operations is exponential. Therefore, we anticipate having to sample exponentially many networks in sample-based NASWOT~\citep{nwot} compared to our proposed Zero-Cost-PT. 

In order to extend the comparison between zero-cost NAS (NASWOT) and our Zero-Cost-PT to the DARTS CNN search space, we have conducted further experiments similar to NASWOT on NAS-Bench-201, the details of these experiments can be found in Appendix~\ref{sec:app:nv_darts} and the results are presented in Table~\ref{tbl:darts-c10}.
In summary, for a similar time budget to ours (25 min), the average performance of the baseline is actually closer to the random search (3.29$_{\pm 0.15}$)~\citep{darts} than to our method, with significant variance.

In addition to the random sampling baseline presented above, here we further extend our study by performing an evolution-based search for a model maximizing the \texttt{nwot} metric (instead of sampling randomly as above) which is then trained.
We denote this baseline as Zero-Cost-EVO.
We allow a similar search budget (2500 sample size, $\sim$25min on a single 2080ti GPU), and follow the same settings as in our experiments (searching with 4 random seeds, each of the discovered models is trained with 4 random seeds).
The results are shown in Table~\ref{tbl:darts-c10}.
We can see that when given a similar search budget, the evolution-based search performs significantly worse than our Zero-Cost-PT (avg. of 2.94 vs. 2.64), confirming the efficacy of the proposed NAS algorithm.

\noindent\textbf{Results on ImageNet.} 
Table~\ref{tbl:darts-imagenet} shows the ImageNet classification accuracy for architectures searched on CIFAR-10.
Our Zero-Cost-PT$_{\texttt{random}}$ algorithm is able to find architectures with a comparable accuracy much faster than previous work, further reinforcing its efficacy in this setting.
While TE-NAS results on CIFAR-10 were very close to Zero-Cost-PT, a much larger difference is observed on ImageNet with an accuracy drop of 1.8 pp and a search time that is $\sim$2.5$\times$ slower than Zero-Cost-PT. 


\subsection{MobileNet-like Search Space}
\label{sub:results-mobilenet}
It is well known that most of the existing NAS algorithms designed for MobileNet-like perform constrained NAS.
However, our method has not been designed for such a context.
To the best of our knowledge, the necessity to consider both scores of operations and their potential contribution to the sum of \#FLOPS/Params of the final model would result in a potentially NP-hard problem.
Therefore, we do not enforce such constraints at this point, as it is less relevant to the proposed approach. 

\noindent\textbf{Results on ImageNet.} Table~\ref{tbl:app:mobilnet} shows the performance (error \%) of the architectures discovered by the proposed Zero-Cost-PT algorithm on ImageNet, using a MobileNet-like search space from ProxylessNAS \cite{proxylessnas}, we, therefore, compare only to results using the same setting.
We see that compared to the existing train-based approaches, our approach allows for finding even better models, but also larger ones, much faster (at least $190\times$ speed up).
Please note, that because our method performs unconstrained NAS, unlike existing baselines, the results should not be interpreted as being objectively better.
Instead, we simply use them as reference points to put our results in perspective -- the goal was to show that our method works well in this type of search space and the results support this claim; as expected, more accurate models, but with a larger footprint, can be found faster compared to the constrained baselines.

\begin{table}[t]
\centering
\caption {Comparison on MobileNet search space (ImageNet)}
\footnotesize
\resizebox{\columnwidth}{!}{
    \begin{threeparttable}[]
    \begin{tabular}{@{}lccccc@{}}
    \toprule
    \multirow{2}{*}{\textbf{Architecture}} & \multicolumn{2}{c}{\textbf{Error [\%]}} & \textbf{Params} & \textbf{Cost}   \\
     & Top-1. & Top-5 & \textbf{[M]} & \textbf{[GPU-days]}   \\\toprule
    MobileNet-V3(1.0)\cite{mobilev3}  & \secondbest{24.8}  & -   & 5.3    & 288                \\
    GreedyNAS \cite{greedynas}         & 25.1  & -   & 3.8    & \secondbest{7.6}               \\
    SPOS \cite{spos}                 & 25.3  & -   & -    & 12.4              \\
    ProxylessNAS (GPU) \cite{proxylessnas} & 24.9  & 7.5   & 7.1    & 8.3                \\ \midrule
    Zero-Cost-PT(best)                  & \best{23.6}  & 6.8   & 8.0   & \best{0.041}         \\
    Zero-Cost-PT(avg)                  & 23.8$_{\pm 0.08}$    & 6.93$_{\pm 0.09}$   & 8.1   & 0.041         \\
    \bottomrule
    \end{tabular}
    \end{threeparttable}
}
\label{tbl:app:mobilnet}
\end{table}

%% file: 60-conclusion.tex
\section{Conclusion}
\label{sec:conclusion}
In this paper, we formalized the implicit operation scoring proxies that are present within differentiable NAS algorithms to both analyze existing methods and propose new ones.
We showed that lightweight operation scoring methods based on zero-cost proxies empirically outperform existing operation scoring functions. We also found that perturbation is more effective than discretization when scoring an operation, leading to our lightweight NAS algorithm, Zero-Cost-PT.
Our approach outperforms the best available differentiable architecture search in terms of searching time and accuracy even in very large search spaces.

\newpage

%% file: 90_appendix.tex
\beginsupplement

\subsection{Additional Details on Notations}
\label{sec:app:notation}

All sets are denoted with stylised capital letters using latex's \texttt{mathcal} font.
Letters denoting elements of different sets use the same letters as the sets, e.g. $e\in\mathcal{E}$, $A\in\mathcal{A}$, etc.
For any element, we use subscript for indexing iterations of the discretization process -- e.g., $A_0$ is a network architecture at the beginning of iteration 0, $e_2$ is an edge that is being investigated in iteration 2, etc.
To identify different elements in any other context we use superscript, e.g. $e^{(1)}$ might denote the first edge in a supernet, which might be different from $e_1$ which denotes the edge that is first going to be dicretized following a relevant discretization order.

\begin{figure}[h]
 \centering
 \includegraphics[width=\textwidth]{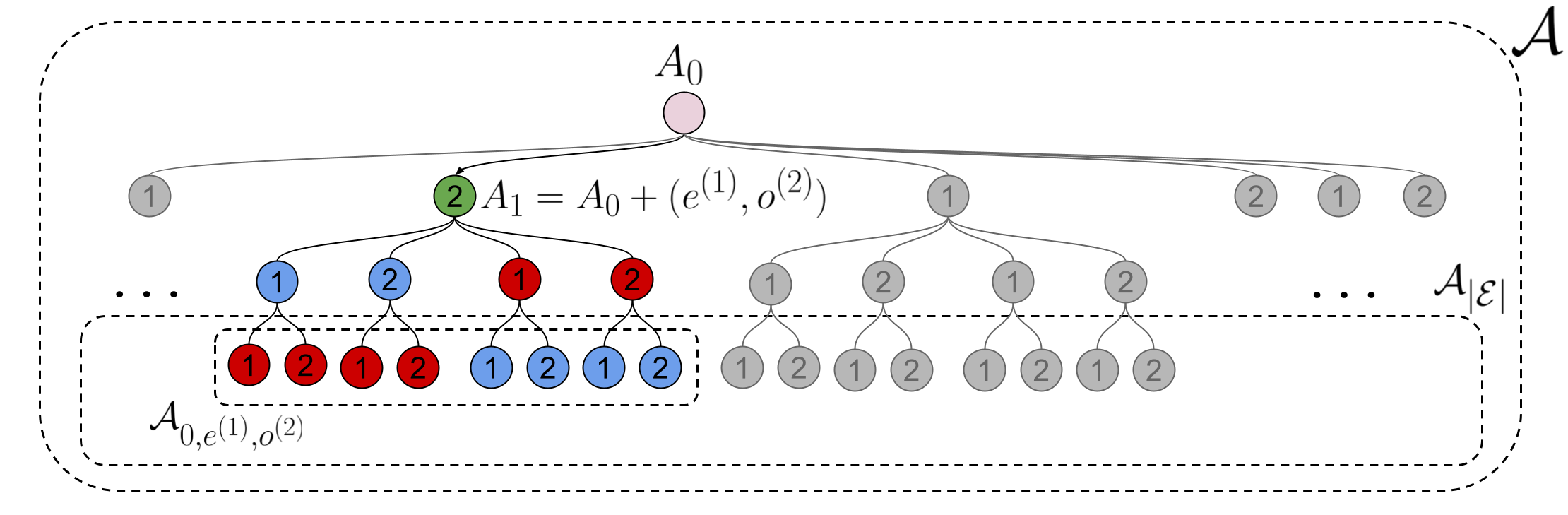}
 \caption{Visualisation of the searching tree related to a hypothetical supernetwork with 3 nodes (represented by different colors) and 2 candidate operations (represented by different numbers). The root node of the tree represents a full supernetwork. Each descendant node represents a non-unique network architecture obtained by discretizing an edge identified by the color of the node, with an operations encoded by a number in the node, given an architecture represented by a parent node. $\mathcal{A}$ is the set of all possible architectures, including the supernet $A_0$, partially-discretized networks like $A_1$ and fully-discretized networks. The set of all fully-discretized networks is additionally denoted as $\mathcal{A}_{|\mathcal{E}|}$, and the set of all fully-discretized networks that are achievable after operation $o^{(2)}$ is assigned to edge $e^{(1)}$, given $A_0$, is labeled as $\mathcal{A}_{0,e^{(1)},o^{(2)}}$. The arrow from $A_0$ to $A_1$ represents a possible first step during the discretization process which closes all grayed-out configurations.
 While performing NAS, at each tree level we need to decide which edge to follow; this is achieved by assigning scores to all/some of the edges that we are willing to consider in each iteration using different scoring functions which are the main focus of our analysis in Section~\ref{sec:scoring}.}
\vspace{-0.5cm}
\label{fig:app:search_tree}
\end{figure}

To better understand meaning of each used symbol, consider a hypothetical supernet with 3 edges -- $e^{(1)}, e^{(2)}, e^{(3)}$ -- repented by different colors (green, blue and red, respectively) and 2 candidate operations -- $o^{(1)}, o^{(2)}$  -- represented by different numbers (1 and 2).
Figure~\ref{fig:app:search_tree} visualises the entire space related to the decision process that is happening in order to perform NAS in this setting, including the first discretization step $A_1=A_0 + (e_0,o_0)$, where $e_0=e^{(1)}, o_0=o^{(2)}$, and the related set of all achievable fully-discretized models $\mathcal{A}_{0,e_0,o_0}$ -- a concept central to our definition of the optimal scoring function.

\subsection{More on Zero-Cost Proxies}
\label{sec:app:zc-proxies}
This section provides more details on the zero-cost proxies considered in our analysis (Section 3), ablation study (Section 4) and experiments (Section 5).

\noindent \texttt{grad\_norm}~\citep{zero-cost}, which is a simple proxy that sums the Euclidean norm of the gradients after a single minibatch of training data.

\noindent \texttt{snip}~\citep{snip}, which approximates the change in loss when a specific parameter is removed:

\begin{equation}
\mathtt{snip}: \mathcal{S}_p(\theta) = \abs{\frac{\partial \mathcal{L}}{\partial \theta} \odot \theta},\,\,\,\,\,\,\,\,\
\end{equation}

\noindent \texttt{grasp}, which approximates the change in gradient norm (instead of loss) when a parameter is pruned in their \texttt{grasp} objective:

\begin{equation}
\mathtt{grasp}: \mathcal{S}_p(\theta) = -(H\frac{\partial \mathcal{L}}{\partial \theta}) \odot \theta,\,\,\,\,\,\,\,\,\
\end{equation}

\noindent \texttt{synflow}, which generalized the above two so-called \textit{synaptic saliency} scores and proposed a modified version which avoids layer collapse when performing parameter pruning: 

\begin{equation}
\mathtt{synflow}: \mathcal{S}_p(\theta) = \frac{\partial \mathcal{L}}{\partial \theta} \odot \theta
\end{equation}

where $\mathcal{L}$ is the loss function of a neural network with parameters $\theta$, $H$ is the Hessian\footnote{The full Hessian does not need to be explicitly constructed as explained by \cite{pearlmutter}.}, $\mathcal{S}_p$ is the per-parameter saliency and $\odot$ is the Hadamard product.

\noindent \texttt{nwot}~\cite{nwot}, which scores an untrained network by examining the  the overlap of activations between data points: 

\begin{equation}
s = \log |\mathbf{K}_H|,  \;\;\; \mathbf{K}_H = \begin{pmatrix}
N_A - d_H(\mathbf{c}_1,\mathbf{c}_1) & \cdots & N_A - d_H(\mathbf{c}_1,\mathbf{c}_N)\\
\vdots & \ddots & \vdots \\
N_A - d_H(\mathbf{c}_N,\mathbf{c}_1) & \cdots & N_A - d_H(\mathbf{c}_N,\mathbf{c}_N)
\end{pmatrix}
\end{equation}


where $N_A$ is the number of rectified linear unites, and $d_H(\mathbf{c}_1,\mathbf{c}_1)$ is the Hamming distance between two binary codes, induced by the untrained network at two inputs.

\noindent \texttt{tenas}~\cite{chen2020tenas}, which uses the spectrum of the neural tangent kernel (NTK) $\kappa_\mathcal{N}$ and the number of linear regions in the input space  $\hat{R}_\mathcal{N}$ to rank architectures, which are defined as

\begin{equation}
\kappa_\mathcal{N} = \frac{\lambda_0}{\lambda_m}, \;\;\; \hat{R}_\mathcal{N} \simeq \mathop{{}\mathbb{E}}_{\theta} R_{\mathcal{N}, \theta}
\end{equation}

where $\lambda_0$, $\lambda_m$ are the eigenvalues of the NTK. 

\noindent \texttt{zen\_score}~\cite{ming_zennas_iccv2021}, which is a re-scaled version of $\Phi$-score (by the product of BN layers' variance statistics), measuring the expressivity of a neural network by its expected Gaussian complexity:

\begin{equation}
\Phi (f) = \log \mathop{{}\mathbb{E}}_{\boldsymbol{x}, \theta} \|\nabla_{\boldsymbol{x}} f(\boldsymbol{x}|\boldsymbol{\theta})\|_{F}
\end{equation}

where $f(\cdot)$ is a vanilla neural network.

\subsection{More on Operation Scoring}
\label{sec:app:scores}

This section provides more experimental details and examples of our analysis of the operation scoring functions introduced in Section~\ref{sec:scoring}.

\subsubsection{Detailed Scoring Methodology}

As discussed in Section~\ref{sub:op-eval-corr}, our analysis on the on the initial operation scoring aims to investigate how well an operation scoring method can perform when making the first discrietization decision, with respect to the perfect search (the best-acc approach). Here ``the first discrietization decision'' is made at the first iteration (iteration 0) of a progressive operation selection algorithm, and in our experiment we compute the score for all operations on an edge and later average across all edges to account for random selection of the first edge. Concretely, we compute the score per operation across all edges, then compute the Spearman rank correlation for operations on each edge. After that, we average the rank correlation coefficient over all edges. 

\begin{figure}[t]
 \centering
 \includegraphics[width=0.8\textwidth]{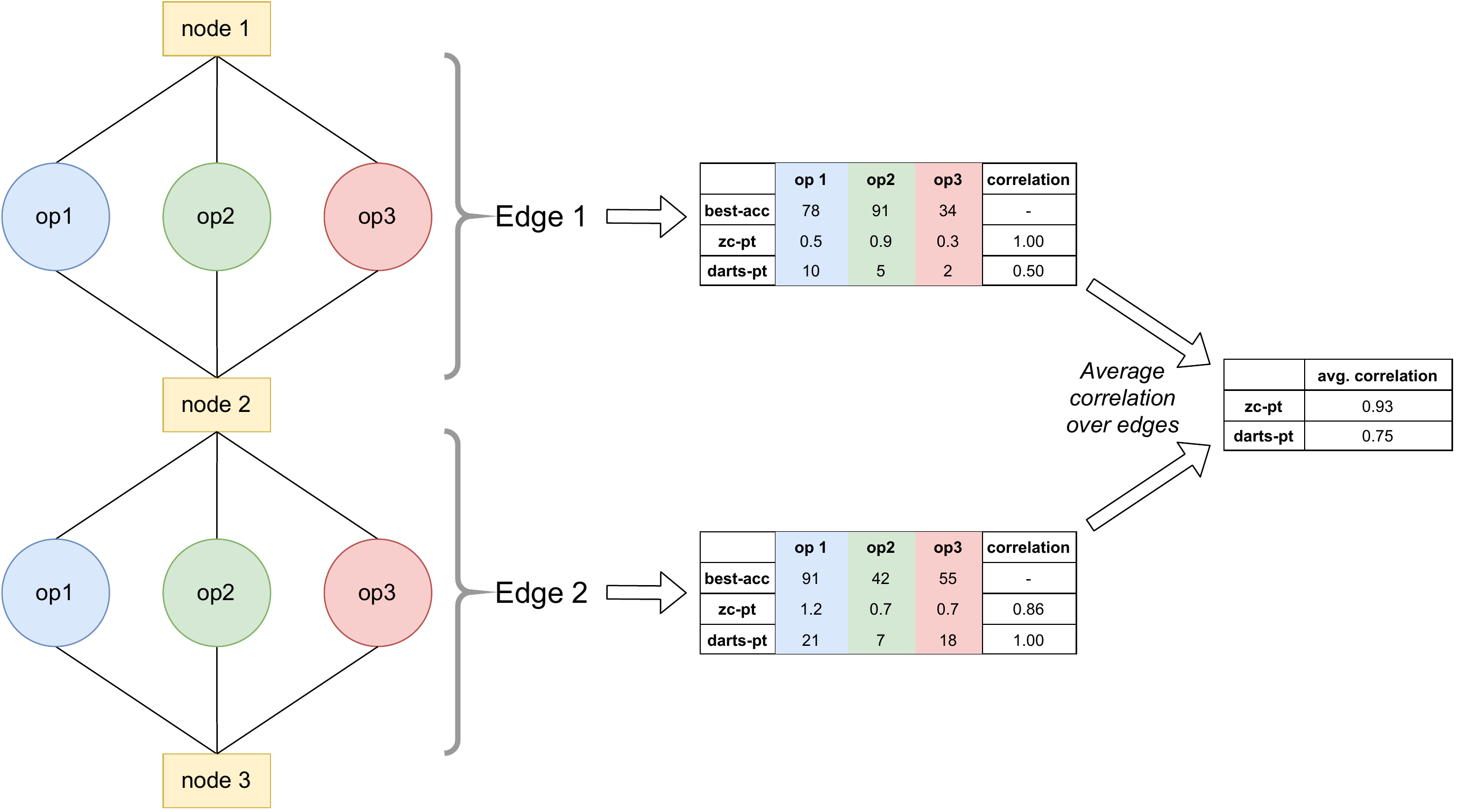}
 \caption{An example showing how correlations of different proxies are computed in our analysis on initial operation scoring in Section~\ref{sub:op-eval-corr}}
\vspace{-0.5cm}
\label{fig:app:op_scoring}
\end{figure}

Consider the example shown in Figure~\ref{fig:app:op_scoring}. Suppose we have a supernet with just two edges as in Figure~\ref{fig:app:op_scoring}. In this case, an operation scoring function should pick just one operation per edge. For a given operation scoring function, we compute the scores for each operation on each edge. Then for each edge, we compute the rank correlation of the oracle scores (best-acc or avg-acc) against the scores from the other operation scoring function (e.g. zc-pt or darts-pt). We then average their correlation coefficient across all edges in a supernet to get an average correlation for each operation scoring function. The resulting average rank correlation is indicative of how well a given operation scoring function (starting with a random edge) would do when making the first discretization decision, relative to the oracle search.

\subsubsection{Experimental Details}
\label{appendix:experimental-details}
Here, we provide some additional experimental details for the data presented in Section~\ref{sec:scoring}.
The following list describes how we compute each operation score.
\begin{itemize} [leftmargin=8mm, topsep=0pt]
    \item \textbf{best-acc}: To get the score for an operation $o$ on a specific edge $e$, we find the maximum test accuracy of all NAS-Bench-201 architectures with $(o,e)$.
    \item \textbf{avg-acc}: Same as best-acc but we \textit{average} all NAS-Bench-201 architecture test accuracies instead of finding the maximum.
    \item \textbf{disc-acc}: We discretize one edge $e$ by selecting an operation $o$, then we train for 5 epochs\footnote{DARTS-PT defines discretization accuracy as the accuracy after \textit{convergence}. We elected to only train for 5 epochs to make our experiments feasible but we are now investigating whether longer training will affect our results.} and record the supernet accuracy -- this is used as the score for $(o,e)$.
    \item \textbf{darts-pt}: We perturb one edge with one operation $A-(e,o)$ and record the validation accuracy. For perturbation-based scoring functions, we multiply the score by $-1$ before computing correlations.
    \item \textbf{disc-zc}: We discretize one edge $e$ by selecting an operation $o$ and then compute the zero-cost metric.
    \item \textbf{zc-pt}: We perturb one edge with one operation $A-(e,o)$ and compute the zero-cost metric. For perturbation-based scoring functions, we multiply the score by $-1$ before computing correlations.
    \item \textbf{darts}: We record the value of the architecture parameters $\alpha$ after 60 epochs of training the supernet.
    \item \textbf{tenas}: We perturb one edge with one operation $A-(e,o)$ and compute the $\kappa_\mathcal{N}$  and the $\hat{R}_\mathcal{N}$ (number of linear regions). Then we rank $\kappa_\mathcal{N}$ ascendingly and descendingly rank the $\hat{R}_\mathcal{N}$. At last, we add those two ranks together to get the final ranks of the operations. We multiply the score by $-1$ before computing correlations.
\end{itemize}

\subsubsection{Detailed Operation Scores}

Table~\ref{tbl:raw-op-scores} (at the end of this Appendix) shows all operation scores at iteration 0.
This data was used to compute Spearman-$\rho$ in Figure~\ref{op}.
Note that we compute Spearman-$\rho$ per edge and average over all edges -- this summarizes how well each score tracks our ``oracle'' best-acc score.

\subsubsection{Additional Analysis on darts-pt and disc-acc}
\label{sec:app:additional-disc-acc}
Darts-pt makes an implicit assumption that changes of accuracy (either due to perturbation or due to discretization followed by fine-tuning) observed in the context of a supernet $A_t$ are representative of changes of accuracy in the context of network $A_{|\mathcal{E}|}$. However, especially when $t < |\mathcal{E}|$, these two networks are in fact very different and there is no reason to take this assumption for granted - even though $A_{|\mathcal{E}|}$ can be derived from $A_t$ (in general, many networks can be obtained from each other with relatively simple transformations but that does not mean that the same architectural transformations will be beneficial for them).

To be even more specific, let us consider disc-acc and darts-pt separately, as they perform scoring slightly differently and hence suffer from slightly different things. (In the case of darts-pt we will directly rely on the analysis presented in Proposition 1 in the original paper so please refer there for background and details.)

darts-pt begins by training a full supernet - our claim is that this results in an undesired bias as the changes in accuracy of a supernet might not be relevant in the context of a subnet. To justify this claim we begin by quoting implication of Proposition 1 from the DARTS-PT paper~\cite{darts-pt} saying that ``in a well-optimized supernet, $x_c$ will naturally be closer to $m^*$ than $o_e(x_e)$'', where $x_c$ is output of a skip-connection component of a mixup operation in a supernet, $o_e$ is convolutional operation in the same mixup operation and $m^*$ is the optimal feature map for the mixup operation. DARTS-PT uses this to show that because of that magnitude of $\alpha_{skip}$ will be larger and hence selection based on magnitude is biased. However, paradoxically, the solution they proposed suffers from a very similar issue (although maybe in a slightly more indirect way).

Specifically, if we assume that  is the optimal output of the mixup operation and the skip connection component is closer to  than any convolutional component then it is also expected that perturbing an edge by removing the skip connection will result in the largest deviation from . Consequently, we can expect that the larger the deviation from the optimal output the larger the drop in accuracy, resulting in selection biased towards selecting skip connections. This behaviour is exactly what we observe on NAS-Bench-201 where in the first iteration of NAS darts-pt tends to assign significantly larger scores to skip connections than any other operation (see Table~\ref{tbl:raw-op-scores}). Please note that this analysis does not consider later iterations, so it is still possible that DARTS-PT as a whole works well due to improved performance in later iterations, but at least in early iterations it should be biased towards selecting skip connections (and empirically we can see that the results are actually quite similar to standard darts).

\begin{figure}[h]
 \centering
 \includegraphics[width=.5\textwidth]{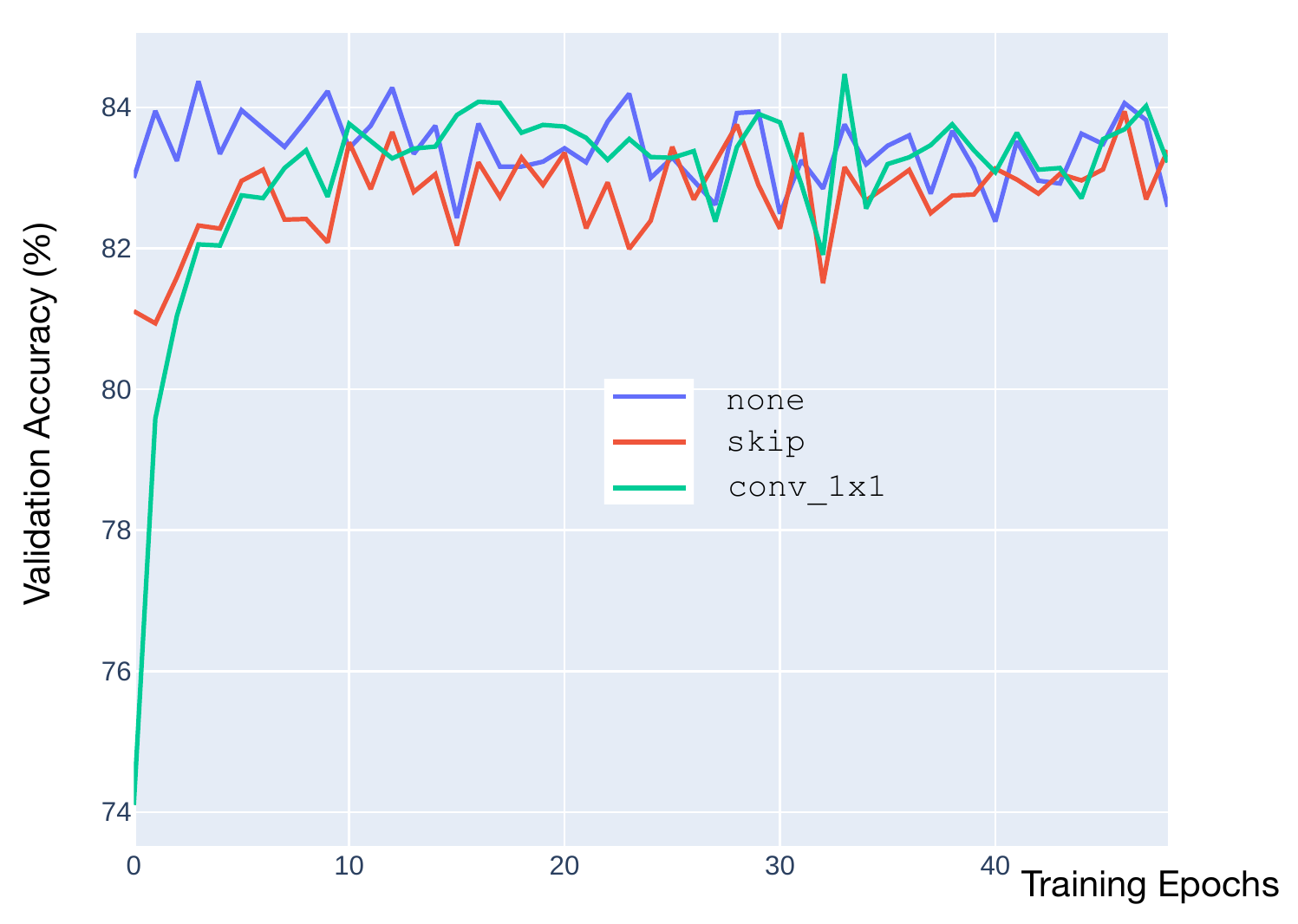}
 \caption{Validation accuracy of three partially discretized supernets at iteration 0 trained on NAS-Bench-201, with none, skip and conv\_1x1 operation selected on edge 1 respectively.  }
\label{fig:disc-acc}
\end{figure}

On the other hand, disc-acc considers performance of a supernet after an edge is discretized, thus (locally) avoiding usage of a mixup operation. However, we also observe similar biased behaviour at the early stage of training, i.e., assigning large scores to skip or none operations. We conducted additional experiments on NAS-Bench-201 (iteration 0), following the same setting as DARTS-PT. We first train the full supernet, and then discretize skip, none and conv\_1x1 on edge 1 respectively, while keeping other edges unchanged (with mixed operations). We train these three partially discretized supernets for 50 epochs and record validation accuracy. Fig.~\ref{fig:disc-acc} plots the validation accuracy of the three resulting partially discretized supernets. 

As we can see, at the early stage of training, the supernets with none or skip operations selected have much higher accuracy than that of conv\_1x1, this is due to the well-known difficulties in training full supernets (see e.g. TE-NAS~\cite{chen2020tenas} 3.2.1) Note that although DARTS-PT defines discretization accuracy as the accuracy after convergence, its algorithm only trains for 5 epochs due to the cost (we mention this in Appendix A.3.2). In that case, disc-acc would very likely make wrong decisions as to which operation to choose. On the other hand, as training continues, the ranking of these three operations computed by disc-acc (discretization accuracy) is still unstable, which may again lead to suboptimal decisions. As mentioned above, such bias towards skip or none in early iterations would result in inferior performance, e.g. weaker correlation with the oracle, as we demonstrated in the paper.

\subsection{Additional Analysis on zc-pt vs. disc-zc}
\label{sec:app:additional-op-scoring}

As shown in Section~\ref{sub:op-eval-corr}, the proposed zc-pt operation scoring function, when using either \texttt{synflow} or \texttt{nwot} metrics, demonstrates strong correlations with the oracle best-acc and avg-acc metrics. However, we also observe that disc-zc, in particular when \texttt{nwot} metric is used, is only weakly or inversely correlated with the oracle. This suggests that \textit{perturbation} is more robust than \textit{discretization}, when combined with the zero-cost metric \texttt{nwot}. In this section, we attempt to explain this observation with further analysis.

\begin{figure*}[t]
 \centering
 \begin{subfigure}{0.43\textwidth}
 \includegraphics[width=\textwidth, trim=0 0 0 1cm]{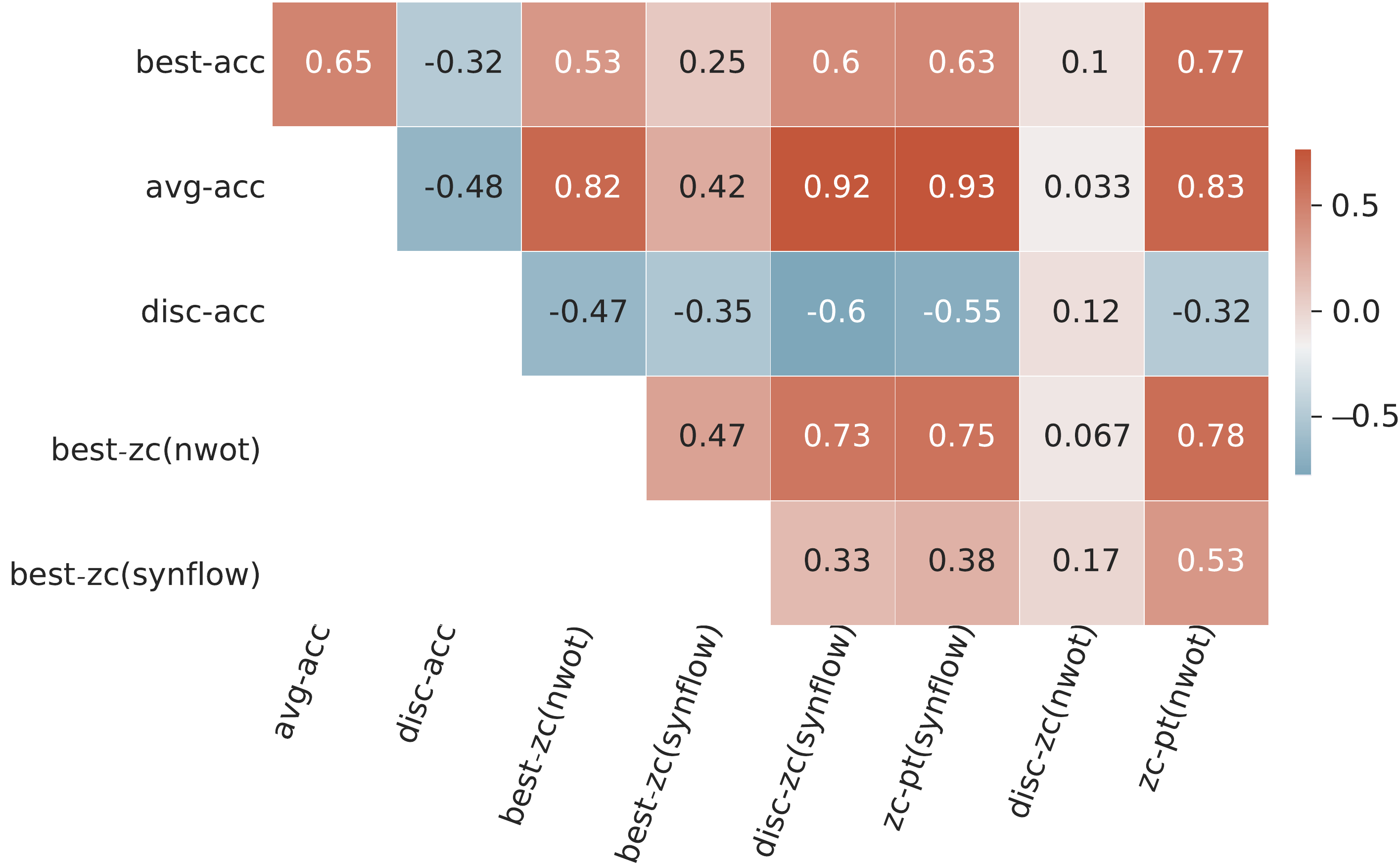}
  \caption{}
 \label{best-zc_correlation}
 \end{subfigure} 
 \begin{subfigure}{0.32\textwidth}
 \includegraphics[width=\textwidth, trim=0 0 0 1cm]{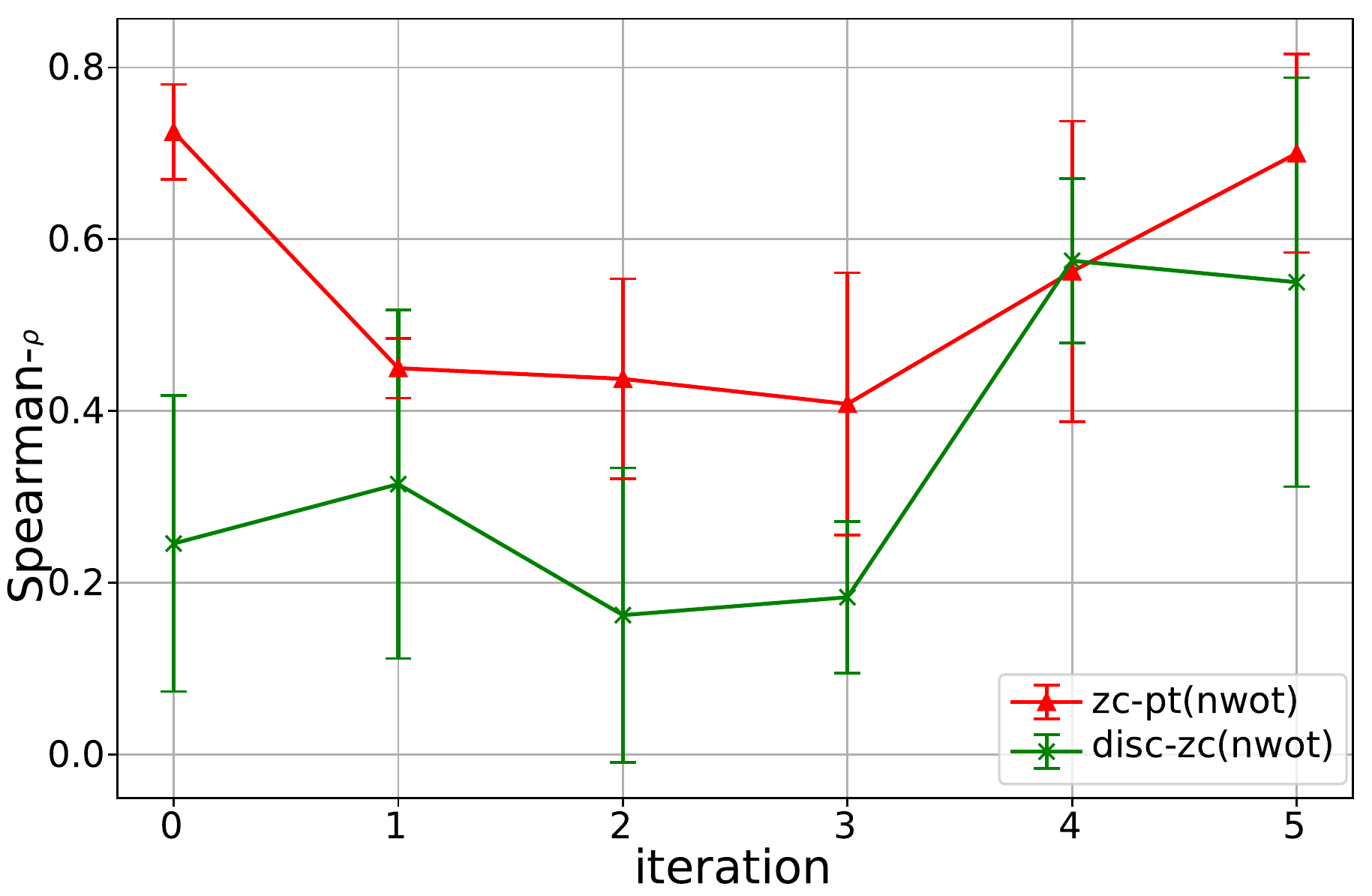}
  \caption{}
 \label{best-zc_correlation_time}
 \end{subfigure} 
  \caption{(a) Spearman's rank correlation coefficient of different zero-cost operation scoring metrics with the oracle metrics at the first iteration of NAS. (b) Rank correlation coefficient of disc-zc(\texttt{nwot}) and zc-pt(\texttt{nwot}) vs. best-zc(\texttt{nwot}) when invoked iteratively for each edge. }
\end{figure*}

As discussed in Section~\ref{sub:op-eval}, our scoring functions disc-zc and zc-pt use a zero-cost proxy $S$ instead of validation accuracy when discretizing an edge or perturbing an operation. As shown by~\cite{zero-cost,nwot}, the zero-cost score of a final architecture $S(A_{|\mathcal{E}|})$ correlates to the model accuracy after full training $V^{*}(A_{|\mathcal{E}|})$, subject to certain levels of fidelity.
In that sense, during iterative NAS process like the one in our work, a well performing zero-cost operation scoring function should lead to models $A_{|\mathcal{E}|}$ with high zero-cost scores $S(A_{|\mathcal{E}|})$, which have better chances to achieve high accuracy. To assess the capability of different zero-cost scoring functions in tracking the final architecture score $S(A_{|\mathcal{E}|})$, here we define another oracle score function $\pi_{\text{best-zc}}$:
\begin{equation}\label{eq:opsel:best-zc}
    \pi_{\text{best-zc}}(A_t,e) = \argmax_{o \in \mathcal{O}_e} \max_{A_{|\mathcal{E}|} \in \mathcal{A}_{t,e,o}} S(A_{|\mathcal{E}|})
\end{equation}


which is the best achievable zero-cost score of all possible architectures under metric $S$ with $(o,e)$ being selected.
Fig.~\ref{best-zc_correlation} shows the correlation of different scoring functions at iteration 0 evaluated on NAS-Bench-201.
First, we see that \texttt{synflow} metric, when combined with both discretization (disc-zc(\texttt{synflow})) and perturbation (zc-pt(\texttt{synflow})) paradigms, correlates well with best-acc and best-zc.
On the other hand, for \texttt{nwot} metric, there is a big discrepancy between the two paradigms.
Specifically, while zc-pt(\texttt{nwot}) shows the strongest overall performance in tracking both best-acc and best-zc its counterpart, disc-zc(\texttt{nwot}), does not correlates well with neither, further confirming our findings in Section~\ref{sub:op-eval-corr}.
Similarly in the progressive setting, as shown in Fig.~\ref{best-zc_correlation}, disc-zc(\texttt{nwot}) is weakly/inversely correlated with best-zc(\texttt{nwot}), explaining its inferior NAS performance (more details shown later by the Zero-Cost-DISC baseline in Appendix~\ref{sec:app:pt_vs_disc}).
Similar analysis on NAS-Bench-1Shot1~\cite{nasbench-1shot1} can be found in Appendix~\ref{sec:app:nb1s1}.

\begin{figure}[t]
 \centering
 \includegraphics[width=0.8\textwidth, trim=0 0 0 1cm]{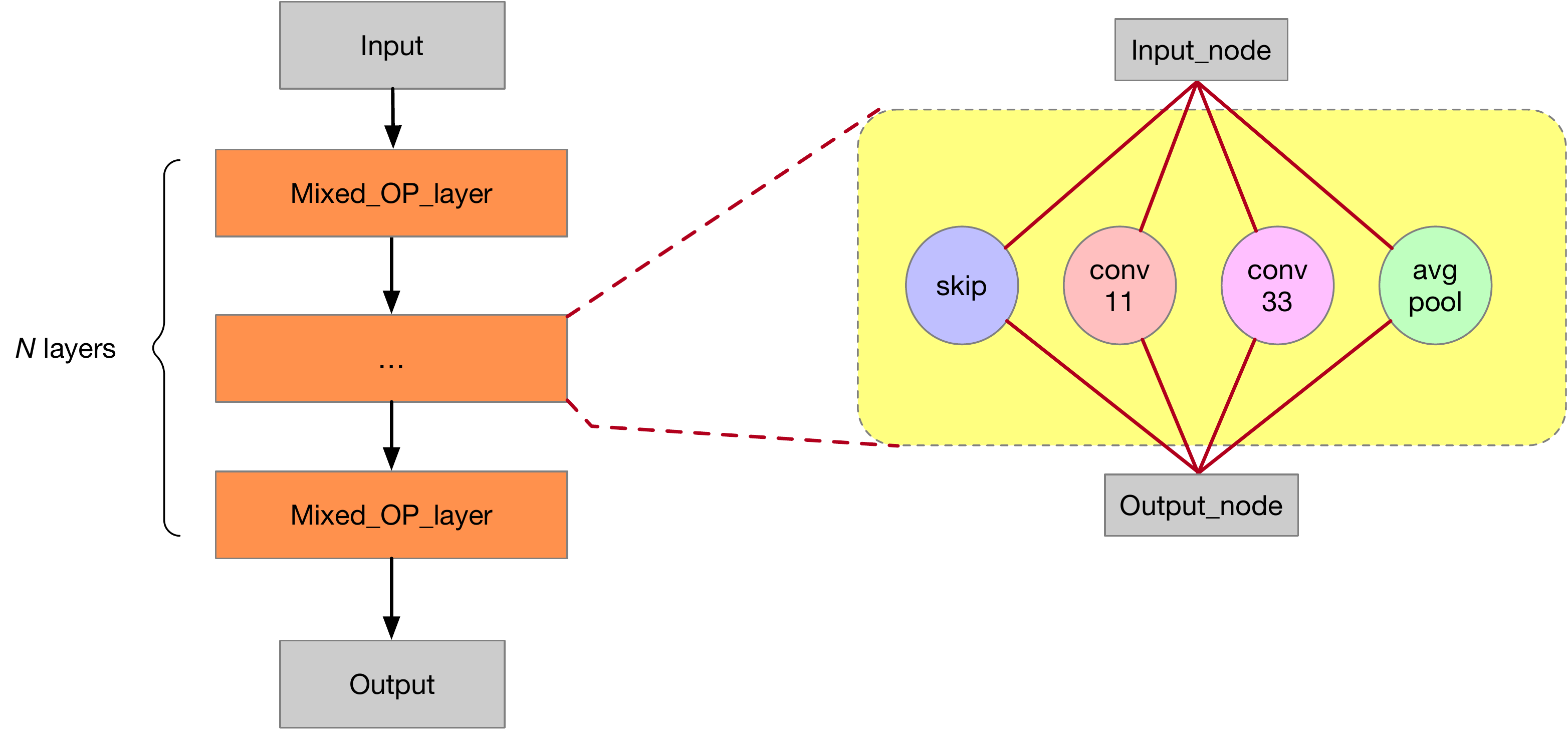}
\caption{A simplified supernetowrk with $N$ layers of mixed operations: \{\texttt{skip}, \texttt{conv\_1x1}, \texttt{conv\_3x3}, \texttt{avg\_pooling}\}.}
 \label{fig:app:toymodel_overview}
\end{figure}

\begin{figure}[h]
 \centering
 \includegraphics[width=0.7\textwidth, trim=0 0 0 1cm]{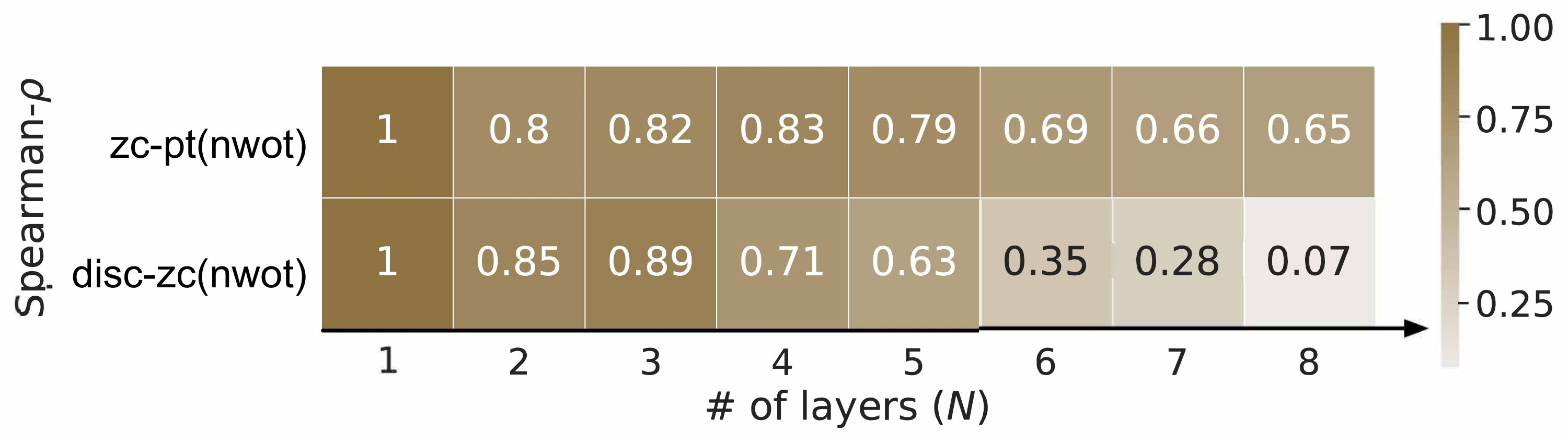}
\caption{Rank correlation coefficient of disc-zc(\texttt{nwot}) and zc-pt(\texttt{nwot}) \textit{vs.} best-zc(\texttt{nwot}) (at iteration 0) with supernetworks of different number of layers $N$.} 
 \label{fig:app:toymodel_skiprank}
\end{figure}

\begin{wraptable}{rt}{8cm}
\setlength\tabcolsep{3pt}
\centering
\caption {Number of times different operations were selected by disc-zc(\texttt{nwot}), zc-pt(\texttt{nwot}) and best-zc(\texttt{nwot}) at iteration 0 in NAS-Bench-201.}
\footnotesize
\begin{threeparttable}
\begin{tabular}{@{}rccc@{}}
\toprule
             & disc-zc(\texttt{nwot}) & zc-pt(\texttt{nwot}) & best-zc(\texttt{nwot}) \\ \midrule
\texttt{none}         & 0               & 0               & 0               \\
\texttt{skip}         & 5               & 1               & 1               \\
\texttt{conv\_1x1}    & 1               & 2               & 4               \\
\texttt{conv\_3x3}    & 0               & 3               & 1               \\
\texttt{avg\_pooling} & 0               & 0               & 0               \\ \bottomrule
\end{tabular}
\scriptsize
\end{threeparttable}
\label{tbl:op_picked-201}
\end{wraptable}

To further investigate the different behaviour of disc-zc(\texttt{nwot}) and zc-pt(\texttt{nwot}), let us consider a simplified supernetwork as in Fig.~\ref{fig:app:toymodel_overview}. The supernetwork has a chain structure of $N$ layers of mixed operations, with four candidate operations: \{\texttt{skip}, \texttt{conv\_1x1}, \texttt{conv\_3x3}, \texttt{avg\_pooling}\}. We vary the number of layers $N$ and calculate Spearman's rank correlation coefficient of disc-zc(\texttt{nwot}) and zc-pt(\texttt{nwot}) \textit{vs.} best-zc(\texttt{nwot}) at iteration 0 (note that here we compute the scores for all operations on an edge and then average across all edges, as explained later in Appendix~\ref{sec:app:scores}). 

Figure~\ref{fig:app:toymodel_skiprank} shows the trend of the ranking correlation coefficient of different approaches as supernet depth $N$ increases. It is straightforward to see that in the simplest case when $N$ = 1, all three approaches make the same decision. As $N$ increases, zc-pt(\texttt{nwot}) keeps tracking best-zc(\texttt{nwot}) well, while on the other hand, disc-zc(\texttt{nwot}) performs particularly bad. For instance, with the $N$ = 8 layer supernet,  disc-zc(\texttt{nwot}) only weakly/inversely correlates with best-zc(\texttt{nwot}) (rank correlation coefficient 0.07), while zc-pt(\texttt{nwot}) remains strongly correlated (rank correlation coefficient 0.65). 

\begin{figure}[H]
 \centering
 \includegraphics[width=0.55\textwidth, trim=0 0 0 1cm]{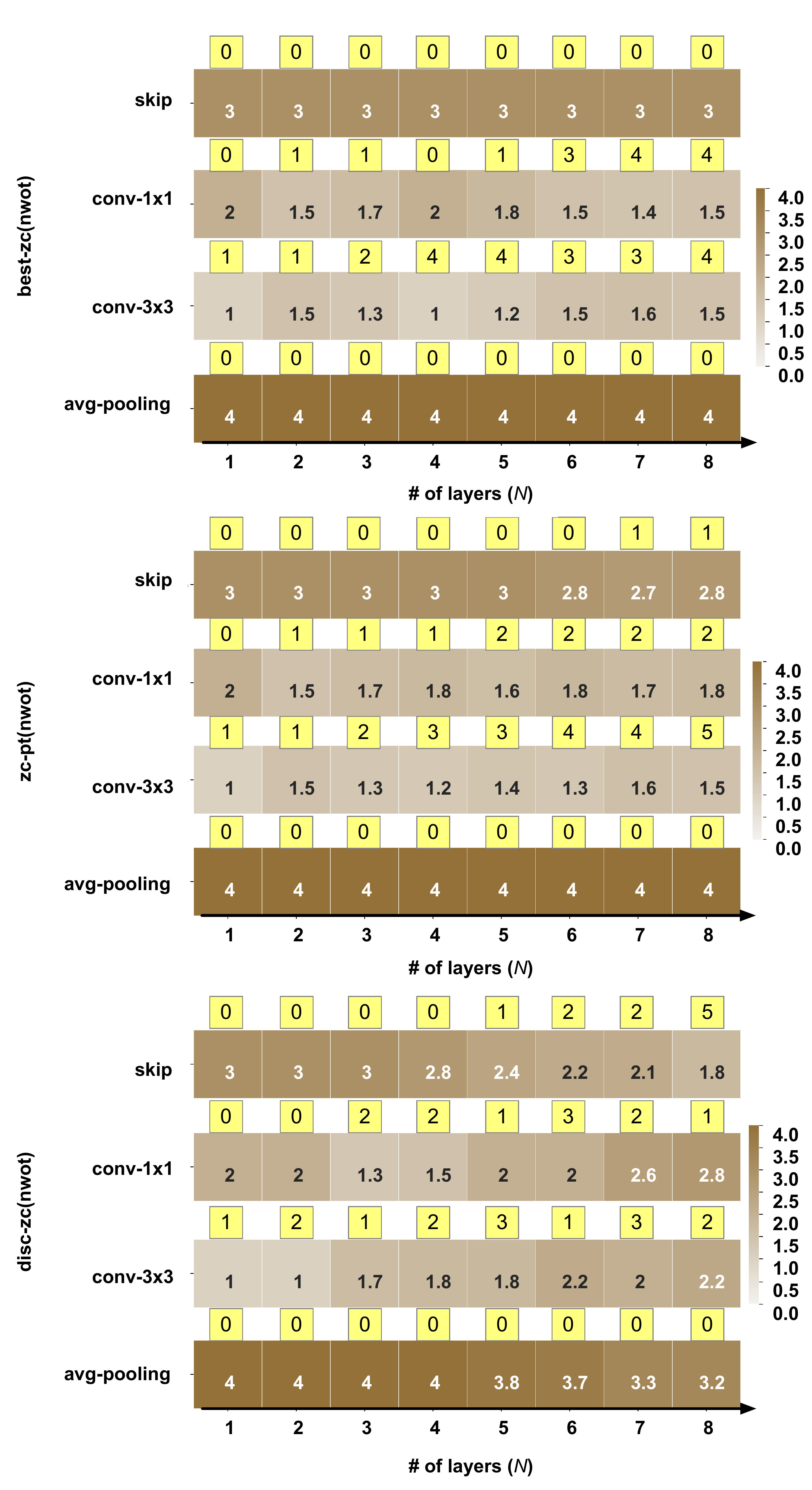}
\caption{Average rank of operations computed by best-zc(\texttt{nwot}) \textbf{(Top)},  zc-pt(\texttt{nwot}) \textbf{(Middle)} and disc-zc(\texttt{nwot}) \textbf{(Bottom)} with supernetworks of different number of layers $N$. The number of times where each operation is selected at iteration 0 ($N$ possible edges to discretize in total) are highlighted in yellow boxes above.}
 \label{fig:app:toymodel_rankselect}
\end{figure}

We further plot the average rank of each candidate operation determined by best-zc(\texttt{nwot}), disc-zc(\texttt{nwot}) and zc-pt(\texttt{nwot}) for supernets with different $N$, and the numbers of a particular operation being selected by different approaches when making the decision at iteration 0 (highlighted in yellow) in Figure~\ref{fig:app:toymodel_rankselect}. Clearly, we notice that as $N$ increases disc-zc(\texttt{nwot}) tends to erroneously favour the \texttt{skip} operation, e.g. for the $N$ = 8 layer supernet, in 5 out of 8 times\footnote{At iteration 0 there are 8 possible edges to discretize in total.} disc-zc(\texttt{nwot}) chose \texttt{skip} as the best operation. This is also evidenced by the the increase of the average rank of \texttt{skip} determined by disc-zc(\texttt{nwot}), resulting in the other operations such as \texttt{conv\_1x1} and \texttt{conv\_3x3} being muted in the NAS process.
On the other hand, we see that zc-pt(\texttt{nwot}) does not suffer from this problem, where the selection of operations is well aligned with the oracle best-zc(\texttt{nwot}). 

\begin{wraptable}{rt}{8cm}
\centering
\caption {Distribution of operations in the 10 final architectures ($\texttt{N}$ = 10 in our Zero-Cost-PT algorithm) generated by disc-zc(\texttt{nwot}) and zc-pt(\texttt{nwot}), compared with the best 10 architectures in the NAS-Bench-201 space.}
\footnotesize
\begin{threeparttable}
\begin{tabular}{@{}rccc@{}}
\toprule
             & Top 10 & disc-zc(\texttt{nwot}) & zc-pt(\texttt{nwot}) \\ \midrule
\texttt{none}         & 0               & 1              & 0            \\
\texttt{skip}         & 12               & 26             & 9            \\
\texttt{conv\_1x1}    & 12               & 14             & 20           \\
\texttt{conv\_3x3}     & 36               & 19             & 31           \\
\texttt{avg\_pooling}  & 0               & 0              & 0            \\ \bottomrule
\end{tabular}
\scriptsize
\end{threeparttable}
\label{tbl:op_picked_n10-201}

\end{wraptable}
With the above analysis on a simplified supernetwork, we assume that it could be the overwhelming selection of \texttt{skip} operation that causes the inferior performance of disc-zc(\texttt{nwot}).
Although the mechanism underpinning this observation is different, this also aligns with the well-known robustness issues of DARTS~\cite{arber2020understanding,shu2020understanding,yu2020evaluation, darts-pt}.
We observe similar behaviour of disc-zc(\texttt{nwot}) in NAS-Bench-201~\cite{nasbench2} space.
As shown in Table~\ref{tbl:op_picked-201}, at iteration 0, disc-zc(\texttt{nwot}) selected \texttt{skip} operation 5 times.
Note that the cell search space of NAS-Bench-201 has 6 edges in total, which means at the first iteration of NAS, disc-zc(\texttt{nwot}) is very likely to choose \texttt{skip} operation to discretize an edge. On the other hand, we see both zc-pt(\texttt{nwot}) and  best-zc(\texttt{nwot}) select \texttt{skip} operation for 1 time, which is reasonable.x 

We also took a closer look at the final architectures selected by disc-zc(\texttt{nwot}) and zc-pt(\texttt{nwot}) (using \texttt{N}=10 architectural proposal iterations as used in our main paper).
As shown in Table~\ref{tbl:op_picked_n10-201}, the 10 architectures selected by disc-zc(\texttt{nwot}) contain 26 \texttt{skip} operation in total, while zc-pt(\texttt{nwot}) only selected 12.
We also see that with the top 10 architectures, i.e. the 10 models with the highest validation accuracy in the entire space contains 12 \texttt{skip}, which is very similar to our zc-pt(\texttt{nwot}).
For completeness we provide the list of the 10 architecture selected by disc-zc(\texttt{nwot}) below, where we can clearly see \texttt{skip} operations dominating other choices: 

\begin{lstlisting}
[{"id": "0", "genotype": "|nor_conv_3x3~0|+|skip_connect~0|nor_conv_3x3~1|+|skip_connect~0|nor_conv_1x1~1|skip_connect~2|"}, 
{"id": "1", "genotype": "|nor_conv_3x3~0|+|skip_connect~0|nor_conv_3x3~1|+|skip_connect~0|nor_conv_3x3~1|skip_connect~2|"},
{"id": "2", "genotype": "|nor_conv_3x3~0|+|skip_connect~0|nor_conv_3x3~1|+|nor_conv_3x3~0|nor_conv_3x3~1|skip_connect~2|"}, 
{"id": "3", "genotype": "|nor_conv_3x3~0|+|skip_connect~0|nor_conv_3x3~1|+|skip_connect~0|nor_conv_1x1~1|skip_connect~2|"}, 
{"id": "4", "genotype": "|nor_conv_3x3~0|+|skip_connect~0|nor_conv_1x1~1|+|skip_connect~0|nor_conv_3x3~1|skip_connect~2|"}, 
{"id": "5", "genotype": "|skip_connect~0|+|nor_conv_1x1~0|nor_conv_1x1~1|+|nor_conv_3x3~0|skip_connect~1|nor_conv_1x1~2|"}, 
{"id": "6", "genotype": "|skip_connect~0|+|nor_conv_3x3~0|nor_conv_1x1~1|+|skip_connect~0|skip_connect~1|nor_conv_1x1~2|"}, 
{"id": "7", "genotype": "|nor_conv_3x3~0|+|skip_connect~0|nor_conv_1x1~1|+|skip_connect~0|nor_conv_1x1~1|none~2|"}, 
{"id": "8", "genotype": "|skip_connect~0|+|nor_conv_3x3~0|nor_conv_3x3~1|+|skip_connect~0|skip_connect~1|nor_conv_1x1~2|"}, 
{"id": "9", "genotype": "|skip_connect~0|+|nor_conv_1x1~0|nor_conv_1x1~1|+|nor_conv_3x3~0|skip_connect~1|nor_conv_1x1~2|"}]
\end{lstlisting}

We conduct similar analysis on NAS-Bench-1Shot1~\cite{nasbench-1shot1} space which does not contain \texttt{skip} in the candidate operation set later in Appendix~\ref{sec:app:nb1s1}, and the results further confirm our analysis with the above simplified supernetwork and NAS-Bench-201. 

It is also worth pointing out that while the aforementioned observations suggest significant limitations of using discretization paradigm together with the \texttt{nwot} metric, the exact mechanism causing those issues remains an open question. In this work we focus on designing a robust zero-cost operation scoring method (zc-pt) and developing a lightweight but effective NAS algorithm (Zero-Cost-PT) based on that, while we leave the investigation of why other approaches might not work well as future work. We include more discussion on the limitations of our work in Appendix~\ref{sec:app:limitations}.

\newpage


\newpage
\subsection{Detailed Zero-Cost-PT Algorithm}
\label{sec:app:algo}
Algorithm~\ref{algo:zero-cost-pt-complete} presents the proposed Zero-Cost-PT algorithm introduced in Section~\ref{sub:arch-search}. 
It has two stages: \textit{searching} and \textit{validation}, where we first iteratively discretize the supernet $A_0$ based on zero-cost-based perturbation function $f_\text{zc-pt}$ (line 1 - 16), and then in the second stage we use the a zero-cost metric (the same which is used in $f_\text{zc-pt}$) to score the candidate architectures (line 17 - 20), and select the one with the highest end-to-end score. In particular for DARTS CNN search space, our Zero-Cost-PT algorithm has an additional topology selection step (line 8 - 14), where for each node in architecture we only retain the top two incoming edges based on the zc-pt score -- this is similar to the vanilla DARTS algorithm~\citep{darts}. For NAS-Bench-201 space our algorithm skips this topology selection step. 

\begin{algorithm}[t]
\footnotesize
\caption{Zero-Cost Perturbation-based Architecture Search (Zero-Cost-PT)}\label{algo:zero-cost-pt-complete}
\SetKwInOut{Input}{Input}
\Input{An \textit{untrained} supernetwork $A_0$ with set of edges $\mathcal{E}$ and set of nodes $\mathcal{N}$, \# of architecture proposal iterations $\mathtt{N}$, \# of validation iterations $\mathtt{V}$}
\KwResult{A selected architecture $A^{\ast}_{|\mathcal{E}|}$}
\tcp*[h]{Stage 1: Architecture Proposal} \\
$\mathcal{C} = \varnothing$\\
\For(){$i=1 : \mathtt{N}$ } 
{
    \For(){$t=1 : |\mathcal{E}|$ } 
    {
        Select next edge $e_t$ using the chosen discretization ordering
        
        
        $o_t = \pi_{\text{zc-pt}}(A_{t-1}, e_t)$ \\
        $A_t = A_{t-1} + (e_t,o_t)$

    }

    \While(\tcp*[h]{prune the edges of the obtained architecture $A_{|\mathcal{E}|}$}){$|\mathcal{N}|>0$} 
    {
        Randomly select a node $n\in\mathcal{N}$
        
        \ForAll{\normalfont{Input edge $e$ to node $n$}}
        {
            Evaluate the zc-pt score of the architecture $A_{|\mathcal{E}|}$ when $e$ is removed
        }
        
        Retain only edges $e_n^{(1)\ast}$, $e_n^{(2)\ast}$ with the 1st and 2nd best zc-pt score, and remove $n$ from $\mathcal{N}$
    }
    Add $A_{|\mathcal{E}|}$ to the set of candidate architectures $\mathcal{C}$ \\
}

\tcp*[h]{Stage 2: Architecture Validation}

\For(){$j=1 : \mathtt{V}$ }
{
Calculate $S^{(j)}(A)$ for each $A \in \mathcal{C}$ using a random mini-batch data;
}

Select the best architecture $A^{\ast}_{|\mathcal{E}|} = \argmax_{A \in \mathcal{C}} \sum_{j=1:\mathtt{V}} S^{(j)}(A)$;

\end{algorithm}


\subsection{Additional Details on NAS-Bench-201 ablations}
\label{sec:app:ablations}

\subsubsection{More Zero-Cost Metrics}

We also include the detailed results of using \texttt{tenas} metric (with multiple search seeds) with our Zero-Cost-PT in Table~\ref{tbl:201-tenas-complete}. For reproducibility, the discovered architectures are listed below: 

\begin{lstlisting}
arch_0=|none~0|+|nor_conv_3x3~0|avg_pool_3x3~1|+|none~0|nor_conv_3x3~1|none~2|
arch_1=|avg_pool_3x3~0|+|nor_conv_1x1~0|nor_conv_1x1~1|+|none~0|nor_conv_1x1~1|skip_connect~2|
arch_2=|nor_conv_1x1~0|+|none~0|none~1|+|none~0|none~1|nor_conv_3x3~2|
arch_3=|skip_connect~0|+|nor_conv_1x1~0|nor_conv_3x3~1|+|none~0|none~1|none~2| 
\end{lstlisting}

\begin{wraptable}{rt}{7.5cm}
\setlength\tabcolsep{3pt}
\centering
\caption {Test error (\%) of Zero-Cost-PT when using \texttt{tenas} metric on NAS-Bench-201.}
\footnotesize
\begin{threeparttable}
\begin{tabular}{@{}ccccc@{}}
\toprule
\textbf{Search Seed}    & \textbf{C10} &\textbf{C100}    &\textbf{ImageNet-16}  \\ \toprule
0 &  90.00 & 99.00 & 99.17 \\
1 &  10.27 & 35.14 & 64.75 \\
2 &  90.00 & 99.00 & 99.17 \\
3 &  90.00 & 99.00 & 99.17 \\
Avg &  70.07$_{\pm39.87}$ & 83.04$_{\pm31.93}$ & 90.57$_{\pm17.21}$ \\
Best &  10.27 & 35.14 & 64.75 \\ 
 \bottomrule 
\end{tabular}
\end{threeparttable}
\label{tbl:201-tenas-complete}
\end{wraptable}

We see that the \texttt{tenas} metric performs badly in this case, which is consistent with our findings reported in Sec.~{\ref{sec:scoring}} where the \texttt{tenas} metric achieves lower correlation than \texttt{synflow} or \texttt{nwot}. We also find that the architectures discovered by \texttt{tenas} metric are dominated by the \texttt{none} operations as shown above, resulting in its inferior performance (also see the detailed scores obtained by \texttt{tenas} metric at iteration 0 in Table~{\ref{tbl:raw-op-scores}}. This shows that as a metric designed to \textit{prune} operations, \texttt{tenas} fails to work well with iterative discretization (aiming to \textit{select} the optimal operations), further confirming our argument that it is not a simple task to design a robust zero-cost NAS methodology. 

\newpage

\subsubsection{Zero-Cost-DISC Baseline}
 \label{sec:app:pt_vs_disc}

In Section.~\ref{sec:scoring}, we propose two zero-cost operation scoring methods disc-zc and zc-pt (corresponding to policies $\pi_{\text{disc-zc}}$ and $\pi_{\text{zc-pt}}$), and study their correlation with an oracle metric best-acc on NAS-Bench-201~\citep{nasbench2}. We also provide more in-depth analysis on the two metrics in the above Appendix~\ref{sec:app:additional-op-scoring}, and present more results in NAS-Bench-1Shot1~\cite{nasbench-1shot1} in Appendix~\ref{sec:app:nb1s1}. 

\begin{wraptable}{rh}{8.5cm}
\centering
\setlength\tabcolsep{4pt}
\caption {Comparison in test error (\%) between zero-cost perturbation-based and discretization-based NAS on NAS-Bench-201.}
\footnotesize
\begin{threeparttable}
\begin{tabular}{@{}lccc@{}}
\toprule
\textbf{Method}\tnote{1}    & \textbf{CIFAR-10} &\textbf{CIFAR-100}    &\textbf{ImageNet-16}  \\ \toprule
\textbf{Zero-Cost-DISC}      &6.22$_{\pm 0.84}$	&28.18$_{\pm 2.01}$	&55.14$_{\pm 1.77}$\\ 
\textbf{Zero-Cost-PT}       &\best{5.97$_{\pm 0.17}$}	&\best{27.47$_{\pm 0.28}$}   &\best{53.82$_{\pm 0.77}$} \\ 
\bottomrule
\end{tabular}
\begin{tablenotes}
\scriptsize
\item[1] We use the same hyperparameter settings as reported in the main paper: \texttt{N}=10, \texttt{V}=100, \texttt{nwot} zero-cost metric and \texttt{random} edge discretization order.
\end{tablenotes}
\end{threeparttable}
\label{tbl:disc-zc-201}
\end{wraptable}

We find that discretization is generally a weaker scoring paradigm than perturbation (especially when \texttt{nwot} is used), as shown by their correlations with respect to the oracle score best-acc and best-zc. To further evaluate the end-to-end NAS performance of discretization vs. perturbation with zero-cost metrics, we consider a baseline named Zero-Cost-DISC, which discretizes the supernet based on $\pi_{\text{disc-zc}}$ instead of $\pi_{\text{zc-pt}}$. Here for simplicity, we only consider \texttt{nwot} metric, and details on how disc-zc computes the operation scores can be found in Appendix~\ref{appendix:experimental-details}. We compare the performance of Zero-Cost-DISC and our proposed Zero-Cost-PT on NAS-Bench-201~\citep{nasbench2}, as shown in Table~\ref{tbl:disc-zc-201}. We see that discretization (Zero-Cost-DISC) results in inferior performance compared to the proposed perturbation-based approach (Zero-Cost-PT) on all datasets, confirming our previous analysis on their correlations with the oracle metric.


\subsection{Additional Ablation Study and Baseline on DARTS CNN Space}
\label{sec:app:nv_darts}

\subsubsection{Architecture Proposal vs. Validation}

\begin{wraptable}{rt}{10.0cm}
\centering
\caption {Detailed performance of Zero-Cost-PT$_{\texttt{random}}$ with \texttt{N}=1, \texttt{V}=0, \texttt{nwot} metric on DARTS CNN space.}
\footnotesize
\begin{threeparttable}[]
\begin{tabular}{@{}ccccccl@{}}
\toprule
\multirow{3}{*}{\textbf{S. seed}~\tnote{1}} & \multicolumn{6}{c}{\textbf{Test Error} (\%)}\\
 \cmidrule(l){2-7}
            & \multicolumn{4}{c}{Training seed~\tnote{2}} & \multirow{2}{*}{Avg.} & \multirow{2}{*}{Best} \\ \cmidrule(lr){2-5}
            & 0 & 1 & 2 & 3 & & \\ \midrule
0           & 2.72 & 2.55 & 2.83 & 2.71 &  &  \\
1           & 3.25 & 3.26 & 3.28 & 3.20 & &  \\
2           & 2.59 & 2.84 & 2.59 & 2.79 &  &  \\
3           & 2.43 & 2.77 & 2.52 & 2.66 & \multirow{-4}{*}{2.81$_{\pm 0.29}$} & \multirow{-4}{*}{2.43} \\ \bottomrule
\end{tabular}
\scriptsize
\begin{tablenotes}
\scriptsize
\item[1] Random seeds for searching the architectures.
\item[2] Random seeds for training the selected architectures.
\end{tablenotes}
\end{threeparttable}
\label{tbl:zcpt-n1}
\end{wraptable}

We conducted an ablation study of our Zero-Cost-PT algorithm on NAS-Bench-201~\citep{nasbench2} in Section~\ref{sec:zc-pt}, aiming to decide the best possible configuration of the main hyperparameters of our algorithm: architecture proposal iterations \texttt{N}, validation iterations \texttt{V}, ordering of edges to follow when discretizing, and the zero-cost metric to use. In the following, we present additional ablations on the much larger DARTS CNN space.

We first study the impact of different architecture proposal iterations \texttt{N} and validation iterations \texttt{V} when Zero-Cost-PT uses \texttt{random} as the search order and \texttt{nwot} metric.

As detailed in Algorithm~\ref{algo:zero-cost-pt-complete}, Zero-Cost-PT$_{\texttt{random}}$ firstly proposes \texttt{N} candidate architectures using the proposed zero-cost perturbation paradigm, which are then evaluated in a lightweight manner during the validation phase to come up with a single outcome of a search. It extends the existing zero-cost NAS approach such as NASWOT~\citep{nwot} by including more sophisticated architecture selection phase based on the combination of zero-cost metrics and perturbations.

We first consider an extreme case, setting architecture proposal iteration \texttt{N}=1, where Zero-Cost-PT only proposes one architecture candidate (with \texttt{random} edge discretization order), and with no validation stage performed. The detailed results are shown in Table~\ref{tbl:zcpt-n1}. As can be seen, the average performance is affected quite significantly. However, the best model still happens to be on-par with our main results. This suggests what has already been mentioned in our main paper, that the searching phase alone tends to ``find'' many different architectures depending on, broadly speaking, random seed, and this randomness is especially visible in the case of random edge ordering (Figure~\ref{fig:n-v-population}). While the high variance might seem undesired at first, we empirically observe that the higher exploration resulting from it is beneficial for finding some very good models, e.g., \texttt{global-op-iter} discretization order tends to be less sensitive to random seed as it takes away one degree of randomness (edge order), producing more stable results on average, but at the same time limiting its ability to maximize the performance of the best model found (Table~\ref{tbl:darts-c10}). In order to maximize the performance of our method, we balance exploration (higher \texttt{N} + random edge order) and exploitation (higher \texttt{V}) in the searching and validation phases respectively.

Admittedly, the interplay between those two phases is crucial for our method. To further showcase how the validation phase complements the searching phase, we run additional ablations on the DARTS CNN space with \texttt{N}=10 and \texttt{V}=\{1,10,100\}, the results are shown in Table~\ref{tbl:zcpt-v1-10-100}. The results are consistent with what is shown in the main paper: higher \texttt{V} produces better results on average but does not affect the best case that much (the best model is still upper-bounded by what was found with \texttt{N}=1).

\begin{wraptable}{rt}{18.0cm}
\setlength\tabcolsep{6pt}
\centering
\caption {Detailed performance of Zero-Cost-PT$_{\texttt{random}}$ with \texttt{N}=10, \texttt{V}=\{1, 10, 100\}, \texttt{nwot} metric on DARTS CNN space.}
\footnotesize
\begin{threeparttable}[]
\begin{tabular}{cccccccc}
\toprule
\multirow{3}{*}{\textbf{\texttt{V}}}                & \multirow{3}{*}{\textbf{S. seed}~\tnote{1}}      & \multicolumn{6}{c}{\textbf{Test Error} (\%)}         \\ \cmidrule(l){3-8}
   &  & \multicolumn{4}{c}{Training seed~\tnote{2}}  & \multirow{2}{*}{Avg.}          & \multirow{2}{*}{Best} \\ \cmidrule(lr){3-6}
                &     & 0 & 1 & 2 & 3 &           & \\ \midrule

\multirow{4}{*}{1}        & 0                    & 3.08                                                                         & 3.16                          & 3.06                          & 2.96                          &                                  &                          \\
                          & 1                    & 2.74                                                                         & 2.91                          & 2.92                          & 2.92                          &                                  &                          \\
                          & 2                    & 2.96                                                                         & 3.10                          & 3.06                          & 2.90                          &                                  &                          \\
                          & 3                    & 2.86                                                                         & 2.85                          & 2.85                          & 2.65                          & \multirow{-4}{*}{2.93$_{\pm0.14}$} & \multirow{-4}{*}{2.65}   \\ \midrule
\multirow{4}{*}{10}       & 0                    & 3.08                                                                         & 3.16                          & 3.06                          & 2.96                          &                                  &                          \\
                          & 1                    & 2.74                                                                         & 2.91                          & 2.92                          & 2.87                          &                                  &                          \\
                          & 2                    & 2.77                                                                         & 2.71                          & 2.65                          & 2.76                          &                                  &                          \\
                          & 3                    & 2.83                                                                         & 3.00                          & 2.82                          & 2.87                          & \multirow{-4}{*}{2.88$_{\pm 0.14}$}  & \multirow{-4}{*}{2.65}   \\ \midrule
\multirow{4}{*}{100}   & 0                    & 2.86                                                                         & 2.97                          & 2.77                          & 2.82                          &                                  &                          \\
                          & 1                    & 2.51                                                                         & 2.47                          & 2.56                          & 2.43                          &                                  &                          \\
                          & 2                    & 2.74                                                                         & 2.73                          & 2.54                          & 2.62                          &                                  &                          \\
                          & 3                    & 2.43                                                                         & 2.77                          & 2.52                          & 2.64                          & \multirow{-4}{*}{2.64$_{\pm 0.16}$}  & \multirow{-4}{*}{2.43}  \\ \bottomrule
\end{tabular}
\scriptsize
\begin{tablenotes}
\scriptsize
\item[1] Random seeds for searching the architectures.
\item[2] Random seeds for training the selected architectures.
\end{tablenotes}
\end{threeparttable}
\label{tbl:zcpt-v1-10-100}
\end{wraptable}

\subsubsection{Different Zero-Cost Metrics}
\label{ssec:app:darts_metrics}

\begin{wraptable}{rt}{8.5cm}
\centering
\setlength\tabcolsep{6pt}
 \vspace{-0.3cm}
\caption {Test error (\%) of Zero-Cost-PT$_\texttt{random}$ when using different zero-cost metrics.}
\vspace{-0.2cm}
\footnotesize
\begin{threeparttable}
\begin{tabular}{@{}lccc@{}}
\toprule
\textbf{Metric}    & \textbf{Average Error} &\textbf{Best Error}     \\ \toprule
\texttt{synflow}\tnote{1}~\cite{synflow}     &3.88$_{\pm 0.56}$	& 3.38 	\\
\texttt{zen\_score}~\cite{ming_zennas_iccv2021}     &3.06$_{\pm 0.31}$	& 2.68	\\
\texttt{nwot}~\cite{nwot} &\best{2.64$_{\pm 0.16}$}	&\best{2.43}  	\\
\bottomrule
\end{tabular}
\begin{tablenotes}
\scriptsize
    \item[1] Only 1 model was selected across all 4 random seeds.    
\end{tablenotes}
\end{threeparttable}
\label{tbl:metrics-darts}
\end{wraptable}

We also conducted experiments on DARTS CNN space using different zero-cost metrics with our Zero-Cost-PT algorithm, under the setting of \texttt{N}=10, \texttt{V}=100, and random edge order. In particular, in addition to the \texttt{nwot} metric, we choose the next two best performing metrics, \texttt{synflow}~\cite{synflow} and \texttt{zen\_score}~\cite{ming_zennas_iccv2021}. We perform search with 4 random seeds, and each discovered model is trained using 4 random seeds respectively. The results are presented in Table~\ref{tbl:metrics-darts}.

We see that the results are consistent with our analysis on the tabular benchmarks, where \texttt{nwot} outperforms both \texttt{synflow} and \texttt{zen\_score} in terms of average and best error by a large margin. In particular, similar to what we observed in NAS-Bench-201, in this case \texttt{synflow} also discovers the same architecture across 4 random seeds, which selects the operation with the most parameters (\texttt{sep\_conv\_5x5}): 

\begin{lstlisting}
Genotype(normal=[["sep_conv_5x5", 0], ["sep_conv_5x5", 1], ["sep_conv_5x5", 1], ["sep_conv_5x5", 2], ["sep_conv_5x5", 2], ["sep_conv_5x5", 3], ["sep_conv_5x5", 3], ["sep_conv_5x5", 4]], normal_concat=[2, 3, 4, 5], reduce=[["sep_conv_5x5", 0], ["sep_conv_5x5", 1], ["sep_conv_5x5", 1], ["sep_conv_5x5", 2], ["sep_conv_5x5", 2], ["sep_conv_5x5", 3], ["sep_conv_5x5", 3], ["sep_conv_5x5", 4]], reduce_concat=range(2, 6))
\end{lstlisting}

\subsubsection{Maximum-param Baseline}

\begin{wraptable}{rt}{8.5cm}
\centering
\caption {Randomly selected architectures with only operation \texttt{sep\_conv\_5x5} on DARTS CNN space.}
\footnotesize
\begin{threeparttable}[]
\begin{tabular}{@{}ccccccl@{}}
\toprule
\multirow{3}{*}{\textbf{S. seed}~\tnote{1}} & \multicolumn{6}{c}{\textbf{Test Error}(\%)}\\
 \cmidrule(l){2-7}
            & \multicolumn{4}{c}{Training seed~\tnote{2}}  & \multirow{2}{*}{Avg.} & \multirow{2}{*}{Best} \\ \cmidrule(lr){2-5}
            & 0    & 1    & 2    & 3    &                   & \\ \midrule
0           & 3.07 & 2.93 & 2.89 & 2.85 & 2.94$_{\pm 0.30}$ & 2.85 \\
1           & 2.92 & 2.93 & 3.17 & 3.05 & 3.02$_{\pm 0.20}$ & 2.92 \\
2           & 2.98 & 2.97 & 2.93 & 2.90 & 2.95$_{\pm 0.06}$ & 2.95 \\
3           & 2.87 & 2.83 & 3.02 & 2.78 & 2.88$_{\pm 0.18}$ & 2.78 \\ \bottomrule
\end{tabular}
\scriptsize
\begin{tablenotes}
\scriptsize
\item[1] Random seeds for searching the architectures.
\item[2] Random seeds for training the selected architectures.
\end{tablenotes}
\end{threeparttable}
\label{tbl:maxmum-param}
\end{wraptable}
We create a baseline approach that trains 4 random models with separable convolution 5x5 (the most expensive operation in the DARTS search space) selected everywhere and random connections between layers. The evaluation methodology follows all other experiments, i.e. we have 4 searched models trained 4 times with different training seeds, and we report the average and minimum error. 

We also show that the results on simple search spaces like NAS-Bench-201 (NB201), this maximum-param baseline may perform relatively better, e.g. it can find the model with test accuracy of 93.76\% (165th position in the ranking), but our method can still do better, discovering the 33rd best model in the search space (test accuracy 94.03\%). 

In summary, on both NB201 and DARTS space, the proposed combination of the perturbation paradigm with zero-shot proxies does better than the naive usage of the proxies presented in~\citep{zero-cost, nwot}, as shown in Tables~\ref{tbl:201}, \ref{tbl:darts-c10}, \ref{tbl:darts-imagenet}. Note that we are better than the plain perturbation-based baseline~\citep{darts-pt} and recent zero-cost NAS~\citep{chen2020tenas} on NB201 (Table~\ref{tbl:201}), and comparable/better on DARTS and derived subspaces (Table~\ref{tbl:darts-c10} and \ref{tbl:darts-imagenet}), while being much cheaper to run.

The structures of models are provided as Genotype objects below for reproducibility.

\begin{lstlisting}
random_max_0 = Genotype(normal=[["sep_conv_5x5", 0], ["sep_conv_5x5", 1], ["sep_conv_5x5", 0], ["sep_conv_5x5", 1], ["sep_conv_5x5", 0], ["sep_conv_5x5", 3], ["sep_conv_5x5", 1], ["sep_conv_5x5", 4]], normal_concat=range(2, 6), reduce=[["sep_conv_5x5", 0], ["sep_conv_5x5", 1], ["sep_conv_5x5", 0], ["sep_conv_5x5", 1], ["sep_conv_5x5", 0], ["sep_conv_5x5", 1], ["sep_conv_5x5", 0], ["sep_conv_5x5", 1]], reduce_concat=range(2, 6))
random_max_1 = Genotype(normal=[["sep_conv_5x5", 0], ["sep_conv_5x5", 1], ["sep_conv_5x5", 0], ["sep_conv_5x5", 2], ["sep_conv_5x5", 0], ["sep_conv_5x5", 2], ["sep_conv_5x5", 1], ["sep_conv_5x5", 4]], normal_concat=range(2, 6), reduce=[["sep_conv_5x5", 0], ["sep_conv_5x5", 1], ["sep_conv_5x5", 0], ["sep_conv_5x5", 1], ["sep_conv_5x5", 0], ["sep_conv_5x5", 1], ["sep_conv_5x5", 0], ["sep_conv_5x5", 1]], reduce_concat=range(2, 6))
random_max_2 = Genotype(normal=[["sep_conv_5x5", 0], ["sep_conv_5x5", 1], ["sep_conv_5x5", 0], ["sep_conv_5x5", 2], ["sep_conv_5x5", 0], ["sep_conv_5x5", 3], ["sep_conv_5x5", 1], ["sep_conv_5x5", 4]], normal_concat=range(2, 6), reduce=[["sep_conv_5x5", 0], ["sep_conv_5x5", 1], ["sep_conv_5x5", 0], ["sep_conv_5x5", 1], ["sep_conv_5x5", 0], ["sep_conv_5x5", 1], ["sep_conv_5x5", 0], ["sep_conv_5x5", 1]], reduce_concat=range(2, 6))
random_max_3 = Genotype(normal=[["sep_conv_5x5", 0], ["sep_conv_5x5", 1], ["sep_conv_5x5", 0], ["sep_conv_5x5", 1], ["sep_conv_5x5", 0], ["sep_conv_5x5", 2], ["sep_conv_5x5", 1], ["sep_conv_5x5", 4]], normal_concat=range(2, 6), reduce=[["sep_conv_5x5", 0], ["sep_conv_5x5", 1], ["sep_conv_5x5", 0], ["sep_conv_5x5", 1], ["sep_conv_5x5", 0], ["sep_conv_5x5", 1], ["sep_conv_5x5", 0], ["sep_conv_5x5", 1]], reduce_concat=range(2, 6))
\end{lstlisting}

\subsubsection{Random Sample Baseline(NASWOT)}
In Section~\ref{sub:results-201}, we compared our method to sampling-based zero-cost NAS in Table~\ref{tbl:201} (see NASWOT lines). Our results are empirically better on all three datasets. Additionally, our method computes the operation score per edge in a supernet, whereas the sampling-based approach computes the end-to-end network score. The relationship between the number of subnetworks and the number of operations is exponential. Therefore, we anticipate having to sample exponentially many networks in sample-based NASWOT~\citep{nwot} compared to our proposed Zero-Cost-PT.

In order to extend the comparison between zero-cost NAS (NASWOT) and our zero-cost PT to the DARTS CNN search space, we have conducted further experiments, in which we allow NASWOT to sample and score random models from the DARTS search space for a specified amount of samples (2500, 20000, 50000, corresponding to roughly 25min, 2h, and 5h on a single 2080Ti GPU) and the best model, according to the \texttt{nwot} metric, is selected as the result of a search. For each of the three sampling budgets, we run the entire process 4 times using different searching seeds (0-3), thus resulting in 3$\times$4 = 12 final architectures (9 unique ones are selected). Each of the final architectures was then trained 4 times using different training seeds (0-3). 

As expected, both the average and best performance of the sample-based zero-cost search increases with more samples. However, the results are visibly behind our proposed method, even for the most expensive searches (5 hours). In addition, we see that increasing the search budget from 2 hours to 5 hours does not result in proportional gains in accuracy, compared to 25min, suggesting diminishing returns. We hypothesize it is related to the mentioned fact that the number of architectures grows exponentially, so we'd need to sample significantly more networks before the probability of hitting a good one increases noticeably. In fact, we find that when increasing the search budget from 20000 to 50000 samples in our experiments, the baseline only gets better results in one out of 4 cases (searching seeds), and in one it actually made the results worse. This further suggests diminishing returns.

\subsection{Experiments on DARTS subspaces (S1-S4)}
\label{sec:app:subspaces}

\subsubsection{Description of DARTS subspace (S1-S4)}
RobustDARTS introduced four different DARTS subspaces to evaluate the robustness of the original DARTS algorithm~\citep{arber2020understanding}.
In our work, we validate the robustness of Zero-Cost-PT against some of the more recent algorithms using the same subspaces originally proposed in the RobustDARTS paper (Section~\ref{sub:results-darts-sub}).
The search spaces are defined as follows:

\begin{itemize} [leftmargin=8mm, topsep=0pt]
  \item in S1 each edge of a supernet consists only of the two candidate operations having the highest magnitude of $\alpha$ in the vanilla DARTS (these operations can be different for different edges);
  \item S2 only considers two operations: \texttt{skip\_connect} and \texttt{sep\_conv\_3x3};
  \item similarly, S3 consists of three operations: \texttt{none}, \texttt{skip\_connect} and \texttt{sep\_conv\_3x3};
  \item finally, S4 again considers just two operations: \texttt{noise} and \texttt{sep\_conv\_3x3}, where \texttt{noise} operation generates random Gaussian noise $\mathcal{N} (0,1)$ in every forwards pass that is independent of the input. 
\end{itemize}

\subsubsection{Results on DARTS subspaces (S1 - S4)}
\label{sub:results-darts-sub}
It is well known that DARTS could generate trivial architectures with degenerative performance in certain cases.
\cite{arber2020understanding} have designed various special search spaces for DARTS to investigate its failure cases on them.
As in DARTS-PT, we consider spaces S1-S4 to validate the robustness of Zero-Cost-PT in a controlled environment (detailed specifications can be found in Appendix~\ref{sec:app:subspaces}).
As shown in Table~\ref{tbl:darts-sub}, our approach consistently outperforms the original DARTS, the state-of-the-art DARTS-PT and DARTS-PT(fix $\alpha$) across S1 to S3 on both datasets CIFAR-10 and CIFAR-100, while on SVHN it offers competitive performance comparing the competing algorithms (best in S1, second best in space S2/S3 with .08/.02\% gap).
This confirms that our Zero-Cost-PT is robust in finding good performing architectures in spaces where DARTS typically fails, e.g. it has been shown~\citep{darts-pt} that in S2 DARTS tends to produce trivial architectures saturated with skip connections.
On the other hand, we observe that Zero-Cost-PT doesn't perform well in search space S4, struggling with operation \texttt{noise}, which simply outputs a random Gaussian noise $\mathcal{N}(0,1)$ regardless of the input. 
This is expected as score $S(A_t-(e,o))$ can be completely random if $o$ = \texttt{noise}. 
However, since \texttt{noise} operation is not useful in NAS, we are satisfied with the robustness of Zero-Cost-PT. 

\begin{table}[t]
\centering
\caption {Comparison in test error (\%) with state-of-the-art perturbation-based NAS on DARTS spaces S1-S4.}
\footnotesize
\setlength\tabcolsep{3pt}
\footnotesize
\begin{threeparttable}
\begin{tabular}{@{}cccccc@{}}
\toprule
\multirow{2}{*}{\textbf{Space}} & \textbf{DARTS}\tnote{1} & \multicolumn{2}{c}{\textbf{DARTS-PT}\tnote{1}} & \multicolumn{2}{c}{\textbf{Zero-Cost-PT}\tnote{2}} \\ 

\cmidrule(lr){2-2} \cmidrule(lr){3-4} \cmidrule(l){5-6}
    & Best & Best & Best (fix $\alpha$) & Avg. & Best \\ \toprule
    \multicolumn{6}{l}{\textbf{CIFAR-10}} \\
    S1 &3.84	&3.5	& \secondbest{2.86}	& \best{2.75$_{\pm 0.28}$} & \best{2.55} \\ 
    S2 &4.85	&2.79	& \secondbest{2.59}	& \best{2.49$_{\pm 0.05}$} & \best{2.45}\\ 
    S3 &3.34	& \secondbest{2.49}	&2.52	& \best{2.47$_{\pm 0.09}$} & \best{2.40} \\ 
    S4 &7.20	& \secondbest{2.64}	&\best{2.58}	&5.23$_{\pm 0.76}$ & 4.69 \\ \midrule
    \multicolumn{6}{l}{\textbf{CIFAR-100}} \\
    S1 &29.64	&24.48	& \secondbest{24.4}	& \best{22.05$_{\pm 0.29}$} & \best{21.84} \\ 
    S2 &26.05	& \secondbest{23.16}	&23.3	& \best{20.97$_{\pm 0.50}$} & \best{20.61} \\ 
    S3 &28.9	&22.03	& \secondbest{21.94}	& \best{21.02$_{\pm 0.57}$} & \best{20.61} \\ 
    S4 &22.85	& \secondbest{20.80}	& \best{20.66}	&25.70$_{\pm 0.01}$ & 25.69\\ \midrule
    \multicolumn{6}{l}{\textbf{SVHN}} \\
    S1 &4.58	&2.62	&\secondbest{2.39}	& \best{2.37$_{\pm 0.06}$} & \best{2.33} \\ 
    S2 &3.53	&2.53	&\best{2.32}	& \secondbest{2.40$_{\pm 0.05}$}& \secondbest{2.36}	 \\ 
    S3 &3.41	&2.42	&\secondbest{2.32}	& 2.34$_{\pm 0.05}$ & \best{2.30} \\ 
    S4 &3.05	&\secondbest{2.42}	&\best{2.39}	&2.83$_{\pm 0.06}$ &2.79 \\ \bottomrule
\end{tabular}
\begin{tablenotes}
\scriptsize
    \item[1] Results taken from~\citep{darts-pt}.
    \item[2] Results obtained using random seeds 0 and 2.
\end{tablenotes}
\end{threeparttable}
\label{tbl:darts-sub}
\end{table}

\subsection{NAS Experimental Details}
\label{sec:app:exp-details}

All searches were run multiple times with different searching seeds (usually 0, 1, 2 and 3).
Additionally, each found architecture was trained multiple times using different training seeds -- for DARTS the same set of seeds was used for training and searching, for NAS-Bench-201 (NB201) training seeds were taken from the dataset (777, 888 and 999, based on their availability in the dataset).
Therefore, for each experiment, we got a total of \texttt{searching\_seeds} $\times$ \texttt{training\_seeds} accuracy values.
Whenever average performance is reported, it is averaged across all obtained results.
Similarly, the best values were selected by taking the best single result from the searching and training seeds.

\subsubsection{Experimental Details -- NAS-Bench-201}
Searching was performed using 4 different seeds (0, 1, 2, and 3) to initialise a supernet.
Whenever we had to perform training of a supernet during the searching phase (Section~\ref{sec:scoring}), we used the same hyperparameters as the original DARTS-PT code used.
When searching using our Zero-Cost-PT we used a batch size of 256, N=10, V=100 and S=\texttt{nwot}, unless mentioned otherwise (e.g., during ablation studies).
Inputs for calculating zero-cost scores came from the training dataloader(s), as defined for relevant datasets in the original DARTS-PT code (including augmentation).
For zero-cost proxies that require a loss function, standard cross-entropy was used.
For any searching method, after an algorithm had identified a final subnetwork, we extracted the final architecture and queried the NB201 dataset to obtain test accuracy – one value for each training seed available in the dataset.

All experiments concerning operation scoring (Sections~\ref{sec:scoring} and \ref{sec:app:scores}) used averaged accuracy of models from NB201 for simplicity.

We did not search for architectures targeting CIFAR-100 or ImageNet-16 directly – whenever we report results for these datasets, we used the same architecture found using CIFAR-10.



\subsubsection{Experimental Details -- DARTS}
DARTS experiments follow a similar methodology to NB201.
Each algorithm was run with 4 different initialisation seeds for a supernet (0, 1, 2 and 3).
When running Zero-Cost-PT, we used the following hyperparameters: the batch size of 64, \texttt{N}=10, \texttt{V}=100 and $S$=\texttt{nwot}.
Inputs and the loss function for zero-cost metrics were defined analogically to NB201.
We did not run any baseline method on the DARTS search space (all results were taken from the literature), so we did not have to perform any train a supernet.
After an algorithm had identified a final subnetwork, we then trained it 4 times using different initialisation seeds again (0, 1, 2 and 3).
When training subnetworks, we used a setting aligned with the previous work \citep{darts,darts-pt}.

Unlike NB201, whenever different datasets were considered, (Section~\ref{sub:results-darts-sub}) architectures were searched on each relevant dataset directly.

For CIFAR-10 experiments, we trained models using a heavy configuration with \texttt{init\_channels} = $36$ and \texttt{layers} = $20$.
Models found on CIFAR-100 and SVHN were trained using a mobile setting with \texttt{init\_channels} = $16$ and \texttt{layers} = $8$.
Both choices follow the previous work~\citep{arber2020understanding,darts-pt}.

\newpage

\subsection{Additional Experiments on NAS-Bench-1shot1}
\label{sec:app:nb1s1}

\begin{figure}[t]
 \centering
 \includegraphics[width=\textwidth, trim=0 0 0 1cm]{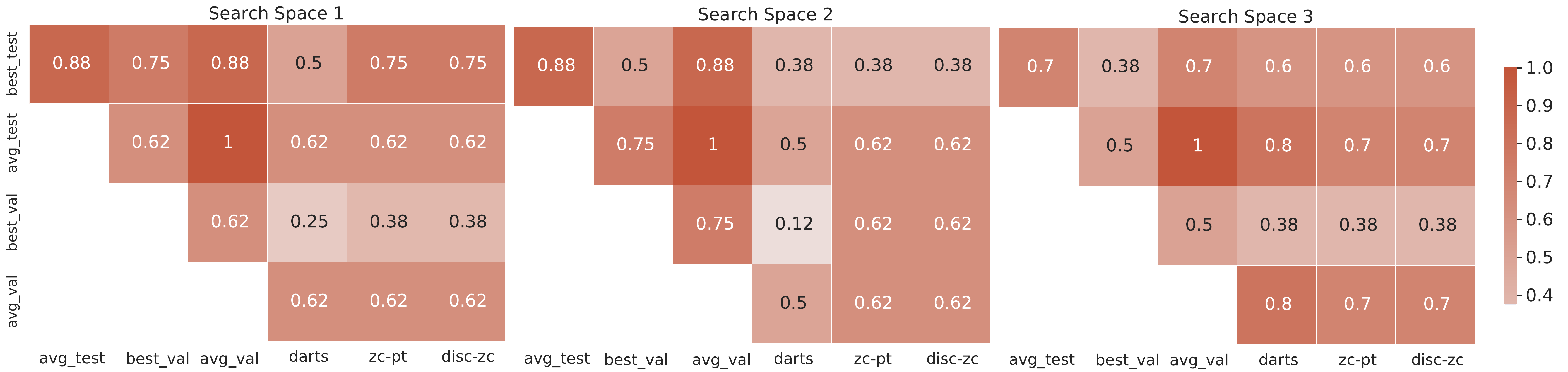}
\caption{Spearman's rank correlation coefficient of different operation scoring metrics with each other at the first iteration of NAS evaluated on NAS-Bench-1Shot1 benchmark.}
 \label{fig:app:correlation-1shot1}
\end{figure}

\begin{figure}[t]
 \centering
 \includegraphics[width=\textwidth, trim=0 0 0 1cm]{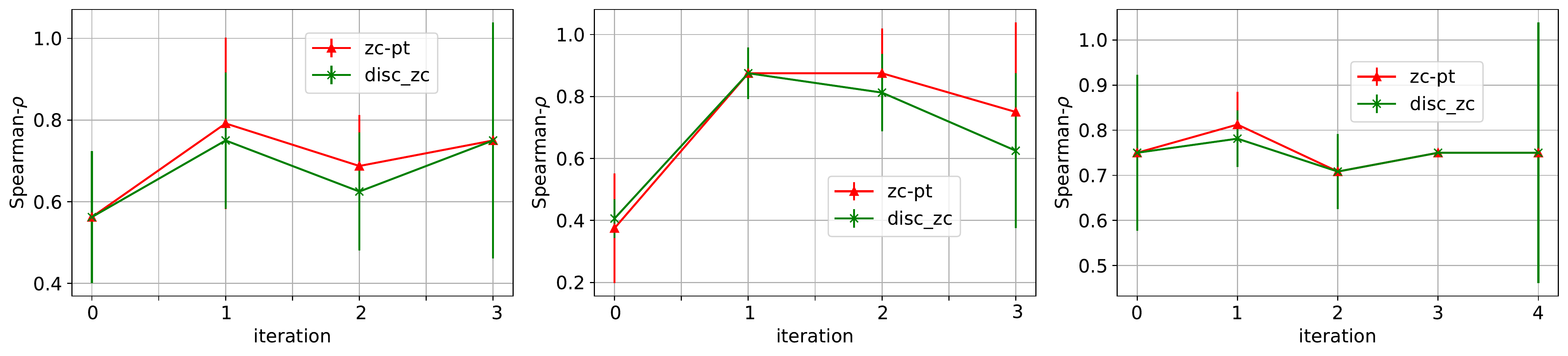}
\caption{Rank correlation coefficient of disc-zc(nwot)
and zc-pt(nwot) vs. best-zc(nwot) when invoked iteratively for each edge on NAS-Bench-1Shot1 benchmark.}
 \label{fig:app:correlation-1shot1_iter}
\end{figure}

\begin{wraptable}{rt}{8cm}
\setlength\tabcolsep{3pt}
\centering
\caption {Comparison in validation and test error (\%) of DARTS~\citep{darts} and our Zero-Cost-PT on NAS-Bench-1Shot1~\citep{nasbench-1shot1} benchmark.}
\footnotesize
\begin{threeparttable}[]
\begin{tabular}{lccccc}
\toprule
 \multirow{2}{*}{\textbf{Method}} & \multirow{2}{*}{\textbf{Space}} & \multicolumn{2}{c}{\textbf{Val Error [\%]} } & \multicolumn{2}{c}{\textbf{Test Error [\%]} } \\ \cmidrule(l){3-6}  
                                  &                                 & Avg.                          & Best          & Avg.                          & Best\\ \midrule
DARTS                             & \multirow{2}{*}{1}              & 6.11$_{\pm 0.01}$             & 5.90          & 6.67$_{\pm 0.08}$             & 6.44 \\
\textbf{Zero-Cost-PT}                      &                                 & \best{5.86$_{\pm 0.05}$}      & \best{5.15}   & \best{6.47$_{\pm 0.64}$}      & \best{5.61} \\ \midrule
DARTS                             & \multirow{2}{*}{2}              & 6.34$_{\pm 0.05}$             & 5.71          & 6.68$_{\pm 0.36}$             & 6.24 \\
\textbf{Zero-Cost-PT}                      &                                 & \best{5.89$_{\pm 0.04}$}      & \best{5.29}   & \best{6.65$_{\pm 0.360}$}     & \best{6.15} \\ \midrule
DARTS                             & \multirow{2}{*}{3}              & 6.15$_{\pm 0.02}$             & 5.90          & 6.64$_{\pm 0.12}$             & 6.5  \\
\textbf{Zero-Cost-PT}                      &                                 & \best{5.90$_{\pm 0.03}$}      & \best{5.31}   & \best{6.45$_{\pm 0.31}$}      & \best{5.84} \\

\bottomrule
\end{tabular}
\end{threeparttable}
\label{tbl:1shot1}
\end{wraptable}
In addition to experiments on NAS-Bench-201~\citep{nasbench2}, we perform additional experiments on NAS-Bench-1shot1~\citep{nasbench-1shot1} to further evaluate the performance of our Zero-Cost-PT algorithm on relevant NAS benchmarks. When searching using Zero-Cost-PT we used batch size of 256, N=10, V=100 and S=\texttt{nwot}. We first extend our correlation analysis from Section~\ref{sec:scoring} (in particular, the initial operation scoring as in Section~\ref{sub:op-eval-corr}). We compare to DARTS as it is already available in the NB1shot1 codebase~\footnote{\url{https://github.com/automl/nasbench-1shot1}}. As in Figure~\ref{fig:app:correlation-1shot1}, the results show that DARTS is surprisingly well-correlated to both best and avg accuracy, in some cases (only in Space 3) even better than our proposed zc-pt. We believe that this is because the search space does not contain skip connections and overall is rather unusual compared to others used with differentiable NAS, so it is possible that it constitutes an edge-case where DARTS performs relatively well. Due to the exact reason, we observe that the performance of disc-zc and zc-pt in tracking the oracle best-zc is very similar, as shown in Figure~\ref{fig:app:correlation-1shot1_iter}, further confirming our hypothesis about degrading effects of skip connections on disc-zc performance, presented in Appendix~\ref{sec:app:additional-op-scoring}.

Table~\ref{tbl:1shot1} shows the end-to-end NAS results. We can see that our Zero-Cost-PT consistently achieves better performance than DARTS in all 3 spaces, and is able to find better models (best-case scenarios). This confirms our analysis and evaluation presented in the main paper, that Zero-Cost-PT could offer strong NAS performance on various search spaces. We would like to notice that unlike DARTS or NAS-Bench-201, NAS-Bench-1shot1 supernet contains architectural parameters associated with entire connections between cells ($\alpha$ and $\gamma$ in the paper~\citep{nasbench-1shot1}), additionally to the standard ones associated with candidate operations in a single layer ($\beta$ in the paper). What it means for our method is that for those parameters we no longer perturb a single edge of a supernet but rather the entire path, making the setting for our algorithm noticeably different. Further extending our method from operation scoring to path scoring is a very relevant goal for future work.

\subsection{Additional Experiments on NAS-Bench-Macro}
\label{sec:app:nbmacro}

To further verify the robustness of our Zero-Cost-PT method, we conduct additional experiments on a different type of tabular NAS benchmark, NAS-Bench-Macro~\citep{nasbench-macro}. Unlike NAS-Bench-201~\citep{nasbench2} and NAS-Bench-1Shot1~\citep{nasbench-1shot1}, NAS-Bench-Macro is a NAS benchmark on macro search space, where candidate blocks are selected during NAS process instead of operations. It consists of 6561 networks with their test accuracies, parameters and FLOPs on CIFAR-10 dataset.
             
\subsubsection{Experimental setting}
We follow the same settings as in~\citep{nasbench-macro}. The NAS-Bench-Macro search space contains 8 searching layers, each of which has the following 3 candidate blocks:
\begin{itemize}
  \item \texttt{Identity}: identity connection (encoded as `0');
  \item \texttt{MB3\_K3}: MobileNetV2 block with kernel size 3 and expansion ratio 3 (encoded as `1');  
  \item  \texttt{MB6\_K5}: MobileNetV2 block with kernel size 5 and expansion ratio 6 (encoded as `2'). 
\end{itemize}

The total size of the search space is therefore $3^8=6561$, containing 3969 unique architectures. We adopt the same input dimension 32, and run our Zero-Cost-PT NAS algorithm with 4 different random seeds (0, 1, 2, 3) as in all other experiments. We query the discovered architectures and obtain their accuracies, parameters and FLOPs from the benchmark. When searching using Zero-Cost-PT we used batch size of 256, N=10, V=100 and S=\texttt{nwot} (same as in experiments on other tabular NAS benchmarks).

\begin{table}[h]
    \footnotesize
    \centering
    \setlength\tabcolsep{1pt}
    \caption {Comparison of MCT-NAS~\citep{nasbench-macro} and our Zero-Cost-PT on NAS-Bench-Macro benchmark.}
    \begin{threeparttable}[]
    \begin{tabular}{@{}cccccc@{}}
        \toprule
        \textbf{Methods}      & \textbf{Avg. Error [\%]} & \textbf{Params [M]}  & \textbf{FLOPs [M]}     & \textbf{Rank}   & \textbf{Search Cost}~\tnote{2} \\ \midrule
         MCT-NAS~\tnote{1} & 6.87$_{\pm 0.253}$ & 2.0 & 85.2 & 1/6561 & 0.68 \\   
        \textbf{Zero-Cost-PT} & 7.05$_{\pm 0.077}$ & 2.9 & 105.7 & 9/6561 & \best{0.05}         \\ \bottomrule
    \end{tabular}
    \begin{tablenotes}
    \scriptsize
    \item[1] Results taken from~\citep{nasbench-macro}. 
    \item[2] Averaged over 3 random seeds (used in training) as reported in~\citep{nasbench-macro}. 
    \item[3] In GPU hours. Cost for MCT-NAS~\citep{nasbench-macro} is estimated by measuring wall clock time of i) a single epoch supernet training and ii) a single forward pass on validation set with a 2080Ti GPU. No information on search cost is reported on CIFAR-10 in the paper, and the NAS code is not available. 
    \end{tablenotes}
    \end{threeparttable}
    \label{tbl:nb-macro}
\end{table}

\subsubsection{Results}
Table~\ref{tbl:nb-macro} compares the NAS results of our Zero-Cost-PT approach and the MCT-NAS algorithm reported on NAS-Bench-Macro~\citep{nasbench-macro}. We see that Zero-Cost-PT is able to discover a good architecture (ranked the 9th in all 6561 models), with much less search cost comparing to MCT-NAS reported in NAS-Bench-Macro~\citep{nasbench-macro}. Since MCT-NAS did not report search cost on CIFAR-10 dataset, nor release search code, we estimate the cost of MCT-NAS by running i) a single epoch for supernet training, and ii) a single forward pass on validation set with a 2080Ti GPU. We then calculate the total cost according to Algorithm 1 in ~\citep{nasbench-macro} (with 120 epoch of supernet training, 50 search numbers and 256 batch size).

In particular, during search our Zero-Cost-PT selects the same architecture across all 4 seeds (0, 1, 2, 3), which assigns \texttt{MB6\_K5} at every candidate block. Note that the top-1 architecture discovered by MCT-NAS~\citep{nasbench-macro} also tends to choose many \texttt{MB6\_K5} block, with an \texttt{Identity} block selected on the last layer. It is also worth pointing out that in NAS-Bench-Macro benchmark, architectures are only trained for 50 epochs, which may change with further training.

\subsection{Experimental Settings on MobileNet-like Search Space}
\label{sec:app:mobilenet}


We adopt the same settings as in~\citep{proxylessnas} and construct a supernet with 21 choice blocks, and each block has the following 7 alternative operations: 
\begin{itemize}
  \item 3$\times$2 = 6 MobileNet blocks, with 3 different kernel sizes \{3, 5, 7\} and 2 expansion ratio \{3, 6\}).
  \item skip connection.
 \end{itemize}


We follow the previous work~\citep{proxylessnas} on this space and search directly on ImageNet dataset~\citep{imagenet}, with \texttt{input\_size} = 224. We use 3 different random seeds (0-2) to perform architectures search, and train the discovered final models (3 models searched and trained in total). We use \texttt{batch-size} = 1024 and training was performed on 8 NVIDIA V100 GPUs for 300 epochs, with initial learning rate set to 0.5. We use colour-jitters, random horizontal flip and random crop for data augmentation, set \texttt{label\_smooth} = 0.1. All other training setting is identical to ProxylessNAS~\citep{proxylessnas}.



\subsection{Discovered Architectures}
\label{sec:app:models}
\begin{figure}[h]
 \centering
 \begin{subfigure}{0.45\columnwidth}
 \includegraphics[width=\textwidth]{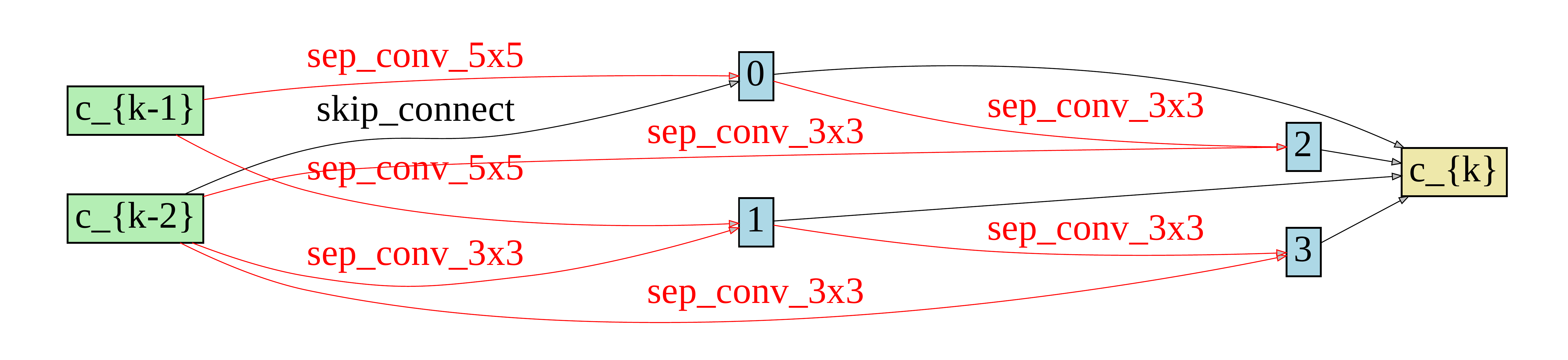}
  \caption{Normal cell}
 \end{subfigure} 
 \begin{subfigure}{0.54\columnwidth}
 \includegraphics[width=\textwidth]{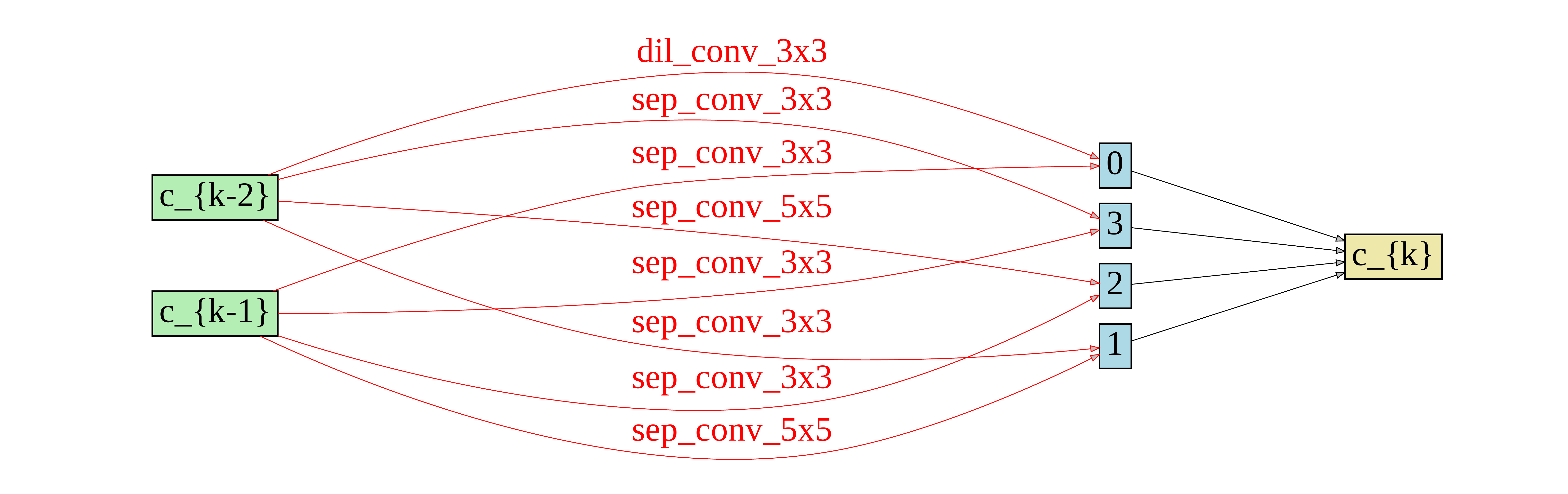}
  \caption{Reduction cell}
 \end{subfigure} 
  \caption{Cells found by Zero-Cost-PT (\texttt{global-op-iter} discretization order) on the DARTS search space using CIFAR-10.}
  \label{fig:app:darts_global}
 \vspace{-0.5cm}
\end{figure}

\begin{figure}[h]
 \centering
 \begin{subfigure}{0.45\columnwidth}
 \includegraphics[width=\textwidth]{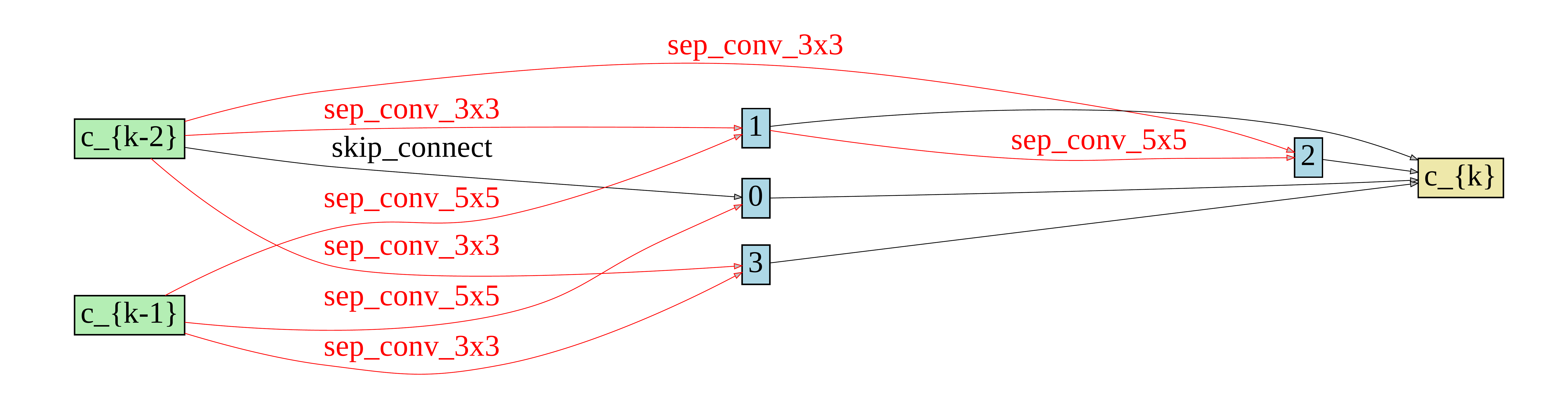}
  \caption{Normal cell}
 \end{subfigure} 
 \begin{subfigure}{0.54\columnwidth}
 \includegraphics[width=\textwidth]{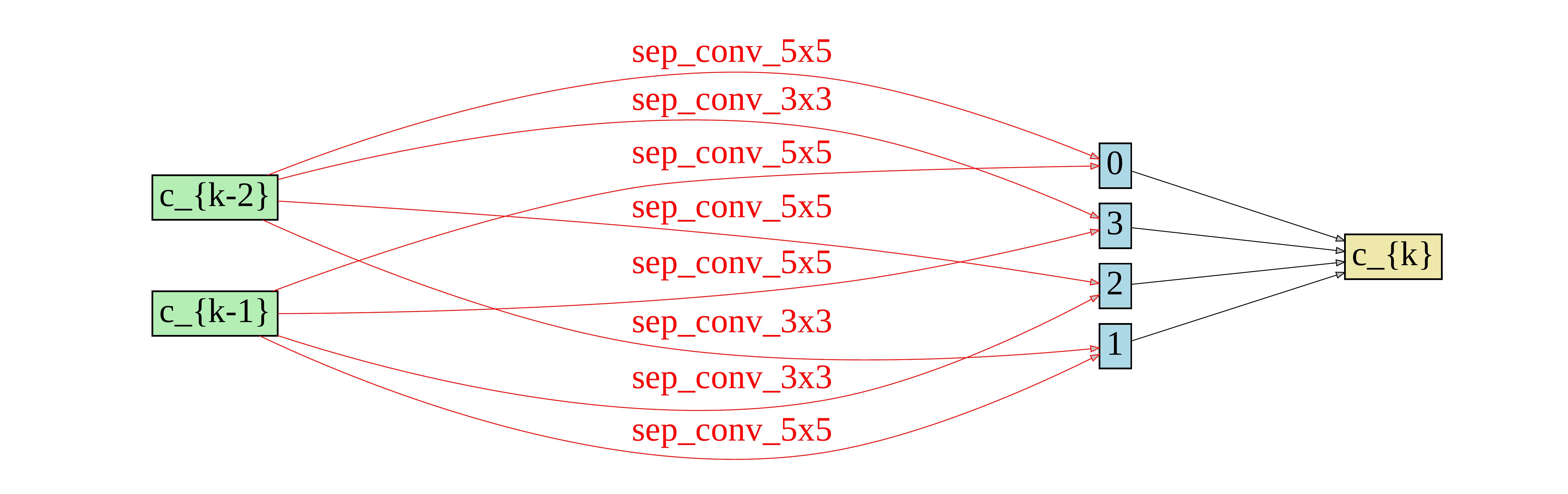}
  \caption{Reduction cell}
 \end{subfigure} 
  \caption{Cells found by Zero-Cost-PT (\texttt{random} discretization order) on the DARTS search space using CIFAR-10.}
  \label{fig:app:darts_random}
 \vspace{-0.5cm}
\end{figure}

\begin{figure}[h]
 \centering
 \begin{subfigure}{0.45\columnwidth}
 \includegraphics[width=\textwidth]{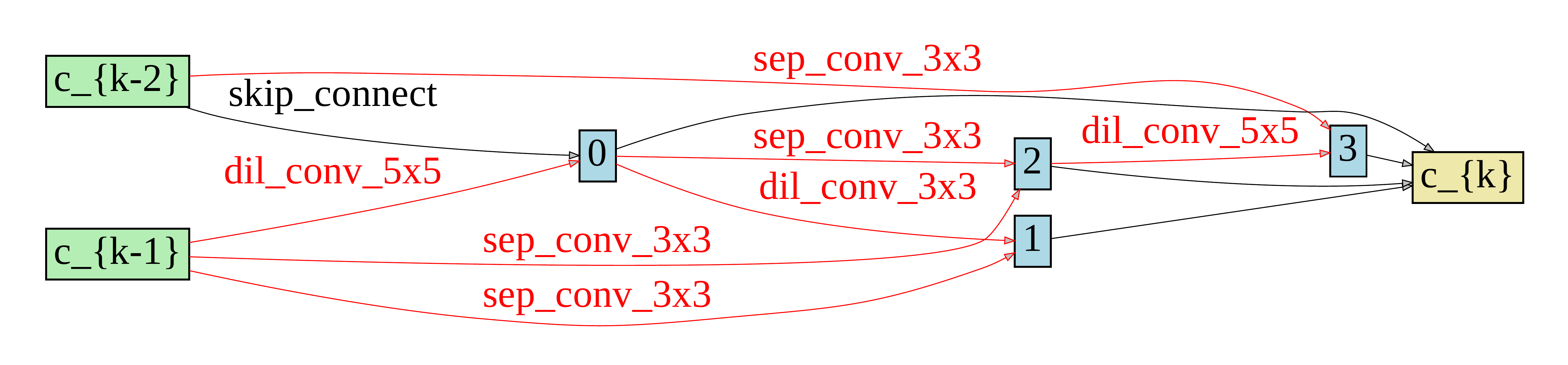}
  \caption{Normal cell}
 \end{subfigure} 
 \begin{subfigure}{0.54\columnwidth}
 \includegraphics[width=\textwidth]{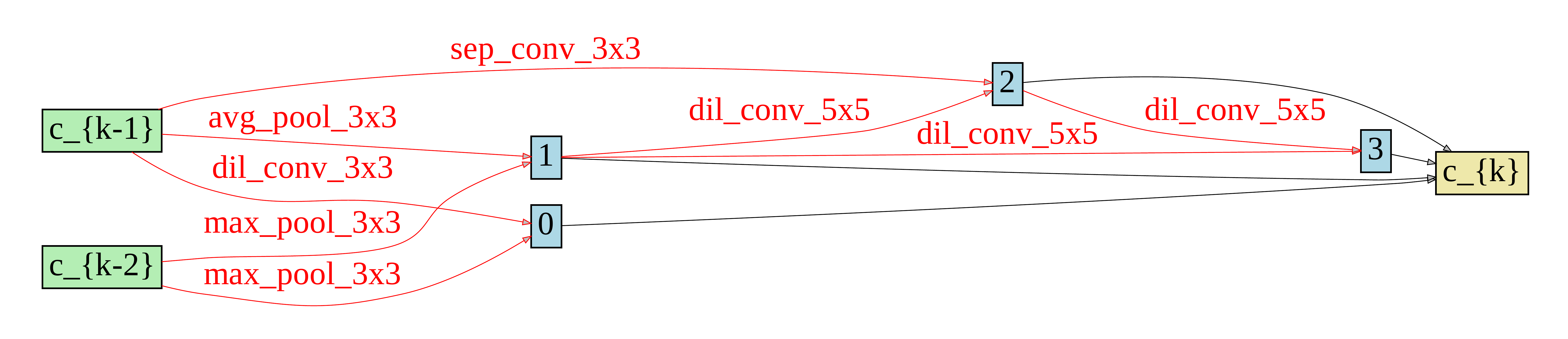}
  \caption{Reduction cell}
 \end{subfigure} 
  \caption{Cells found by Zero-Cost-PT (\texttt{random} discretization order) on the DARTS-S1 space using CIFAR-10.}
  \label{fig:app:darts_sub_first}
 \vspace{-0.5cm}
\end{figure}

\begin{figure}[t]
 \centering
 \begin{subfigure}{0.45\columnwidth}
 \includegraphics[width=\textwidth]{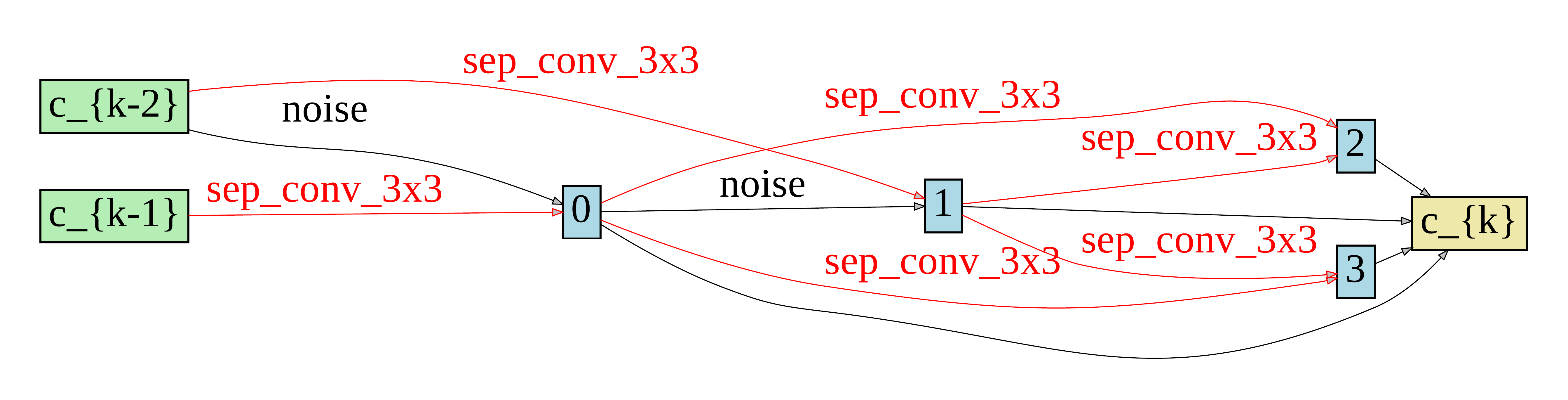}
  \caption{Normal cell}
 \end{subfigure} 
 \begin{subfigure}{0.54\columnwidth}
 \includegraphics[width=\textwidth]{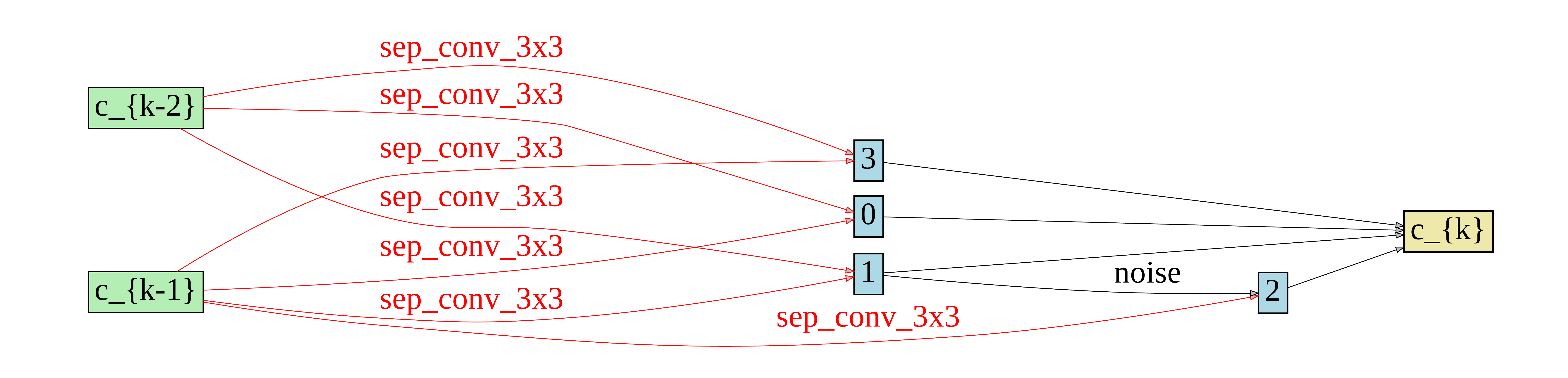}
  \caption{Reduction cell}
 \end{subfigure} 
  \caption{Cells found by Zero-Cost-PT (\texttt{random} discretization order) on the DARTS-S4 space using SVHN.}
  \label{fig:app:darts_sub_last}
 \vspace{-0.5cm}
\end{figure}

\begin{figure}[t]
 \centering
 \includegraphics[width=\textwidth]{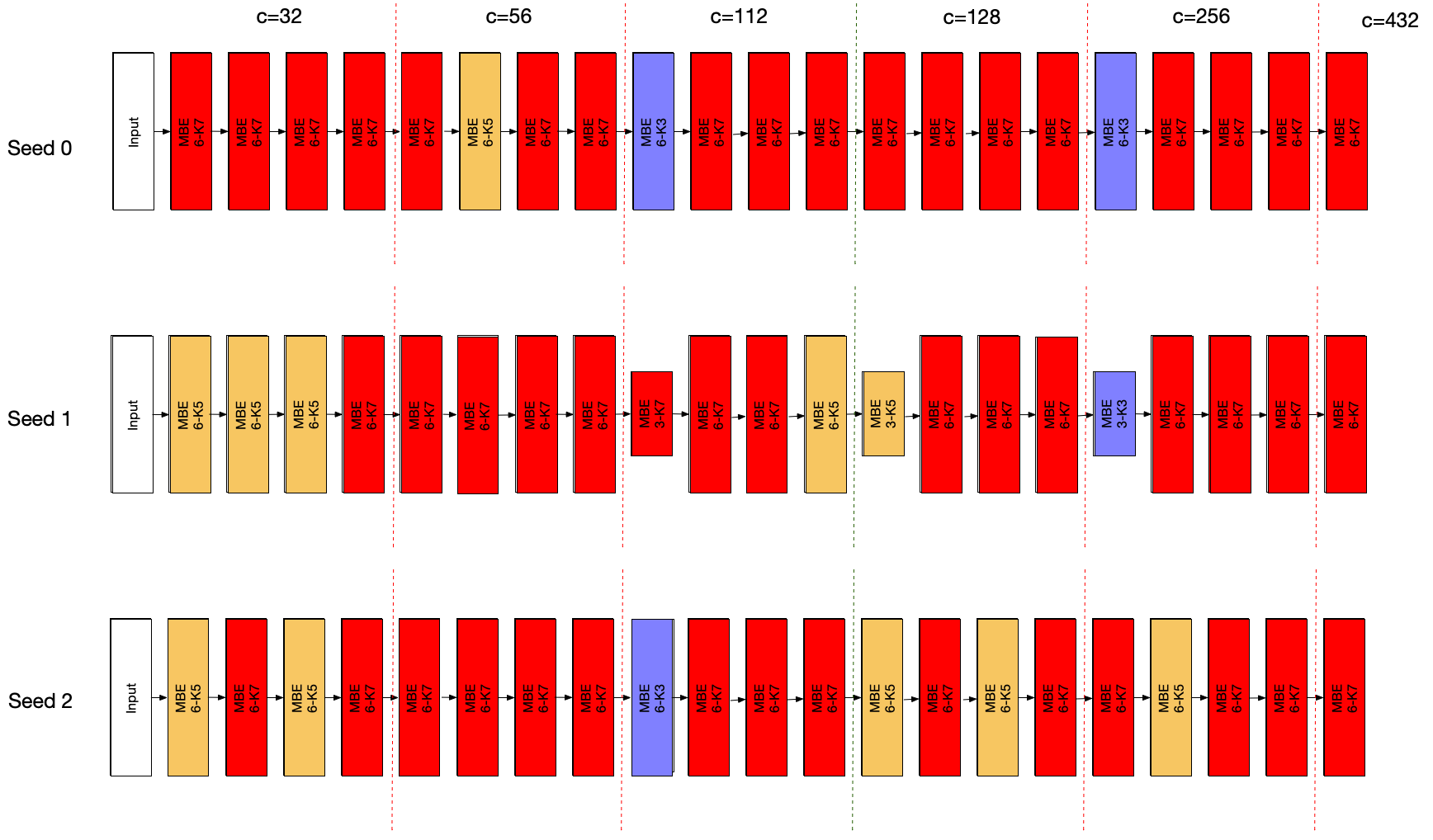}
 \caption{Discovered Architectures in MobileNet-like Search Space}
 \vspace{-0.5cm}
\label{fig:mobilnet-arch}
\end{figure}

Figures~\ref{fig:app:darts_global} and \ref{fig:app:darts_random} present cells found by our Zero-Cost-PT on the DARTS CNN search space (Section~\ref{sub:results-darts}) when using \texttt{global-op-iter} and \texttt{random} discretization orders, respectively (see Section~\ref{sec:zc-pt} for the definition of the two discretization orders).
Figures~\ref{fig:app:darts_sub_first} through \ref{fig:app:darts_sub_last} show cells discovered on the four DARTS subspaces and the three relevant datasets (Sections~\ref{sub:results-darts-sub} and \ref{sec:app:subspaces}). Figure~\ref{fig:mobilnet-arch} shows architectures found by our Zero-Cost-PT approach on MobileNet-like search space.

\subsection{Societal Impact and Limitations}
\label{sec:app:limitations}

In this paper we study operation scoring in the differentiable NAS context. We propose, evaluate and compare perturbation-based zero-cost operation scoring methods for differentiable NAS building upon recent work on training-free NAS proxies, leading to a lightweight NAS algorithm, Zero-Cost-PT. Our approach outperforms the best available differentiable architecture search in terms of searching time and accuracy even in very large search spaces. As our approach can significantly reduce searching time of NAS, especially for large search spaces, a potential negative societal impact is that GPU time needed to evaluate (training fully) the large number of discovered architectures can be substantial. 

One limitation of this work is that our analysis and ablations on operation scoring (Sections 3 and 4) is primarily conducted on tabular NAS benchmarks. Although the superior NAS performance of our Zero-Cost-PT on various larger search spaces indicates that the insights gained from our analysis can indeed transfer to other spaces, we didn't perform the same analysis on large search spaces such as DARTS due to the prohibitive cost, i.e. computing the correlations with oracle scoring methods will require the full training of all $10^{18}$ models, which is clearly not feasible.

In addition, the proposed approaches and analysis are primarily quantitative and empirical, and in this paper we did not try to provide theoretical explanations to questions revolving around the common problem of explaining why zero-cost proxies work well, or what makes a good zero-cost proxy, or how to select a reasonable zero-cost metric for a specific task. We think that those questions are of great interest to the whole community working on zero-cost proxies in NAS.
However, we find them extremely difficult to answer fully at this stage.
One of the reasons is simply the fact that those zero-cost proxies work under very different principles, and hence would require different approaches to be analysed -- if we strive for an answer that is well-grounded in theory, it will likely be different for proxies like \texttt{synflow} that utilize synaptic-flow and \texttt{nwot} that counts linear regions. 
At the same time, if we want a theoretically-grounded, analytical answer to those questions, we would need to understand in depth what properties are crucial for those zero-cost proxies to achieve high correlation with post-training accuracy. 
This naturally involves problems of analysing the (stochastic) training process, its generalization capabilities etc., which are all hard problems and areas of active research in modern ML.
Therefore, admittedly in the scope of this paper we do not have conclusive and qualitative answers to those questions.

However, in this paper we show that we can approach the problems from a quantitative perspective. In particular, the investigation in our paper provides one possible way to select (empirically) a reasonable zero-cost metric for a specific context, i.e., we could first conduct quantitative correlation analysis and ablation studies on available tabular NAS benchmarks, and then use the obtained insights (e.g., which metric/hyper-parameters to use) on the actual problem space. We demonstrate that even without qualitative analysis we can still design a robust and sound methodology by following a meticulous and systematic design process with clearly (and correctly) defined quantitative objectives, like those proposed in Section~\ref{sec:scoring}.
Our hope is that our results will be of interest to parts of the research community and will spur further research into low-cost NAS approaches.

\newpage

\begin{figure}[h]
 \centering
 \begin{subfigure}{0.45\columnwidth}
 \includegraphics[width=\textwidth]{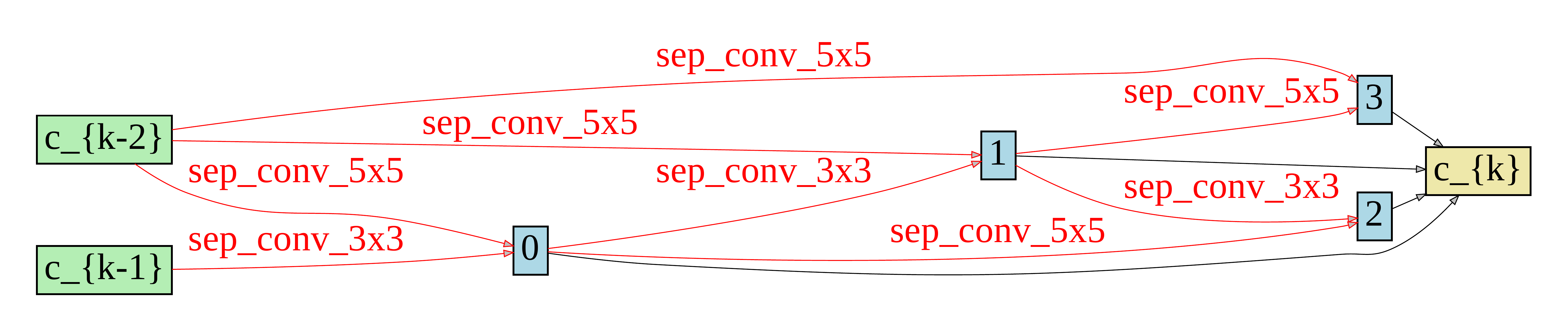}
  \caption{Normal cell}
 \end{subfigure} 
 \begin{subfigure}{0.54\columnwidth}
 \includegraphics[width=\textwidth]{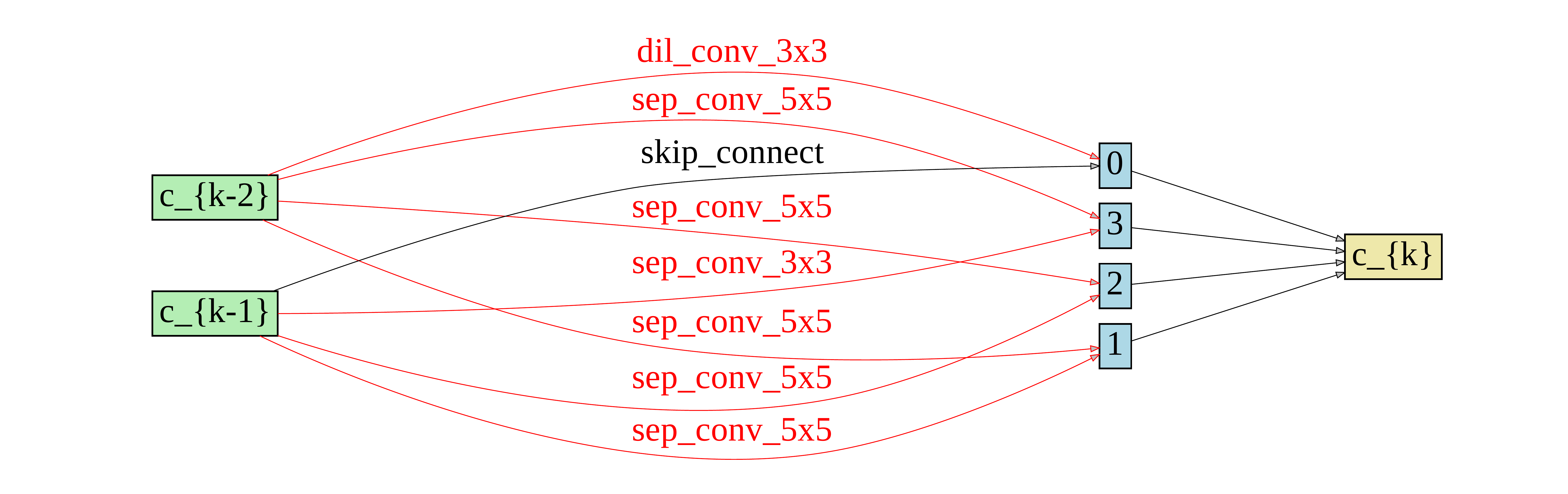}
  \caption{Reduction cell}
 \end{subfigure} 
  \caption{Cells found by Zero-Cost-PT (\texttt{random} discretization order) on the DARTS search space using ImageNet.}
  \label{fig:app:darts_random_im}
 \vspace{-0.5cm}
\end{figure}


\begin{figure}[t]
 \centering
 \begin{subfigure}{0.45\columnwidth}
 \includegraphics[width=\textwidth]{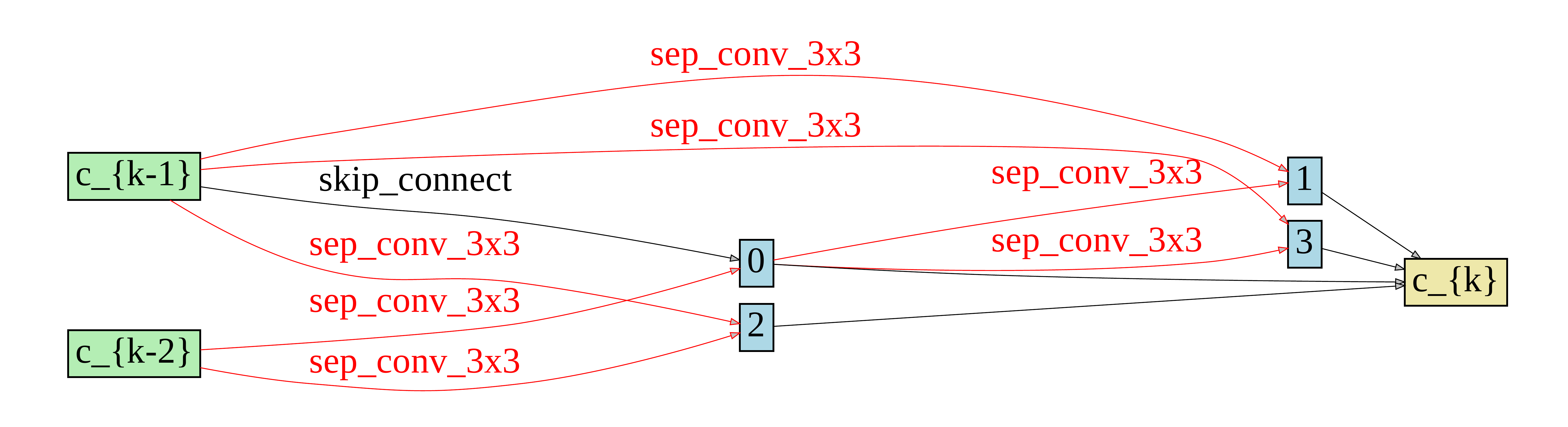}
  \caption{Normal cell}
 \end{subfigure} 
 \begin{subfigure}{0.54\columnwidth}
 \includegraphics[width=\textwidth]{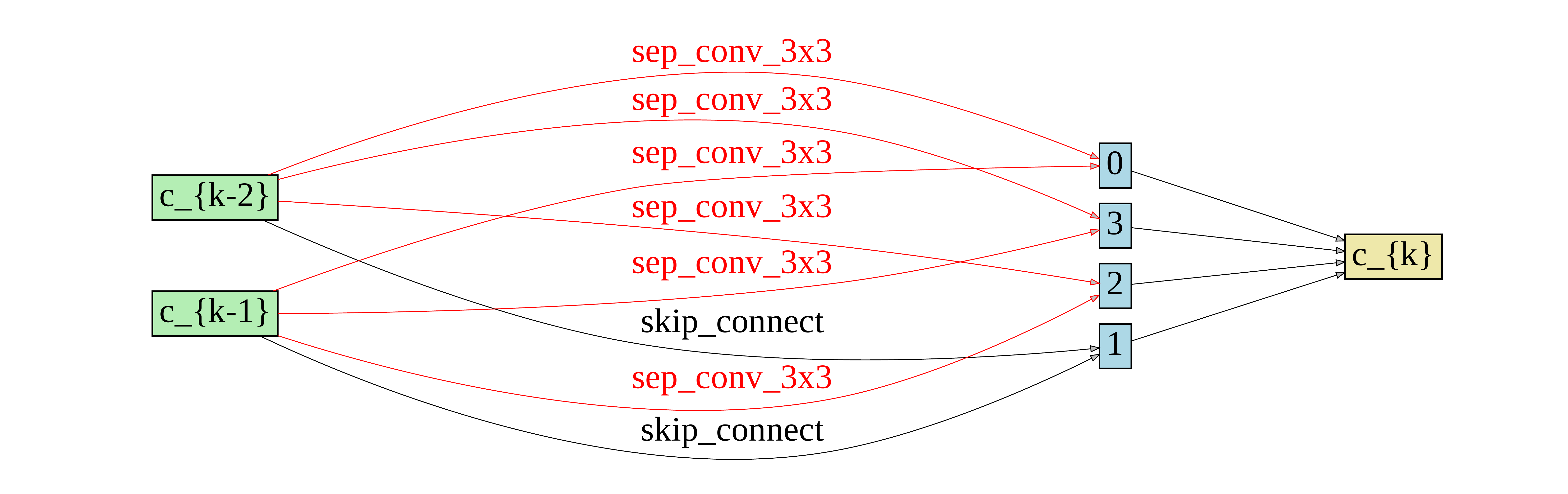}
  \caption{Reduction cell}
 \end{subfigure} 
  \caption{Cells found by Zero-Cost-PT (\texttt{random} discretization order) on the DARTS-S2 space using CIFAR-10.}
 \vspace{-0.5cm}
\end{figure}

\begin{figure}[h]
 \centering
 \begin{subfigure}{0.45\columnwidth}
 \includegraphics[width=\textwidth]{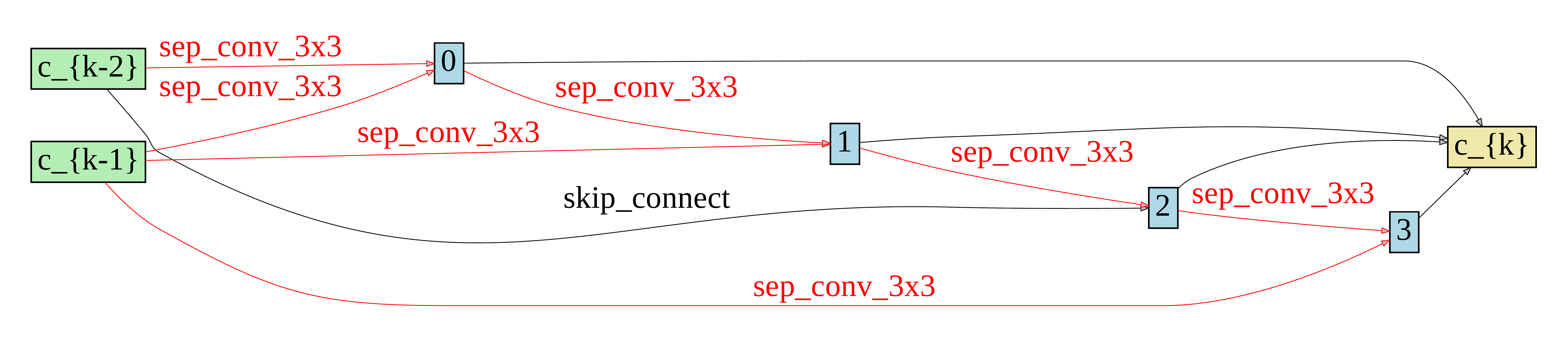}
  \caption{Normal cell}
 \end{subfigure} 
 \begin{subfigure}{0.54\columnwidth}
 \includegraphics[width=\textwidth]{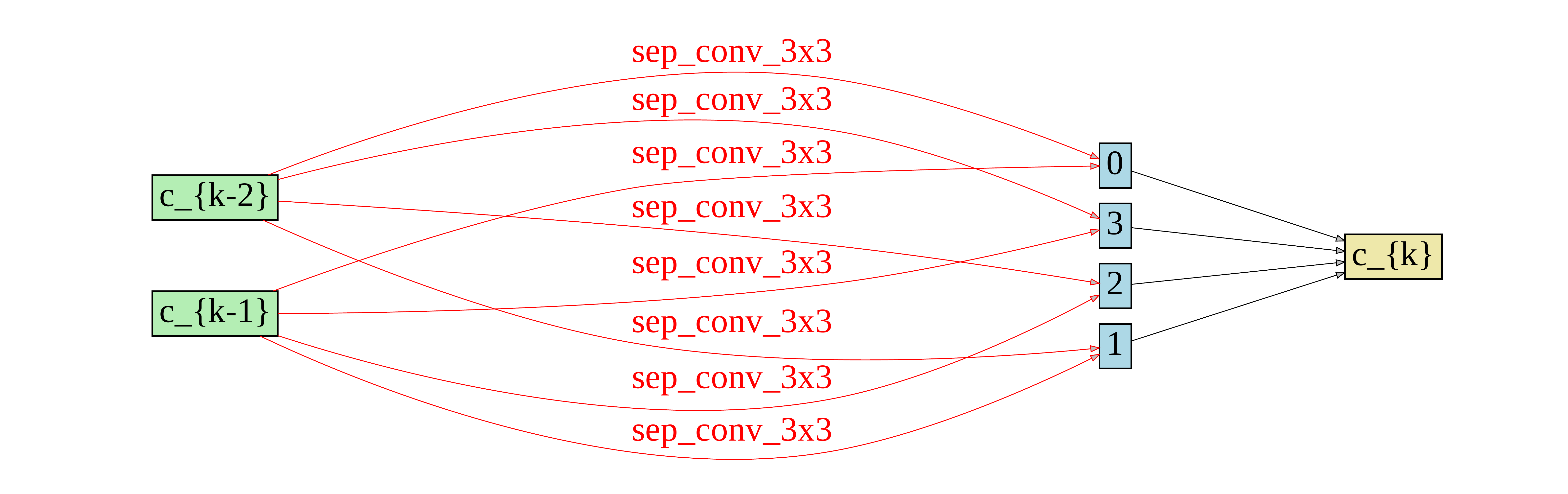}
  \caption{Reduction cell}
 \end{subfigure} 
  \caption{Cells found by Zero-Cost-PT (\texttt{random} discretization order) on the DARTS-S3 space using CIFAR-10.}
 \vspace{-0.5cm}
\end{figure}

\begin{figure}[h]
 \centering
 \begin{subfigure}{0.45\columnwidth}
 \includegraphics[width=\textwidth]{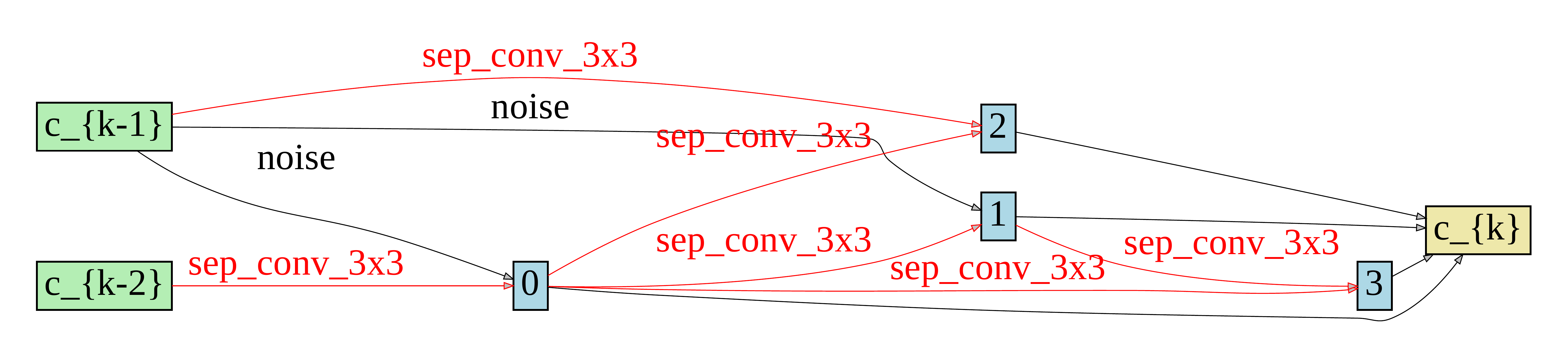}
  \caption{Normal cell}
 \end{subfigure} 
 \begin{subfigure}{0.54\columnwidth}
 \includegraphics[width=\textwidth]{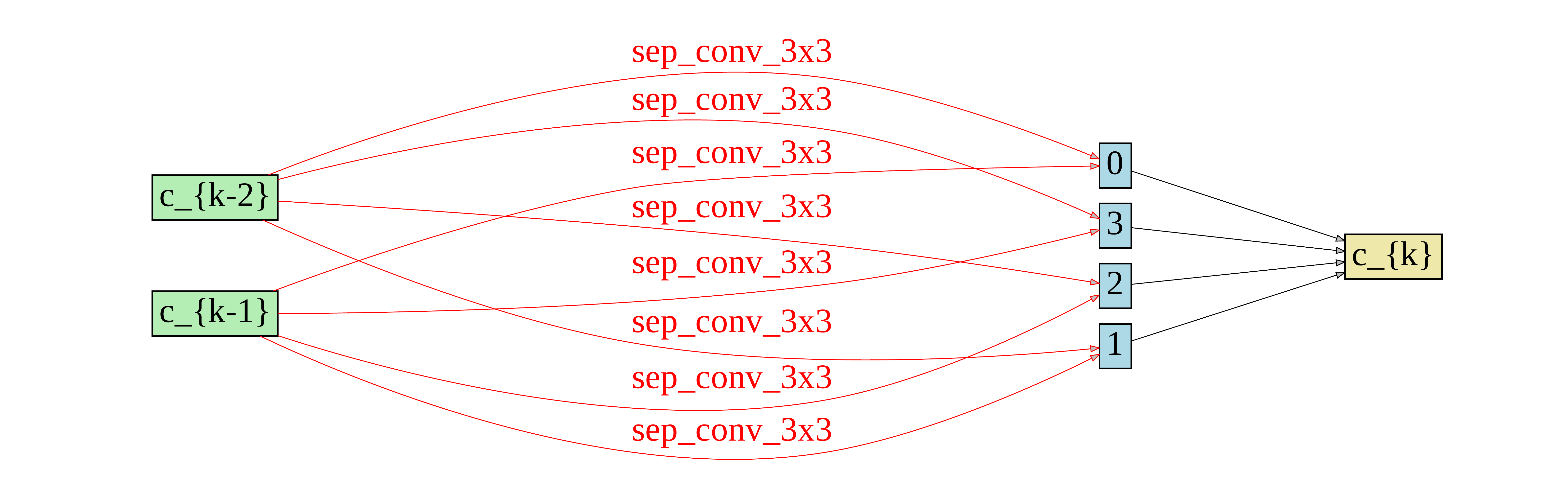}
  \caption{Reduction cell}
 \end{subfigure} 
  \caption{Cells found by Zero-Cost-PT (\texttt{random} discretization order) on the DARTS-S4 space using CIFAR-10.}
 \vspace{-0.5cm}
\end{figure}


\begin{figure}[h]
 \centering
 \begin{subfigure}{0.45\columnwidth}
 \includegraphics[width=\textwidth]{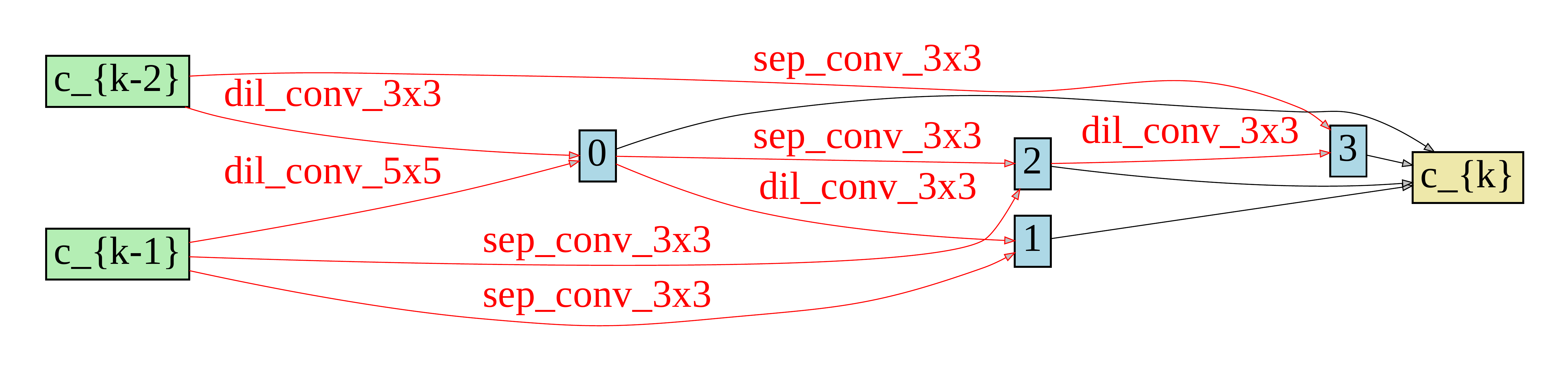}
  \caption{Normal cell}
 \end{subfigure} 
 \begin{subfigure}{0.54\columnwidth}
 \includegraphics[width=\textwidth]{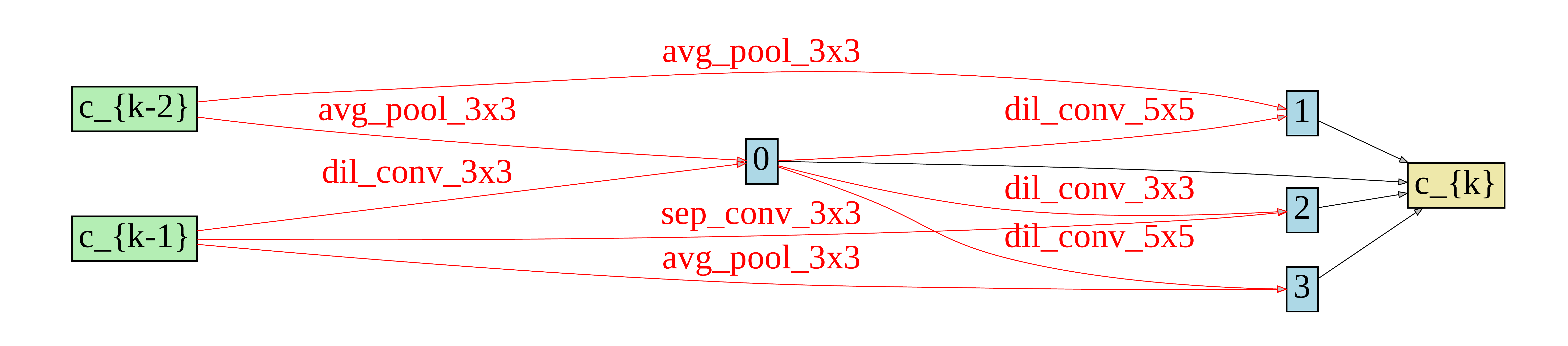}
  \caption{Reduction cell}
 \end{subfigure} 
  \caption{Cells found by Zero-Cost-PT (\texttt{random} discretization order) on the DARTS-S1 space using CIFAR-100.}
 \vspace{-0.5cm}
\end{figure}

\begin{figure}[h]
 \centering
 \begin{subfigure}{0.45\columnwidth}
 \includegraphics[width=\textwidth]{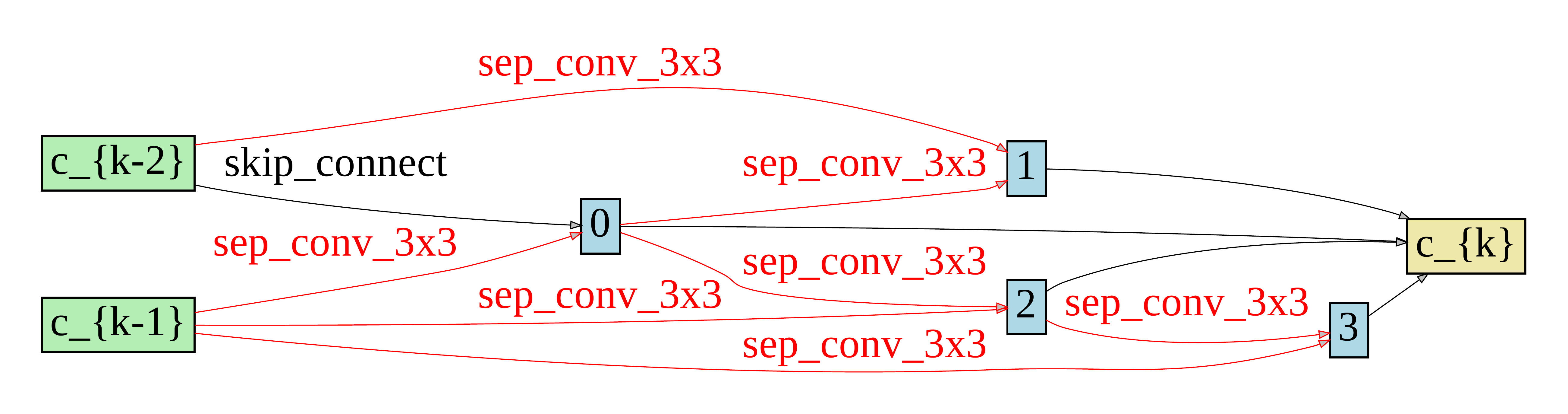}
  \caption{Normal cell}
 \end{subfigure} 
 \begin{subfigure}{0.54\columnwidth}
 \includegraphics[width=\textwidth]{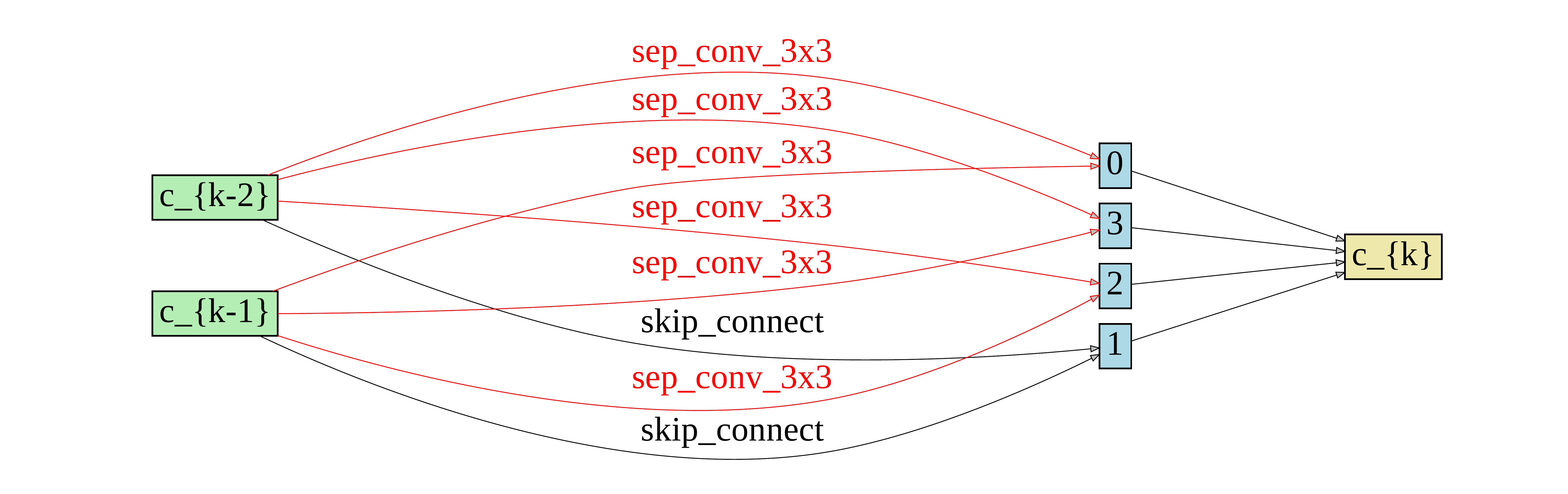}
  \caption{Reduction cell}
 \end{subfigure} 
  \caption{Cells found by Zero-Cost-PT (\texttt{random} discretization order) on the DARTS-S2 space using CIFAR-100.}
 \vspace{-0.5cm}
\end{figure}

\begin{figure}[h]
 \centering
 \begin{subfigure}{0.45\columnwidth}
 \includegraphics[width=\textwidth]{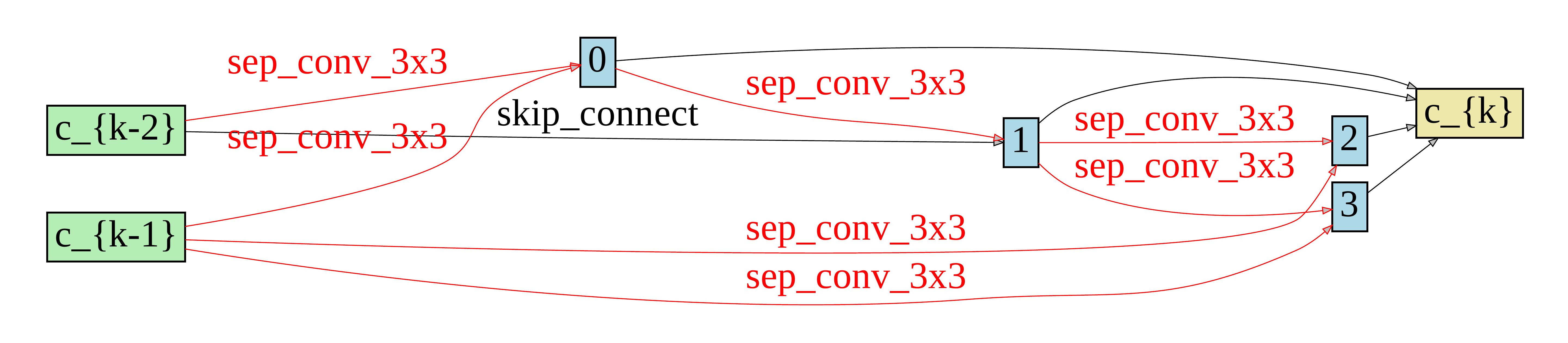}
  \caption{Normal cell}
 \end{subfigure} 
 \begin{subfigure}{0.54\columnwidth}
 \includegraphics[width=\textwidth]{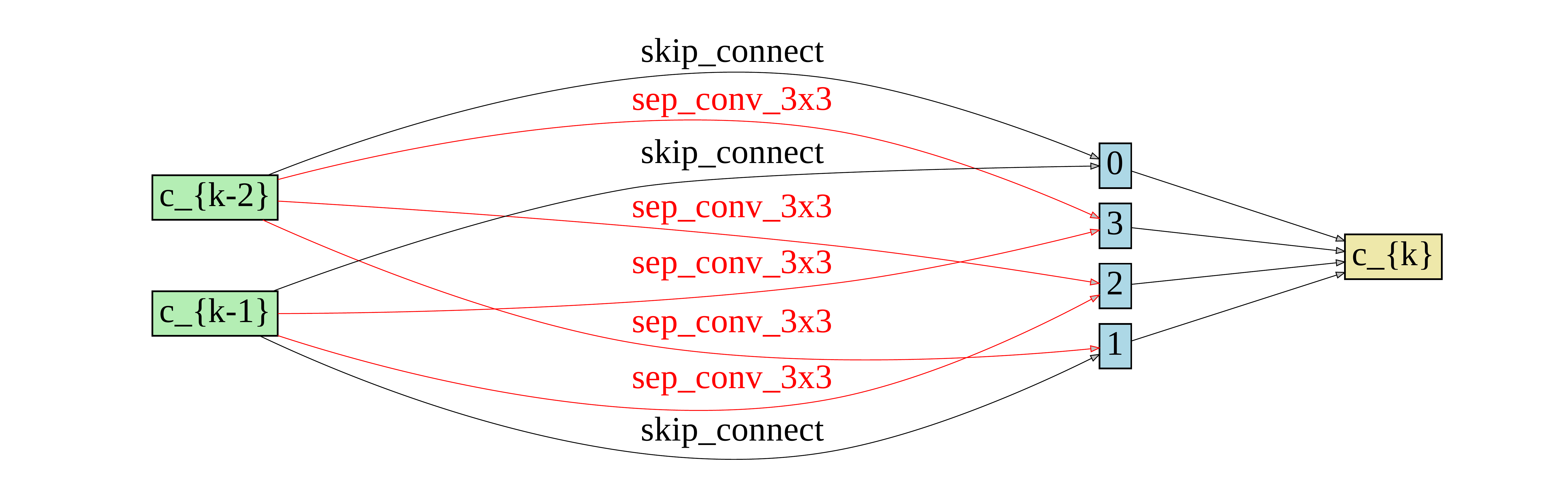}
  \caption{Reduction cell}
 \end{subfigure} 
  \caption{Cells found by Zero-Cost-PT (\texttt{random} discretization order) on the DARTS-S3 space using CIFAR-100.}
 \vspace{-0.5cm}
\end{figure}

\begin{figure}[h]
 \centering
 \begin{subfigure}{0.45\columnwidth}
 \includegraphics[width=\textwidth]{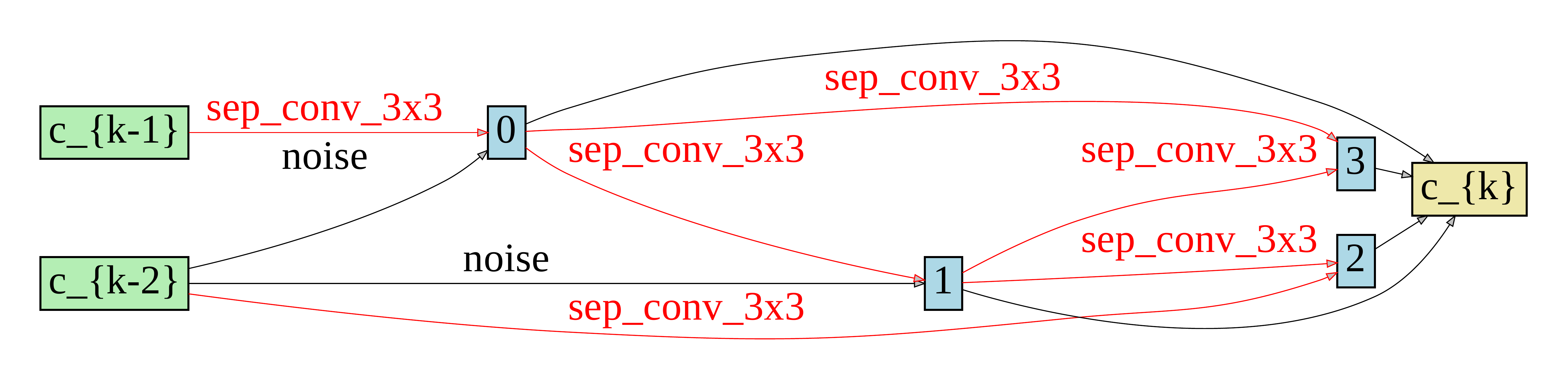}
  \caption{Normal cell}
 \end{subfigure} 
 \begin{subfigure}{0.54\columnwidth}
 \includegraphics[width=\textwidth]{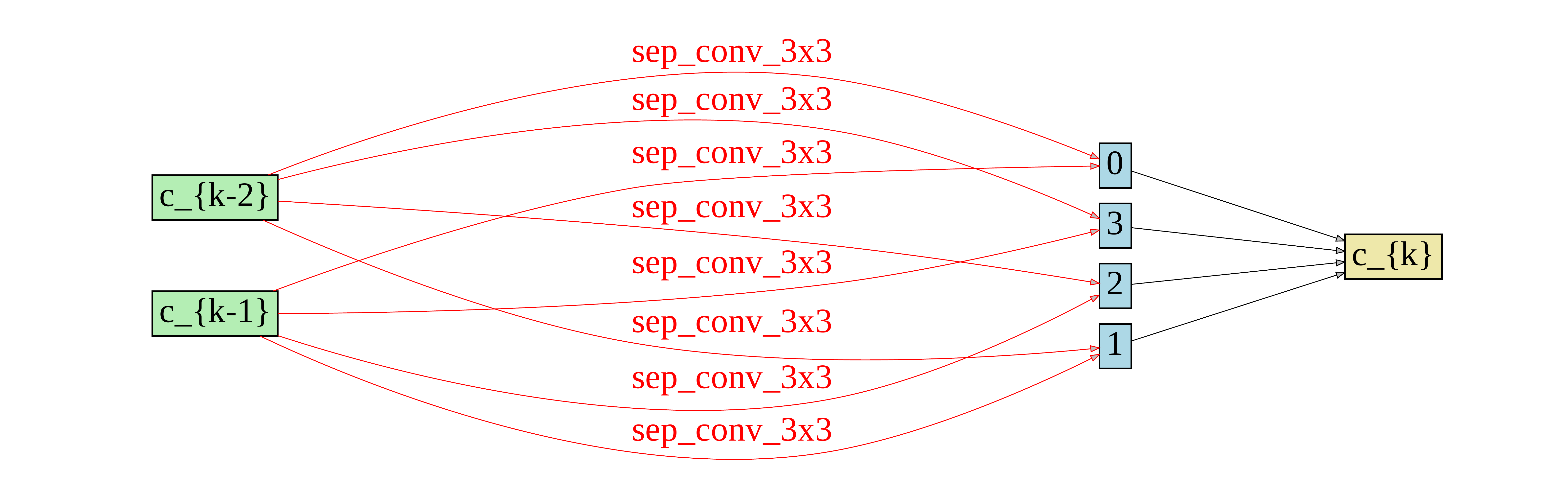}
  \caption{Reduction cell}
 \end{subfigure} 
  \caption{Cells found by Zero-Cost-PT (\texttt{random} discretization order) on the DARTS-S4 space using CIFAR-100.}
 \vspace{-0.5cm}
\end{figure}


\begin{figure}[h]
 \centering
 \begin{subfigure}{0.45\columnwidth}
 \includegraphics[width=\textwidth]{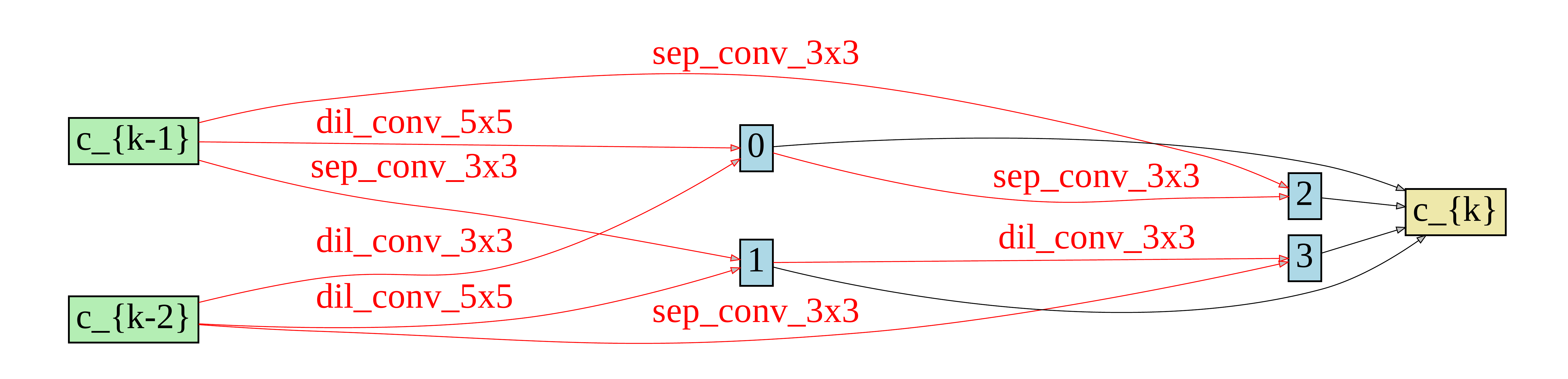}
  \caption{Normal cell}
 \end{subfigure} 
 \begin{subfigure}{0.54\columnwidth}
 \includegraphics[width=\textwidth]{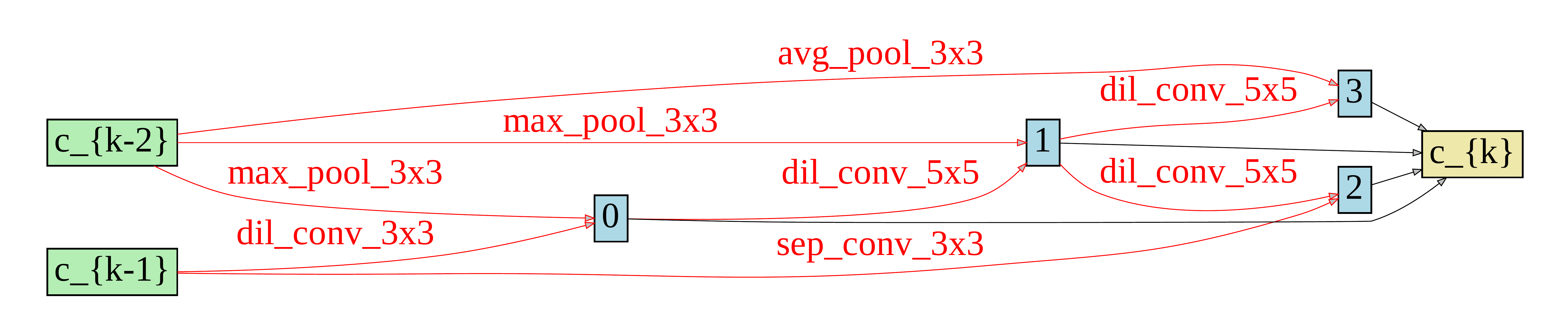}
  \caption{Reduction cell}
 \end{subfigure} 
  \caption{Cells found by Zero-Cost-PT (\texttt{random} discretization order) on the DARTS-S1 space using SVHN.}
 \vspace{-0.5cm}
\end{figure}

\begin{figure}[h]
 \centering
 \begin{subfigure}{0.45\columnwidth}
 \includegraphics[width=\textwidth]{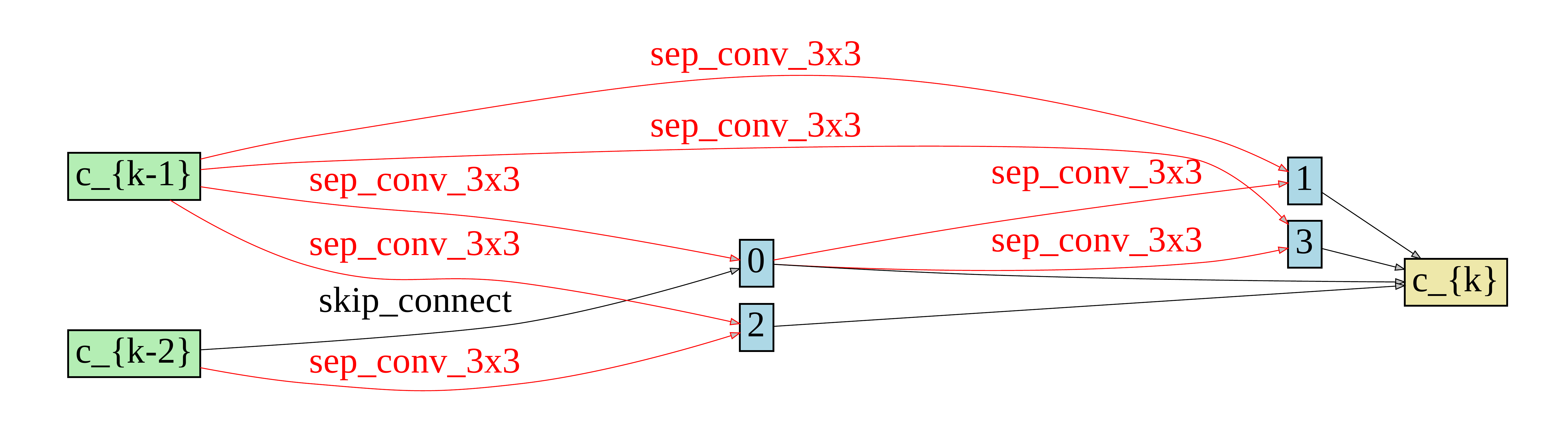}
  \caption{Normal cell}
 \end{subfigure} 
 \begin{subfigure}{0.54\columnwidth}
 \includegraphics[width=\textwidth]{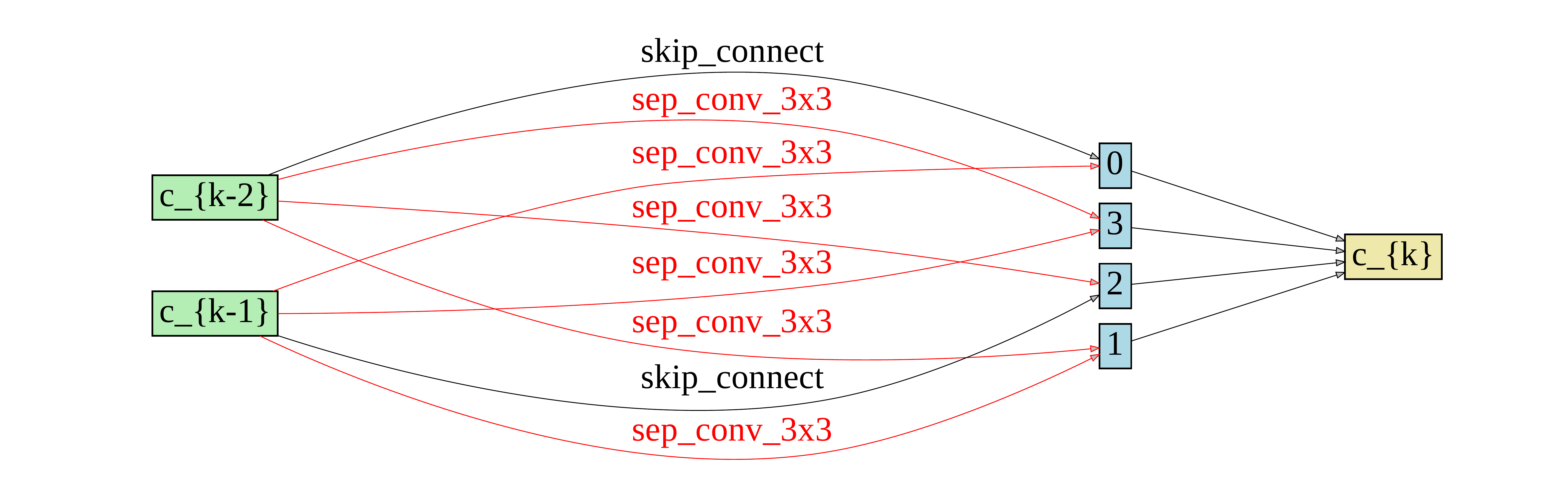}
  \caption{Reduction cell}
 \end{subfigure} 
  \caption{Cells found by Zero-Cost-PT (\texttt{random} discretization order) on the DARTS-S2 space using SVHN.}
 \vspace{-0.5cm}
\end{figure}

\begin{figure}[h]
 \centering
 \begin{subfigure}{0.45\columnwidth}
 \includegraphics[width=\textwidth]{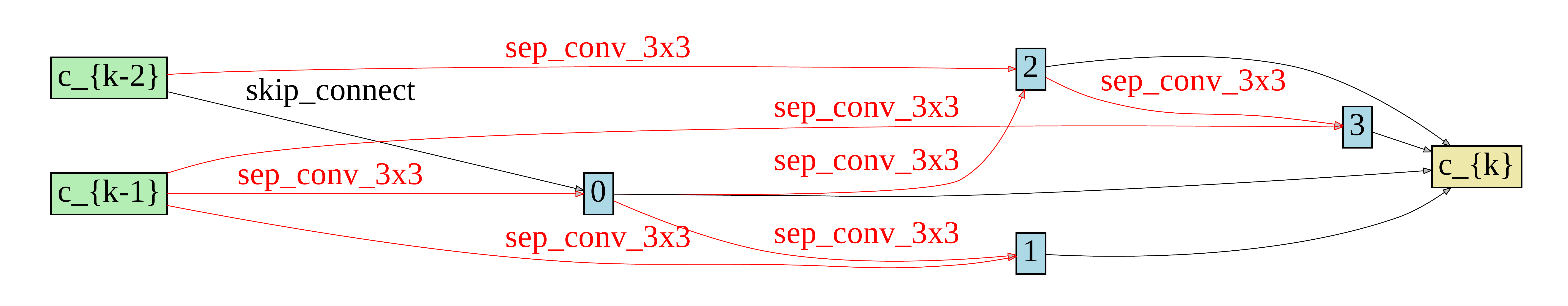}
  \caption{Normal cell}
 \end{subfigure} 
 \begin{subfigure}{0.54\columnwidth}
 \includegraphics[width=\textwidth]{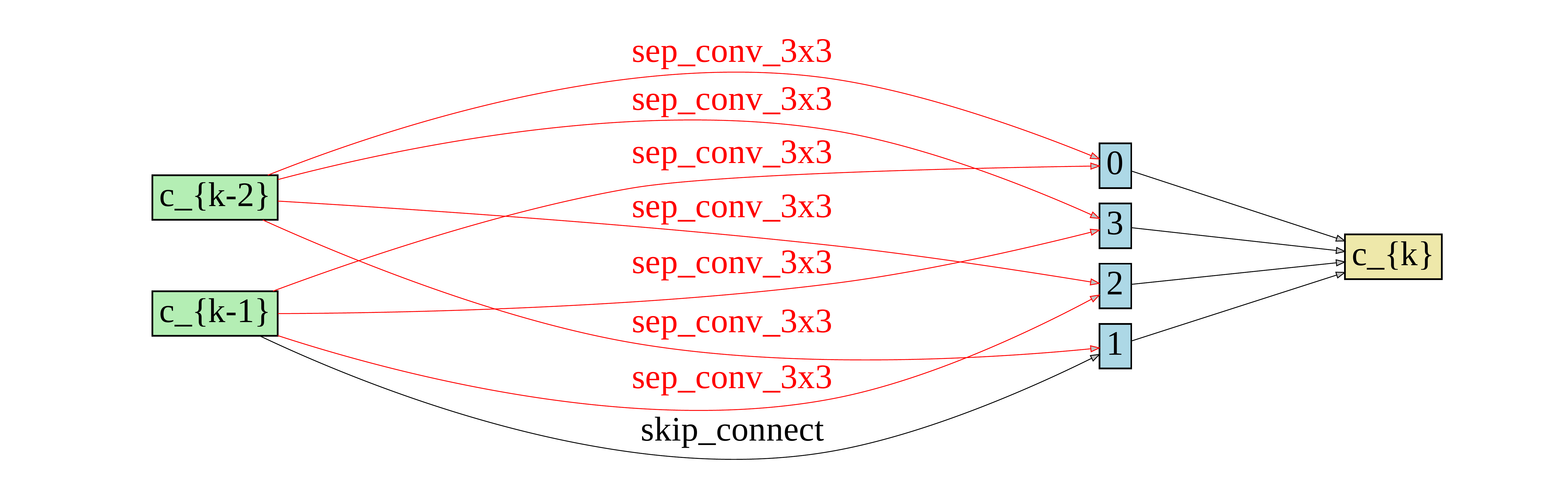}
  \caption{Reduction cell}
 \end{subfigure} 
  \caption{Cells found by Zero-Cost-PT (\texttt{random} discretization order) on the DARTS-S3 space using SVHN.}
 \vspace{-0.5cm}
\end{figure}

\input{91_metric-details}

%% file: 91_metric-details.tex
\begin{table}[t]
\centering
\caption {Raw values of operation scoring functions at iteration 0 to reproduce Figure~\ref{op}.}
\small 
\begin{threeparttable} 
\begin{tabular}{@{}lcccccc@{}} 
\toprule
& $edge\backslash op$ & none & skip\_connect & nor\_conv\_1x1 & nor\_conv\_3x3 & avg\_pool\_3x3  \\ \toprule
\parbox[t]{2mm}{\multirow{6}{*}{\rotatebox[origin=c]{90}{best-acc}}} & 0 & 94.15 & 94.18 & \secondbest{94.44} & \best{94.68} & 93.86 \\
 & 1 & 94.24 & 94.16 & \secondbest{94.49} & \best{94.68} & 94.09 \\
 & 2 & 94.25 & 94.43 & \secondbest{94.49} & \best{94.68} & 94.19 \\
 & 3 & \secondbest{94.16} & \best{94.68} & 94.03 & 94.04 & 93.85 \\
 & 4 & 94.29 & 94.18 & \secondbest{94.56} & \best{94.68} & 94.23 \\
 & 5 & 94.05 & 94.16 & \best{94.68} & \secondbest{94.56} & 94.1 \\
\midrule
\parbox[t]{2mm}{\multirow{6}{*}{\rotatebox[origin=c]{90}{avg-acc}}} & 0 & 77.36 & 81.02 & 83.81 & \secondbest{86.38} & \best{87.32} \\
 & 1 & 80.03 & 83.11 & \secondbest{85.23} & \best{85.99} & 81.52 \\
 & 2 & 82.9 & 82.44 & \secondbest{84.05} & \best{84.49} & 81.98 \\
 & 3 & 74.02 & 85.17 & \secondbest{87.3} & \best{88.28} & 81.38 \\
 & 4 & 80.14 & 83.05 & \secondbest{85.09} & \best{85.7} & 81.89 \\
 & 5 & 77.61 & 83.43 & \secondbest{86.18} & \best{86.95} & 81.74 \\
\midrule
\parbox[t]{2mm}{\multirow{6}{*}{\rotatebox[origin=c]{90}{disc-acc}}} & 0 & \best{83.27} & \secondbest{82.24} & 65.0 & 71.76 & 54.31 \\
 & 1 & \best{84.94} & 83.23 & 73.23 & 76.77 & \secondbest{83.45} \\
 & 2 & \best{83.87} & \secondbest{83.73} & 77.33 & 76.83 & 83.25 \\
 & 3 & 65.77 & \best{84.44} & 75.82 & \secondbest{78.68} & 62.7 \\
 & 4 & \best{83.57} & 82.03 & 75.02 & 76.09 & \secondbest{82.56} \\
 & 5 & \best{83.95} & \secondbest{82.45} & 66.69 & 71.36 & 80.31 \\
\midrule
\parbox[t]{2mm}{\multirow{6}{*}{\rotatebox[origin=c]{90}{darts-pt\tnote{1}}}} & 0 & -85.43 & \best{-17.02} & -78.13 & \secondbest{-59.09} & -85.34 \\
 & 1 & -85.52 & \best{-36.1} & -84.39 & \secondbest{-80.95} & -85.49 \\
 & 2 & -85.51 & \secondbest{-80.29} & -81.86 & \best{-77.68} & -85.32 \\
 & 3 & -85.49 & \best{-9.86} & -81.79 & \secondbest{-59.18} & -85.48 \\
 & 4 & -85.45 & \best{-51.15} & -78.84 & \secondbest{-64.64} & -85.14 \\
 & 5 & -85.54 & \best{-32.43} & -81.04 & \secondbest{-72.75} & -85.51 \\
\midrule
\parbox[t]{2mm}{\multirow{6}{*}{\rotatebox[origin=c]{90}{disc-zc}}} & 0 & 3331.01 & \best{3445.49} & 3366.88 & \secondbest{3437.55} & 3423.18 \\
 & 1 & 3429.07 & \best{3435.75} & 3407.87 & \secondbest{3434.58} & 3421.44 \\
 & 2 & 3428.8 & 3423.36 & \best{3440.93} & \secondbest{3437.29} & 3416.89 \\
 & 3 & 3408.99 & \best{3464.05} & 3359.89 & 3382.18 & \secondbest{3431.81} \\
 & 4 & \secondbest{3433.99} & \best{3435.57} & 3424.47 & 3431.14 & 3423.15 \\
 & 5 & \secondbest{3434.42} & \best{3437.66} & 3418.57 & 3397.52 & 3424.17 \\
\midrule
\parbox[t]{2mm}{\multirow{6}{*}{\rotatebox[origin=c]{90}{zc-pt\tnote{1}}}} & 0 & -3455.23 & -3449.9 & \secondbest{-3449.54} & \best{-3441.82} & -3461.18 \\
 & 1 & -3452.15 & -3448.7 & \secondbest{-3441.81} & \best{-3440.65} & -3453.74 \\
 & 2 & -3446.52 & -3447.61 & \best{-3435.46} & \secondbest{-3436.4} & -3449.28 \\
 & 3 & -3453.81 & \best{-3435.99} & \secondbest{-3444.04} & -3445.6 & -3447.07 \\
 & 4 & -3451.06 & -3449.8 & \secondbest{-3442.63} & \best{-3441.13} & -3453.31 \\
 & 5 & -3450.97 & -3448.21 & \best{-3440.8} & \secondbest{-3443.24} & -3452.99 \\
\midrule
\parbox[t]{2mm}{\multirow{6}{*}{\rotatebox[origin=c]{90}{darts}}} & 0 & 0.14 & \best{0.48} & 0.13 & \secondbest{0.18} & 0.07 \\
 & 1 & 0.12 & \best{0.55} & 0.11 & \secondbest{0.12} & 0.09 \\
 & 2 & \secondbest{0.24} & \best{0.33} & 0.15 & 0.17 & 0.11 \\
 & 3 & 0.06 & \best{0.65} & 0.08 & \secondbest{0.13} & 0.07 \\
 & 4 & 0.12 & \best{0.48} & 0.13 & \secondbest{0.17} & 0.1 \\
 & 5 & \secondbest{0.16} & \best{0.49} & 0.12 & 0.14 & 0.09 \\
 \midrule
\parbox[t]{2mm}{\multirow{6}{*}{\rotatebox[origin=c]{90}{tenas}}} 
 & 0 & -38.5 & -48.0 & \secondbest{-31.0} & \best{-6.0} & -37.5 \\
 & 1 & \best{-7.0}  & -55.0 & \secondbest{-10.0} & -15.0 & -39.0 \\
 & 2 & -31.5 & \best{-10.0} & -30.0 & \secondbest{-16.5} & -36.5 \\
 & 3 & -34.0 & -44.0 & -53.5 & \best{-23.0} & \secondbest{-30.0} \\
 & 4 & \secondbest{-32.0} & -32.5 & -36.5 & \best{-32.0} & -52.0 \\
 & 5 & -38.5 & \best{-16.0} & -20.0 & \secondbest{-17.0} & -27.5 \\
\bottomrule
\end{tabular}
\begin{tablenotes}
\scriptsize
\item[1] Lower is better so we add a negative sign to *-pt scores.
\end{tablenotes}
\end{threeparttable}
\label{tbl:raw-op-scores}
\end{table}